\documentclass[nohyperref]{article}

\usepackage{microtype}
\usepackage{graphicx}
\usepackage{booktabs} % for professional tables

\usepackage{hyperref}

\usepackage[accepted]{icml2022}

\usepackage{amsmath}
\usepackage{amssymb}
\usepackage{mathtools}
\usepackage{amsthm}

\usepackage[capitalize,noabbrev]{cleveref}

\theoremstyle{plain}
\newtheorem{theorem}{Theorem}[section]

\newtheorem{lemma}[theorem]{Lemma}
\newtheorem{corollary}[theorem]{Corollary}
\theoremstyle{definition}
\newtheorem{definition}[theorem]{Definition}

\theoremstyle{remark}

\usepackage[textsize=tiny]{todonotes}

\usepackage{amsmath}
\usepackage{amssymb}
\usepackage{mathtools}
\usepackage{subcaption}
\usepackage{multirow}
\usepackage{adjustbox}
\usepackage{xcolor}
\usepackage{enumitem}
\usepackage{booktabs}
\usepackage{pifont}
\usepackage{dsfont}
\usepackage{xstring}
\usepackage{textcomp}
\usepackage{algorithm, algorithmic}
\usepackage[capitalize]{cleveref}
\usepackage{balance}

\newcommand{\eg}{{\it e.g.,\ }}
\newcommand{\ie}{{\it i.e.,\ }}
\newcommand{\E}{\mathbb{E}}
\newcommand{\Loss}{\mathcal{L}}
\newcommand{\Arch}{\mathcal{A}}

\newcommand{\argmax}{\mathop{\arg\max}}

\newcommand{\acc}{\mathrm{acc}}
\newcommand{\flops}{\mathrm{flops}}
\newcommand{\params}{\mathrm{params}}

\newcommand{\diff}[1]{\IfSubStr{#1}{+}{{\color{blue}#1}}{{\color{red}#1}}}
\newcommand{\meanstd}[2]{#1{\footnotesize $\pm$#2}}

\newcommand{\highlight}[1]{\noindent\textbf{#1}}

\newcommand{\li}{\mathrm{Li}}
\newcommand{\NDCG}{\textrm{NDCG}}
\newcommand{\NDCGExpr}{\NDCG(f; \mathbf{a}, \mathbf{r})}
\newcommand{\TopK}{\textrm{TopK}}
\newcommand{\TopKExpr}{\TopK(f; \mathbf{a}, \mathbf{r}, k)}
\newcommand{\pif}{\pi_{f}}
\newcommand{\logsum}[3][i]{\ensuremath{\sum_{#1=#2}^{#3} \frac{1}{\log_2(#1+1)}}}
\newcommand{\relexpsum}[3][i]{\ensuremath{\sum_{#1=#2}^{#3} 2^{r_{#1}}}}
\newcommand{\rstar}{r^{*}}
\newcommand{\invlog}[1]{\ensuremath{\frac{1}{\log_2(#1+1)}}}
\newcommand{\cnu}[1]{\ensuremath{c^{(\nu,k)}_{#1}}}

\allowdisplaybreaks

\newcommand{\algname}{AceNAS}
 \setlist[itemize]{align=parleft,left=0pt..1em}
 \usepackage[compact]{titlesec}
 \usepackage[skip=1ex]{caption}
 \titlespacing{\section}{0pt}{2ex}{0.5ex}
 \titlespacing{\subsection}{0pt}{0.5ex}{0ex}
 \titlespacing{\subsubsection}{0pt}{0.5ex}{0ex}
\icmltitlerunning{\algname{}}

\begin{document}

\twocolumn[
\icmltitle{Learning to Rank Ace Neural Architectures via \\ Normalized Discounted Cumulative Gain}
% \icmltitle{\algname{}: Learning to Rank Ace Neural Architectures with \\ Weak Supervision of Weight Sharing}



\icmlsetsymbol{equal}{*}

\begin{icmlauthorlist}
\icmlauthor{Yuge Zhang}{ms}
\icmlauthor{Quanlu Zhang}{ms}
\icmlauthor{Li Lyna Zhang}{ms}
\icmlauthor{Yaming Yang}{ms}
\icmlauthor{Chenqian Yan}{xmu}
\icmlauthor{Xiaotian Gao}{ms}
\icmlauthor{Yuqing Yang}{ms}

\end{icmlauthorlist}

\icmlaffiliation{ms}{Microsoft Research}
\icmlaffiliation{xmu}{Xiamen University, work done as an intern at MSRA}

\icmlcorrespondingauthor{Yuge Zhang}{Yuge.Zhang@microsoft.com}

\icmlkeywords{Machine Learning, ICML}

\vskip 0.3in
]



\printAffiliationsAndNotice{\icmlEqualContribution} % otherwise use the standard text.

\begin{abstract}
One of the key challenges in Neural Architecture Search (NAS) is to efficiently rank the performances of architectures. The mainstream assessment of performance rankers uses ranking correlations (\eg Kendall's tau), which pay equal attention to the whole space. However, the optimization goal of NAS is identifying top architectures while paying less attention on other architectures in the search space. In this paper, we show both empirically and theoretically that Normalized Discounted Cumulative Gain (NDCG) is a better metric for rankers. Subsequently, we propose a new algorithm, \algname{}, which directly optimizes NDCG with LambdaRank. It also leverages weak labels produced by weight-sharing NAS to pre-train the ranker, so as to further reduce search cost. Extensive experiments on 12 NAS benchmarks and a large-scale search space demonstrate that our approach consistently outperforms SOTA NAS methods, with up to 3.67\% accuracy improvement and 8$\times$ reduction on search cost.

\end{abstract}

\section{Introduction}

% Neural Architecture Search (NAS) has shown its effectiveness on various tasks including computer vision~\cite{zoph2018learning}, natural language processing~\cite{so2019evolved}, and is increasingly spanning to more domains and tasks. Many popular models (\eg MobileNetV3~\cite{howard2019searching}, EfficientNet~\cite{tan2020efficientnet}) and effective modules (\eg Swish~\cite{ramachandran2017searching}) are found with the help of NAS. %Recent trend shows that NAS has become an increasingly important role on innovating new neural architectures.

Neural Architecture Search (NAS) has shown its effectiveness on various tasks~\cite{howard2019searching,tan2020efficientnet,so2019evolved}. Its core idea is to automatically design suitable network architectures for a given dataset.
Although early NAS~\cite{zoph2016neural} is successful in outperforming human-designed models, its computational cost can be prohibitive, because it takes up to 1,800 GPU days to train thousands of architectures. % For example, early NAS works~\cite{zoph2016neural} % that adopts Reinforcement Learning as search strategy
% takes 1,800 GPU days to train thousands of architectures.
Such excessive training cost motivates the followed-up works to develop efficient strategies to estimate the rankings of architecture performances. % accelerate performance estimation~\cite{elsken2018neural}, by developing efficient \emph{performance rankers} to \emph{rank} architectures with partial-trained or even untrained~\cite{mellor2020neural} neural networks.
A reliable \emph{performance ranker}~\cite{elsken2018neural,white2021powerful} is known to be a key component in NAS to help quickly filter out the bad-performing architectures and identify the best.
% Representative works include performance predictors~\cite{luo2018neural,wen2020neural,wei2020npenas} and weight sharing based methods~\cite{bender2018understanding,liu2018darts}.  \lz{what's the connection between weight sharing and performance predictors?}
% Early NAS~\cite{zoph2016neural} adopt Reinforcement Learning (RL) as their search strategy. Although they have outperformed hand-crafted architectures, the excessive computation cost (\eg 1,800 GPU days) makes those methods impractical.
% A series of follow-up works are done to reduce the search cost. For example, weight-sharing NAS~\cite{bender2018understanding,liu2018darts} trains one super-net to estimate the performance of all candidates. Performance predictors~\cite{luo2018neural,wen2020neural,wei2020npenas} build a estimator that... Recently, the performance of an architecture can be even inferred without training~\cite{mellor2020neural}. Despite the difference in efficiency and effectiveness, all these works share the same goal, \ie to build a \emph{performance ranker} that \emph{ranks} the architectures, such that the performances of top architectures suggested are maximized.

\begin{figure}[t]
\centering
\includegraphics[width=\linewidth]{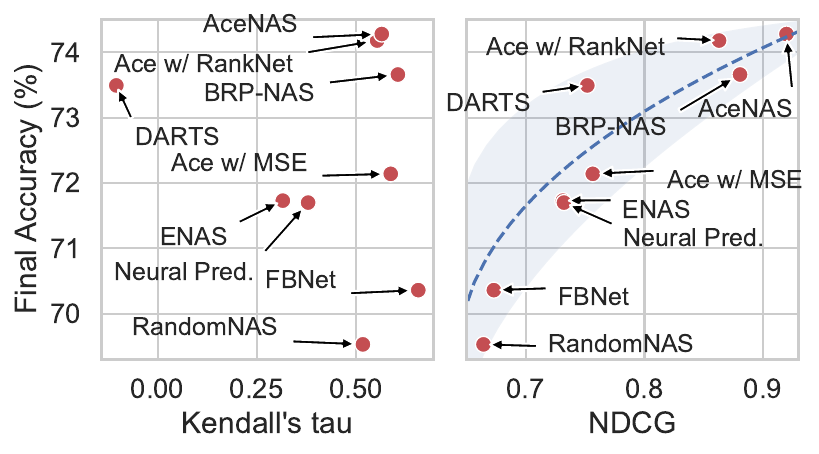}
\caption{We evaluate KdT, NDCG and final NAS performance of 11 rankers. \textbf{(Left)} Better KdT does not necessarily yield better accuracy. \textbf{(Right)} NDCG and accuracy are clearly positively correlated.}
\label{fig:kendalltau-vs-ndcg-1}
\end{figure}

This paper firstly concerns the assessment of performance rankers. Other than reporting the performance of final searched architecture, many works~\cite{white2021powerful,wen2020neural,pourchot2020share,dong2020nasbench201,zhao2021few,guo2020single} have used ranking correlations like \emph{Kendall's tau (KdT)}, to demonstrate the robustness of their method, or to achieve a deeper understanding.
Although widely used, whether such metric is actually eligible to assess a ranker's quality remains in doubt.
To answer this question, we curate 11 state-of-the-art NAS performance rankers from NNI~\cite{zhang2020retiarii} as well as \algname{} (described below), and measure their KdT and final accuracy on NAS-Bench-201. The result, shown in the left figure of \cref{fig:kendalltau-vs-ndcg-1}, is surprising: \emph{(i)} the worst and the best ranker both have KdT around only 0.5, yet their final accuracy differ a lot; \emph{(ii)} although some rankers have poor KdT, their final accuracy can be still very competitive. These observations suggest that KdT is not a good assessment metric in NAS, which we believe might mislead research directions.
% However, the NAS community almost takes such metric for granted \yuge{too sharp}, and few efforts are devoted to verifying whether such metric is actually qualified to assess a ranker's quality. Surprisingly, by experimenting on a variety to rankers \yym{not very clear} \yuge{context: sota, controlled setting}, we find that Kendall's tau might not be a good metric for performance rankers. In the left figure of \cref{fig:kendalltau-vs-ndcg-1}, we see that a ranker with higher Kendall's tau does not necessarily achieve better accuracy in an end-to-end NAS experiment \yym{the argument is not that strong, better to give more comprehensive evidences} \yuge{More evidences are in sec 3?} \yuge{can't difference best rankers, worst doesn't necessarily worst}. \yuge{not the best indicator}
% The wide adoption of Kendall's tau can be misleading for the NAS community, catalyzing researches that treat all the architectures equally~\cite{wen2020neural}, or even minimize pairwise ranking errors and implicitly maximize ranking correlation~\cite{chau2020brp,xu2021renasrelativistic}.
% Unfortunately, the wide adoption of Kendall's tau has already misled 
% This motivates us to find a better objective to measure and optimize a ranker.
% Specifically, it is to measure the correlation between the predicted rank and ground-truth rank (\ie rank of full-training performances).

We identify one of the reasons why KdT is not a good metric to be, KdT pays equal attention to all the architectures in search space. It can not reflect the effectiveness of a ranker when it is used in a NAS experiment, which only cares about the top-tier architectures. Recently, we notice another popular ranking metric named \emph{Normalized Discounted Cumulative Gain (NDCG)}. Unlike KdT which considers every ranked items equally, \emph{NDCG} attaches larger weights to the best items, thus gives more attention to most important items. NDCG is widely used to assess \emph{Learning to Rank (LTR)} algorithms in Information Retrieval (IR) community. %Furthermore, how to maximize NDCG has also been a well-studied problem in IR. It seems promising to adapt NDCG to NAS, yet there are two research challenges.
Inspired by this, it is intuitive to ask \textit{can we apply NDCG in NAS to better rank top-performing architectures?}

%In this paper, we propose to apply NDCG, which focus more on top-performing items, as the performance ranking metric in NAS. However, we face two major challenges.

Despite the success of NDCG in IR~\cite{liu2011learning}, we face two major challenges when using NDCG to find top-performing architectures in NAS.
Firstly, NDCG was initially designed for document ranking problems. The applicability of NDCG in NAS is only an intuition, which lacks theoretical and empirical support.
Secondly, since it is time-consuming to obtain the train-from-scratch accuracy of a neural network, it is challenging to efficiently get a large number of architectures' accuracies to optimize the NDCG in NAS. 
%it is challenging to efficiently compute NDCG in NAS. 
%as architectures' accuracies are very costly to obtain, techniques to cut down the need of ``labelled'' architectures are much desired.
% \yuge{not an identical problem. What is NDCG. Can it be applied? (i) can NDCG be applied to NAS with final accuracy? continuous vs discrete, no theory to support (ii) often requires ground-truth label to learn, but ground-truth label is costly. Contribution: Apply and theoretically prove, weight sharing.} %\yuge{A parallel research community... inspire us to apply. but how to apply? final goal is different, ground-truth is lacking}
% We also observe that NAS is very similar with document ranking problem in Information Retrieval (IR), as they are both trying to identify top architectures/documents. Interestingly, the IR community once took a similar detour to setup objectives that treat all documents equally~\cite{fuhr1989optimum}, but nowadays it has been a common sense that these objectives are not reasonable.
% Thus, we revisit NAS from the perspective of \emph{Learning to Rank (LTR)}~\cite{liu2011learning} from IR, and hypothesis that \emph{Normalized Discounted Cumulative Gain (NDCG)}~\cite{jarvelin2002cumulated} is a better metric for performance rankers. Unlike correlation metrics (\eg Kendall’s tau) which consider every architecture equally, \emph{NDCG} attaches larger weights to the best architectures, thus gives more attention on top-performing ones. It is shown in right figure of \cref{fig:kendalltau-vs-ndcg-1} that NDCG and final performance are positively correlated.

To tackle the above challenges, we first show that NDCG, with careful adaptation, is a good metric for performance rankers in NAS. This is proved both theoretically and empirically. We then propose a novel algorithm called \algname{}. We utilize LambdaRank~\cite{burges2007learning}, a list-wise LTR approach, to directly maximize NDCG of a ranker. % This relaxes the optimization objective to focus more on top architectures, making the optimization easier.
To reduce the time cost of computing ground-truth architecture-accuracy pairs, we leverage the weak supervision obtained from weight-sharing NAS. Though the rank obtained with weight sharing is not accurate~\cite{niu2020disturbanceimmune}, it contains implicit semantics of architectures~\cite{zhang2020does} that is helpful for down-streaming tasks (\eg accuracy prediction) if used properly.
To the end, \algname{} is a two-stage algorithm, where we first pre-train the ranking model on massive number of inaccurate but easily-obtained weight-sharing labels, and then fine-tune the ranker with only a small number of fully-trained models' accuracies with LambdaRank. % The whole process is illustrated in \cref{fig:method}.

We conduct extensive experiments on 12 combinations of search spaces and datasets with benchmarks and ProxylessNAS search space~\cite{cai2019proxylessnas}. The results demonstrate that {\algname{}} consistently outperforms the SOTA performance ranker. Specifically, we achieve a same-level accuracy with only 110 ground-truth architectures on NAS-Bench-101~\cite{ying2019bench}, which reduces the search cost by 18$\times$ than GBDT-NAS~\cite{luo2020neural}, 8$\times$ than SemiNAS~\cite{luo2020semisupervised} and BONAS~\cite{shi2020bridging}. % RE~\cite{real2019regularized} and RL~\cite{zoph2016neural}. %, BOHB~\cite{falkner2018bohb}, SemiNAS~\cite{luo2020semisupervised} and BONAS~\cite{shi2020bridging}.
On NAS-Bench-201~\cite{dong2020nasbench201}, we achieve an improvement of up to 3.67\% in accuracy under similar cost. On ProxylessNAS~\cite{cai2019proxylessnas}, \algname{} % is twice faster than Neural Predictor~\cite{wen2020neural}, and 
achieves SOTA under mobile settings (ImageNet top-1 75.13\%, 84ms). Remarkably, {\algname{}} surpasses two GCN-based accuracy predictors on all the benchmarks with even smaller costs.

To sum up, our main contributions are listed as follows:

\begin{itemize}
\item We advocate NDCG as a new metric of performance rankers in NAS, and prove its effectivess both theoretically and empirically. % relax the objective from equally considering all architectures in the search space to focusing on those best-performing architectures, and propose a novel algorithm named \algname{}.
\item We propose a novel algorithm, \algname{}, which directly optimizes NDCG, and leverages weight sharing as weak labels to accelerate the search process. % To the best of our knowledge, we are the first to leverage weight sharing as weak labels and combine them with ground-truth labels. \yuge{looks weird} % , and combines sampling-based and weight-sharing approaches to get the ranking model.
% \item To the best of our knowledge, we are the first work to leverage weight sharing as weak labels and combine them with ground-truth labels. % transfer implicit knowledge from weight sharing by utilizing the weak labels produced by the super-net.
\item We comprehensively evaluate and demonstrate superiority of  our approach  over state-of-the-art methods on various search spaces and datasets. % The results demonstrate the superiority of our approach on performance and efficiency over SOTA NAS methods.
We will open-source the whole code base to facilitate future NAS research.
\end{itemize}

\section{Related works}

\highlight{Performance rankers in NAS.}
% What are performance rankers and efforts to combine different methods. 
% As \cite{elsken2018neural} pointed out, the three key components of NAS are search space, search strategy and performance estimation strategy.
% This paper focuses on performance estimation strategy. The ideal approach is to fully train and evaluate each sampled architecture. However, they come at a large computational cost and are impractical. 
We focus on performance estimation strategy in NAS. In this paper, we call them \emph{performance rankers}, because as pointed out by  \cite{elsken2018neural}, for most of the times only relative rankings matter.
One popular method is called weight sharing~\cite{pham2018efficient,liu2018darts,wu2019fbnet,cai2019proxylessnas}. It creates a super-net which contains all the architectures in the search space, such that every possible architecture is a sub-net. Then sub-net's accuracy is used to estimate its accuracy when trained alone. % Thus, the super-net is trained and sub-nets' accuracies are used to  weights of sub-nets are shared so as to save the cost to train each architecture from scratch.
Although it can save the cost to train each architecture from scratch, its effectiveness is still a question under debate~\cite{yu2019evaluating,li2019random,zhang2020deeper,bender2020can}.
Recently, another method called ``performance predictor''~\cite{baker2017accelerating,dai2018chamnet,chau2020brp,wen2020neural} is gaining popularity. It collects data-points of architecture-accuracy pairs, and trains a ranking model in a supervised learning manner. Although it outperforms weight-sharing on many benchmarks, % This is renowned as ``performance predictor''~\cite{baker2017accelerating,dai2018chamnet,chau2020brp,wen2020neural}. %,ning2020generic}.
the cost of collecting enough data-points remains high. % Another method is called weight sharing~\cite{pham2018efficient,liu2018darts,wu2019fbnet,cai2019proxylessnas}. It combines all the architectures in the search space into a super-net, such that every possible architecture is a sub-net. Thus, the weights of sub-nets are shared so as to save the cost to train each architecture  from scratch. Despite its efficiency, its effectiveness is still a question under debate~\cite{yu2019evaluating,li2019random,zhang2020deeper,bender2020can}. %~\cite{yu2019evaluating,li2019random,yang2019nas,zhang2020deeper,adam2019understanding,singh2019study,bender2020can}.  % Zhang \etal~\cite{zhang2020does} conducts extensive experiments on 5 search spaces to conclude that weight sharing is useful in distinguishing relatively good architectures from bad, but fails to identify the top ones. This inspires us to treat weight sharing accuracy as weakly supervised labels and transfer the knowledge from weight sharing to our performance predictor. Most recent works have shown the potentials to leverage a cheap metric (\eg FLOPs~\cite{dai2020fbnetv3} and latency~\cite{chau2020brp}) to improve the accuracy predictor. These techniques are also shown to be helpful to us.
Aiming to take the best of both worlds, we combine these two approaches in our design. % when designing our algorithm. % in this paper, to take the best of both worlds.
% There has already been the philosophy of ``hybrid'' rankers in NAS community~\cite{white2021powerful,zhao2021few} , but they have fundamentally different motivations and methodologies. % details are totally different from us.

% Early NAS methods like Reinforcement Learning (RL)~\cite{zoph2016neural}, Evolutionary Algorithms (EA)~\cite{liu2018progressive} or Bayesian Optimization (BO)~\cite{kandasamy2018neural} come with a huge cost to train many architectures.
% Therefore, a good ranker of neural architecture performance becomes the key component to help quickly filter out the bad-performing architectures and identify the best. The most popular approach to build a ranker is based on ``sampling'', which are also known as \emph{performance predictors}.

\highlight{Assessment metrics for performance rankers.}
% In the literature, the evaluation of performance rankers usually includes two parts. Firstly, it is combined with a particular search strategy and search space. 
The ideal and intuitive evaluation of  performance rankers is to integrate it with a search strategy and conduct a search on a particular space. However, as argued by \cite{yu2019evaluating}, such evaluation is coupled with many factors such as sampling strategy, making fair comparison under controlled settings difficult. Moreover, since only the best model of the search process is reported, the ranker's robustness is not reflected in the evaluation. %~\cite{yu2019evaluating}. % it is too random because only the best result in the whole search process is reported.
% To demonstrate that the performance gain comes from the performance ranker itself,
On the other hand, literature often adopts ranking correlations (\eg KdT) that shows the consistency between the predicted rank and ground-truth rank~\cite{yu2019evaluating,pourchot2020share,zhang2020deeper,yu2020train,white2021powerful}. The wide usage of ranking correlations has even catalyzed recent works~\cite{xu2021renasrelativistic,chau2020brp} that directly minimizes ranking errors.
However, correlation metrics neglect the fact that the goal of NAS is to find top-performing architectures in the search space. % architectures are  in NAS.
% Therefore, recent works~\cite{xu2021renasrelativistic,chau2020brp} that optimizes a global pairwise ranking errors are setting an over-strict goal.
Treating all the architectures equally is not well aligned with such goal, resulting in less efficient algorithms.
Therefore, we advocate NDCG that takes the full ranking list into consideration but also emphasizes top-performing architectures.

\highlight{Learning to Rank and Information Retrieval.}
Learning to rank (LTR)~\cite{liu2011learning} uses machine learning technologies to build effective ranking models, and is widely used in solving document ranking problem in Information Retrieval (IR). We come to notice LTR because NAS is also trying to solve a ranking problem. Over the past decades, many LTR methods have been proposed and deployed in modern IR systems like search engines. They can be categorized into pointwise~\cite{crammer2001pranking, cooper1992probabilistic}, pairwise~\cite{burges2005learning, tsai2007frank} and listwise~\cite{burges2007learning}. Our method relies on LambdaRank, which falls into the category of listwise approaches. % The past decades have witnessed a trend that LTR algorithms evolved from pointwise~\cite{crammer2001pranking, cooper1992probabilistic}, to pairwise~\cite{burges2005learning, tsai2007frank}, to  listwise~\cite{burges2007learning}. 

\section{NDCG: a new metric for rankers in NAS} \label{sec:ndcg}

\subsection{Problem formulation}
\label{sec:problem-formulation}

The NAS problem can be formulated as,
\begin{equation}
\label{eq:nas-formulation}
\alpha^{*} = \argmax_{\alpha \in \mathcal{A}} \textsc{Acc}(\alpha)
\end{equation}
which is finding the architecture from a search space $\mathcal{A}$, $\alpha \in \mathcal{A}$, such that $\textsc{Acc}(\alpha)$ (\ie \emph{ground-truth accuracy}\footnote{We will use ``accuracy'' for short throughout this paper. Note that test accuracy can be easily replaced with other metrics. We use the term ``ground-truth accuracy'' to denote the test accuracy when the model is fully trained.} of $\alpha$) is maximized. Since obtaining the precise $\textsc{Acc}(\alpha)$ requires fully training a neural network, which is costly, previous works often design a performance ranker $f$, and use $f(\alpha)$ as a proxy of $\textsc{Acc}(\alpha)$.%  Ideally, the ranker should be good enough so that computing the argmax of $f$ is equivalent to argmax of $\textrm{ACC}$. % Examples of performance rankers include accuracy predictor in predictor-based NAS~\cite{wen2020neural}, well-trained super-net in weight sharing NAS~\cite{guo2020single} and performance estimation based on untrained neural networks~\cite{mellor2020neural}.

Focusing on performance rankers, we first introduce the notations we will use throughout the rest of this paper. Let $\mathbf{a} = \{\alpha_1,\alpha_2,\ldots,\alpha_n\}$ be the architectures to rank, whose ground-truth accuracy is $\mathbf{acc} = \{\acc_1, \acc_2, \ldots, \acc_n \}$ respectively. The performance ranker is $f: \mathcal{A} \rightarrow {\mathbb{R}}$, \ie a ranking function $f$ in function space $\mathcal{F}$. $\pi_f (i)$ ($1 \le i \le n$) is the ranked list produced by $f$, such that $f(\alpha_{\pi_f(1)}) \ge f(\alpha_{\pi_{f}(2)}) \ge \cdots \ge f(\alpha_{\pi_{f}(n)})$.

With the formulations above, \cref{eq:nas-formulation} can be written as the following bi-level optimization:
\begin{equation}
\begin{split}
\alpha^{*} &= \argmax_{\alpha \in \mathcal{A}} f^{*}(\alpha) \\
\textrm{s.t.} ~~~  f^{*} &= \argmax_{f \in \mathcal{F}} \textsc{Objective} (f)
\end{split}
\end{equation}
One of the challenges in this optimization is to find a good $\textsc{Objective}(\cdot)$ to guide the optimization of $f$. We identify two essential properties of rankers that a good objective should encourage. \emph{(i)} $\argmax f^* (\alpha)$ and $\argmax \textsc{Acc} (\alpha)$ should be as close as possible, which means the ranker should precisely predict the top-tier architectures' ranking. \emph{(ii)} As $\mathcal{A}$ is often large and solving $\argmax f^{*} (\alpha)$ requires optimization techniques, it should also give reasonable rankings to worse architectures to hint the ``argmax optimizer'' to find global maximum. %, so that heuristics will be able to distinguish them from better ones.
Existing widely used objectives (\eg minimizing pairwise ranking errors)~\cite{xu2021renasrelativistic} often satisfy property (ii) but neglect property (i).

Property (i) is critical for identifying the best among good performing architectures. To possess this property, we resort to NDCG, a renowned ranking metric in Information Retrieval, as the objective, which satisfies the two properties mentioned above simultaneously. This design choice is inspired by the observation that NAS has similar optimization goal to IR. In IR, when retrieving the best matched documents from a large number of documents, the ranking quality of high relevant documents is more important than that of low relevant documents. Similarly in NAS, model developers care more about identifying the top architecture from those relatively good-performing models, while distinguishing which one is worse among bad-performing models is less important.

\subsection{Normalized Discounted Cumulative Gain}
\label{sec:adapt-ndcg-to-nas}

\emph{Normalized Discounted Cumulative Gain (NDCG)}~\cite{jarvelin2002cumulated},  which has been proved effective and has been widely adopted in IR~\cite{liu2011learning},
is a metric of ranking quality and is often used to measure effectiveness of a ranker. It properly shifts the focus of the ranking towards the front part of the rank. It takes into account the graded relevance values and encourages the highly relevant items to come up into the top of recommended lists. 
%In applications of IR, NDCG is a popularly used metric, and its effectiveness has been well studied~\cite{10.1145/1645953.1646032,wang2013theoretical}.
% In applications of IR, NDCG has shown its effectiveness in improving top-ranked results and has been well studied in quite a few previous works, both theoretically and empirically~\cite{10.1145/1645953.1646032,wang2013theoretical}.
In the context of NAS, with the notation given in \cref{sec:problem-formulation}, NDCG can be computed as,
\begin{equation}
\label{eq:ndcg}
\NDCGExpr = \frac{\sum_{i=1}^n \frac{1}{\log_2(i+1)} \left( 2^{r_{\pif(i)}} - 1 \right)}{\sum_{i=1}^n \frac{1}{\log_2(i+1)} \left( 2^{r_{i}} - 1 \right)}
\end{equation}
where the numerator is often written as $\textrm{DCG} (f; \mathbf{a}, \mathbf{r})$ and the denominator is known as IDCG (Ideal DCG) because it is the DCG with $f$ as a perfect ranker. $\mathbf{r} = \{r_1,r_2,\ldots,r_n\}$ ($0 \le r_i \le S$) are relevance scores which are proportional to accuracy.
Hence, NDCG is the normalized DCG in $[0,1]$. If a model with high accuracy gets ranked poorly, DCG gets penalized. The ``$2^{r_i}$'' part emphasizes on models with higher accuracy, thus encourages the ranker to retrieve more of them, rather than focusing equally on the whole rank.

The architectures' accuracy should be properly mapped to the relevance score. Specially in NAS, the distribution of accuracy is usually highly skewed, \ie the values span in the whole range while most of them gather in a much smaller range. Simply normalizing the whole distribution to the range (\eg 0-10) used in IR will not work. For instance, when dealing with a long-tail distribution with most architectures' accuracy above 90\% and a few additional low-accuracy outliers, using min-max normalization vanishes the ability of distinguishing the accuracy of most architectures. Therefore, we compute new lower and upper bound to clip the distribution. Due to the exponential effect of ``$2^{r_i}$'', the distribution should be scaled to a proper range (\ie relevance scale $S$) to maximize the effectiveness of identifying top architectures. The empirical study of the clipping and scaling is detailed in \cref{sec:appendix-ndcg-details}.

\subsection{Analysis of NDCG in the context of NAS}

\cref{eq:ndcg} implies that the ranking focus is skewed towards the front part of the rank, which properly balances the ability of discriminating good-performing architectures and the ability of distinguishing good architectures from bad ones. As a metric (\eg NDCG) of a performance ranker should eventually serve the final goal of NAS, it is necessary to prove the alignment between the metric and the final goal. In NAS, \textit{top-$k$ accuracy} is a widely recognized metric to reflect how well the final goal is achieved~\cite{wen2020neural,bender2018understanding,mellor2020neural}. Specifically, \textit{top-$k$ accuracy} is computed by evaluating the ground-truth accuracy of top $k$ architectures indicated by the ranker and pick out the best one from them.
%\yuge{To show that NDCG is highly correlated with the final goal of NAS, we follow \cite{wen2020neural,bender2018understanding,mellor2020neural} to apply the ranker to a scenario (NAS pipeline?) where the ranker is firstly trained and used to predict a large number of sampled architectures. The top-$k$ predicted architectures are selected to be fully trained. Our study in this section are two parts. We first show the relationship between NDCG and \emph{top-$k$ accuracy} theoretically, and then we use empirical study to illustrate the property and effectiveness of NDCG.}
Below, we prove the correlation between NDCG and
\textit{top-$k$ accuracy} theoretically, then use empirical study to illustrate the property and effectiveness of NDCG.

% \yuge{To this end, we consider a common scenario~\cite{wen2020neural,bender2018understanding,mellor2020neural} where a ranker is used: evaluating the ground-truth accuracy of top $k$ architectures indicated by the ranker and pick out the best one. ...}

%There are subtle differences of NDCG used in IR and that adapted to NAS: \emph{(i)} Accuracies belong to a continuous space, while relevance scores in IR are usually discrete; \emph{(ii)} In NAS, we want to maximize the best accuracy, while the majority of IR works are maximizing NDCG itself~\cite{liu2011learning}. In this subsection, we show that NDCG of a ranker is highly correlated to the final performance (\ie top-$k$) of NAS experiment. % a good metric, which is able to well reflect a ranker's performance in a NAS experiment.

\subsubsection{Theoretical analysis}

\begin{definition}
Consider a scenario where the ranker $f$ selects $k$ ($1 \le k \le n$) models with highest prediction scores from $\mathbf{a} = \{ \alpha_1, \alpha_2, \ldots, \alpha_n \}$ and fully trains them. With the budget $k$, the final NAS performance is defined as,
\begin{equation*}
\TopKExpr = \max_{1 \le i \le k} r_{\pi_f(i)}.
\end{equation*}
Here we used relevance scores instead of accuracy. As they are proportional, the conclusions are similar.
\end{definition}

%We then show that NDCG and TopK are highly correlated to each other, with \cref{theorem:1} and \cref{theorem:2}. Proofs and corollaries can be found in \cref{sec:proof}. 
\begin{theorem}
\label{theorem:1}
If $\NDCGExpr \ge \nu$ and $k > (n+1) - (n+1)^{\nu}$,  $\TopKExpr \ge \nu \left( \frac{\sum_{i=1}^{n} r_i c_i }{\sum_{i=1}^n c_i } \right)$, where,
\begin{equation*}
\cnu{i} = \begin{cases}
\invlog{i} - \frac{1}{\nu \log_2(i+k+1)} &\text{ if $1 \le i \le n -k$} \\
\invlog{i} &\text{ if $n-k < i \le n$} 
\end{cases}
\end{equation*}
\end{theorem}

In \cref{theorem:1}, $h(\nu) = \nu \left( \frac{\sum_{i=1}^{n} r_i c_i }{\sum_{i=1}^n c_i } \right)$ is an increasing function of $\nu$ (proved in \cref{sec:proof}). Thus, TopK is lower bounded by a function that is monotonically increasing to NDCG.

\begin{theorem}
\label{theorem:2}
Assume two rankers $f_1$ and $f_2$ are randomly drawn from $\mathcal{F}$, $n \gg K$. Let $\TopK (f_1; \mathbf{a}, \mathbf{r}, k) \ge r_{t_1}$, $\TopK (f_2; \mathbf{a}, \mathbf{r}, k) \ge r_{t_2}$. If $r_{t_1} > r_{t_2}$ and $r_{t_1}, r_{t_2} \ge \log_2 \left( \frac{1}{n} \relexpsum{1}{n} \right)$, $\E [\NDCG(f_1; \mathbf{a}, \mathbf{r})] > \E [\NDCG(f_2; \mathbf{a}, \mathbf{r})]$ holds.
\end{theorem}

Further, \cref{theorem:2} proves that a ranker with better TopK is expected to have a higher NDCG. The full proof is detailed in \cref{sec:proof}.

\subsubsection{Empirical analysis}

\begin{figure}[t]
    \centering
    % \vspace{-0.5em}
    \begin{subfigure}{0.48\linewidth}
        \centering
        \includegraphics[width=\textwidth]{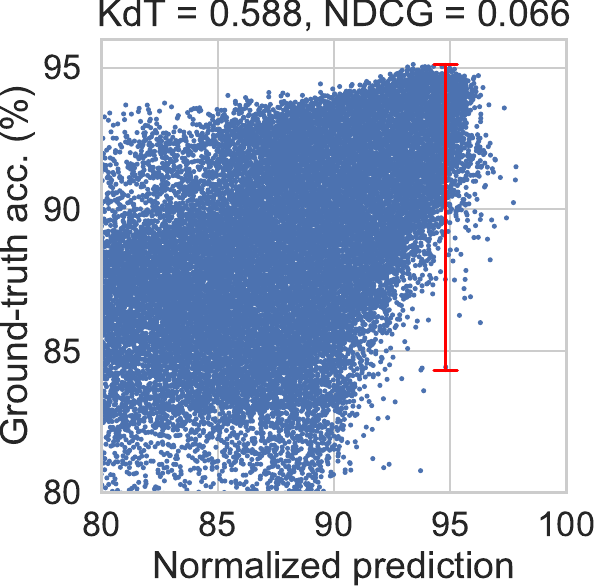}
        % \subcaption{Better k-tau, top-1 = 90.98}
    \end{subfigure}
    \begin{subfigure}{0.48\linewidth}
        \centering
        \includegraphics[width=\textwidth]{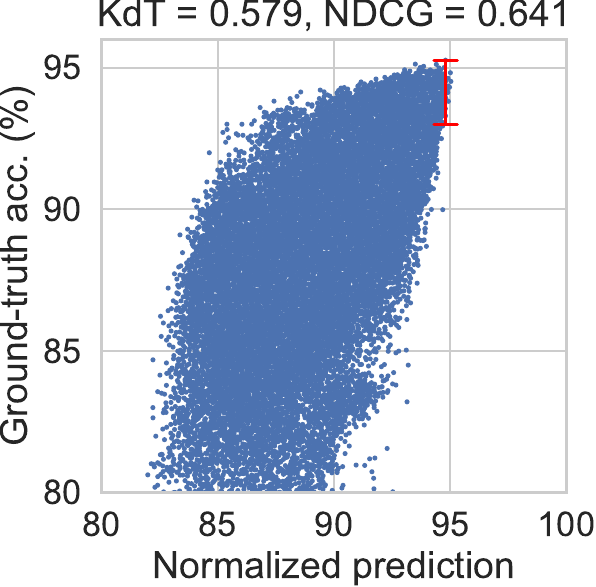}
        % \subcaption{Better NDCG, top-1 = 94.51}
    \end{subfigure}
    % \caption{The prediction ground-truth scatter plot of two predictors. Although two figures have a close Kendall's tau, in the left figure, the top architectures look more scattered, resulting in the failure of finding the best architectures.}
    \caption{The prediction -- ground-truth scatter plot of two rankers. The two rankers have identical network structures. (\textbf{Left}) Ranker optimized for KdT and the top-1 accuracy based on it is 90.98\%. The red line marks the region where the ranker gives high scores but its ground-truth accuracy are low. (\textbf{Right}) Ranker is optimized for NDCG and the top-1 accuracy based on it is 94.51\%.}
    % \vspace{-1em}
    \label{fig:ndcg-kdt-example}
\end{figure}

One experiment is to illustrate the correlation between prediction accuracy and ground-truth accuracy in \cref{fig:ndcg-kdt-example}, where the prediction accuracy is produced by optimizing KdT and NDCG respectively (see next section).
The two accuracy rankers have identical structures~\cite{wen2020neural}. When the ranker is trained to maximize KdT (\ie full-ranking correlation~\cite{chau2020brp,xu2021renasrelativistic}), the final KdT of the prediction accuracy is 0.588 as shown in the left figure. However, the architectures with 95\% predicted accuracy have varied ground-truth accuracy between 85\% and 95\%. In contrast, the ranker optimized for NDCG has a much sharp head (\ie top right of the right figure), the architectures with 95\% predicted accuracy are precisely located in a much smaller range (\ie 93\%-95\%) of the ground-truth accuracy. Besides, this ranker produces equally good KdT (\ie 0.579), which means it is also qualified to distinguish good architectures from bad ones.
%Although two figures have a close Kendall's tau (0.588 vs. 0.579), the NAS performance (top-1 accuracy) varies a lot (90.98\% vs. 94.51\%).
%We highlight the region with red line, where ranker gives high prediction scores (above 95\%) but the ground-truth accuracy is relatively low (spans widely from 85\% to 95\%). As Kendall's tau pays equal attention to the whole space, failure to recall top architectures is not properly penalized. Therefore, Kendall's tau can not reflect a ranker's real quality in NAS.
%As illustrated in \cref{fig:kendalltau-vs-ndcg-1}, Kendall's tau might not be suitable for NAS either. 

% We collect 12 NAS performance rankers, spanning from sampling-based performance predictors to weight-sharing based methods (implementation detailed in \cref{sec:impl-details}). For each method, we train a ranker and measure its NDCG and top-10 test accuracy on each search space and dataset, respectively. Results on more search spaces are available in \cref{sec:correlation-ndcg-accuracy}. The scatter plot is shown in \cref{fig:ndcg-accuracy-12-ranker}. The upward trend in this figure indicates a positive correlation between NDCG and test accuracy. Notably, for rankers achieving test accuracy between 70\% and 75\%, NDCG spans a wide range from 0.6 and 0.9, indicating that NDCG is numerically sensitive to top-performing architectures.

To further demonstrate the effectiveness of NDCG, we collect rankers from 12 search spaces (listed in \cref{sec:experiments}) and pairwisely compare them. Suppose we have $m$ search spaces, and for the $i$-th search space, we get $d_i$ rankers $f_{i,1}, f_{i,2}, \ldots, f_{i,d_i}$. In total, we get $\sum_{i=1}^m \frac{d_i (d_i - 1)}{2}$ ranker pairs. In this experiment, $m=12$ and we get 486 ranker pairs in total. For each pair $f_{i,j}, f_{i,k}$, we determine whether the ranker with higher NDCG also produces better top-$k$ test accuracy. If it does, we count this pair as a ``successful'' distinguished ranker pair for NDCG. Then for each $k$, we compute ``success rate'', which is number of successful distinguished pairs divided by total number. The success rate is similarly computed for ranking correlations including Kendall's tau and Spearman's rho. We also include top-1, \ie the accuracy of architecture with the highest prediction score, into comparison.

\begin{figure}[t]
\centering
\includegraphics[width=\linewidth]{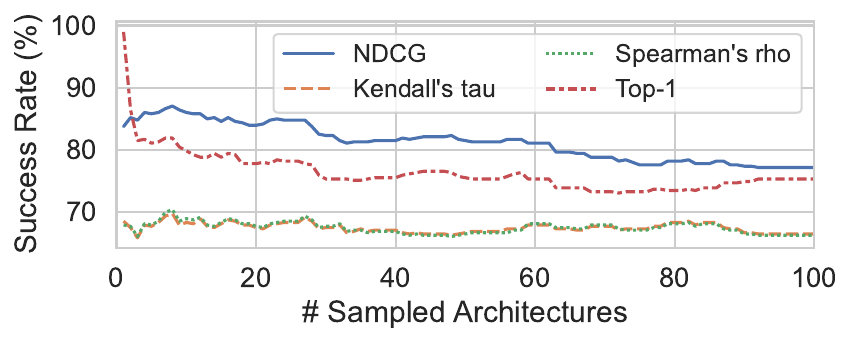}
\caption{Comparison of the success rate with four different ranking metrics. $k$ sampled architectures use top-$k$ to evaluate success rate.}
\label{fig:metric-reliability-success-rate}
% \label{fig:metric-reliability-main-text}
\end{figure}

% \begin{figure}[t]
% \centering
%     \begin{subfigure}{.6\linewidth}
%         \centering
%         \includegraphics[width=\textwidth]{figures/metric_reliability_scatterplot_nasbench201.pdf}
%         \subcaption{NDCG and accuracy of 12 rankers.}
%         \label{fig:ndcg-accuracy-12-ranker}
%     \end{subfigure}
%     \begin{subfigure}{.38\linewidth}
%         \centering
%         \includegraphics[width=\textwidth]{figures/metric_reliability_success_rate_v2.pdf}
%         \subcaption{Success rate.}
%         \label{fig:metric-reliability-success-rate}
%     \end{subfigure}
% \caption{(\textbf{left}): We select 12 performance rankers and compute their NDCG and final test performance. (\textbf{right}): Success rate of NDCG in distinguish a good ranker and a pool ranker, as compared to other ranking metrics.}
% \label{fig:metric-reliability-main-text}
% \end{figure}

%We also compute the success rate for different number of sampled architectures, and when using other metrics such as Kendall's tau.
Results are shown in \cref{fig:metric-reliability-success-rate}. There are two findings: \emph{(i)} NDCG can distinguish a good ranker from a poor one with more than 80\% probability, which is much higher than correlations like KdT. \emph{(ii)} The success rate of NDCG even surpasses top-1 under various budgets. This is because top-1 only takes the highest prediction into account and it might not be robust enough to comprehensively assess a ranker's quality, which echoes the argument in \cite{yu2019evaluating}.
% (ii) NDCG can be especially useful when number of sampled architectures is small, while such pattern is not clearly observed for other metrics. This is because NDCG emphasizes the first items of resulting ranked list with $\frac{1}{\log_2(i+1)}$ term.

More empirical analysis are in \cref{sec:correlation-ndcg-accuracy}.

% \vspace{-0.5em}

% \paragraph{NDCG vs. rank correlation.}

%As we have demonstrated above, NDCG matches the goal of NAS. 
% We use a simple experiment to compare NDCG and Kendall's tau in~\cref{fig:ndcg-kdt-example}. In the experiment, we train a vanilla accuracy predictor~\cite{wen2020neural} using two different training configurations. The left figure shows higher Kendall's tau, but its ability of identifying top architectures is weaker. The accuracy of the architectures whose predicted accuracy are around 95\% spans from 85\% to 95\%, and its NDCG is very low. In contrast, the right figure has a much higher NDCG. Accordingly, top architectures are more accurately identified. Therefore, NDCG is a better optimization objective and metric for NAS than rank correlation. More concrete experiments are available in the next section.

%as metric to indicate the ability of identifying top-$k$ architectures using a simple experiment in~\cref{fig:ndcg-kdt-example}. We train a vanilla accuracy predictor~\cite{wen2020neural} by using different training configurations. As shown in \fixme{the right figure}, the shape is sharper on the top right, which indicates that NDCG has higher discrimination on top architectures, though its Kendall's tau is lower.

\section{\algname{}: Learning to Rank Ace Architectures}
\label{sec:method}

Once we set up NDCG as a new metric to evaluate NAS performance rankers, a natural follow-up idea is to directly optimize NDCG of performance rankers. By incorporating techniques from information retrieval, we show that direct optimization of NDCG is feasible with LambdaRank. Meanwhile, to make our algorithm more practical, we pre-train the ranker with weak labels obtained from a well-trained super-net to greatly reduce architecture-accuracy pairs needed. % Meanwhile, as it is hard and expensive to collect sufficient architecture-accuracy pairs, we pre-train the ranker with weak labels obtained from a well-trained super-net to greatly reduce architecture-accuracy pairs needed.

\begin{figure}[t]
    \centering
    % \vspace{-2em}
    \includegraphics[width=\linewidth,trim=1.5cm 0cm 1.5cm 1cm,clip]{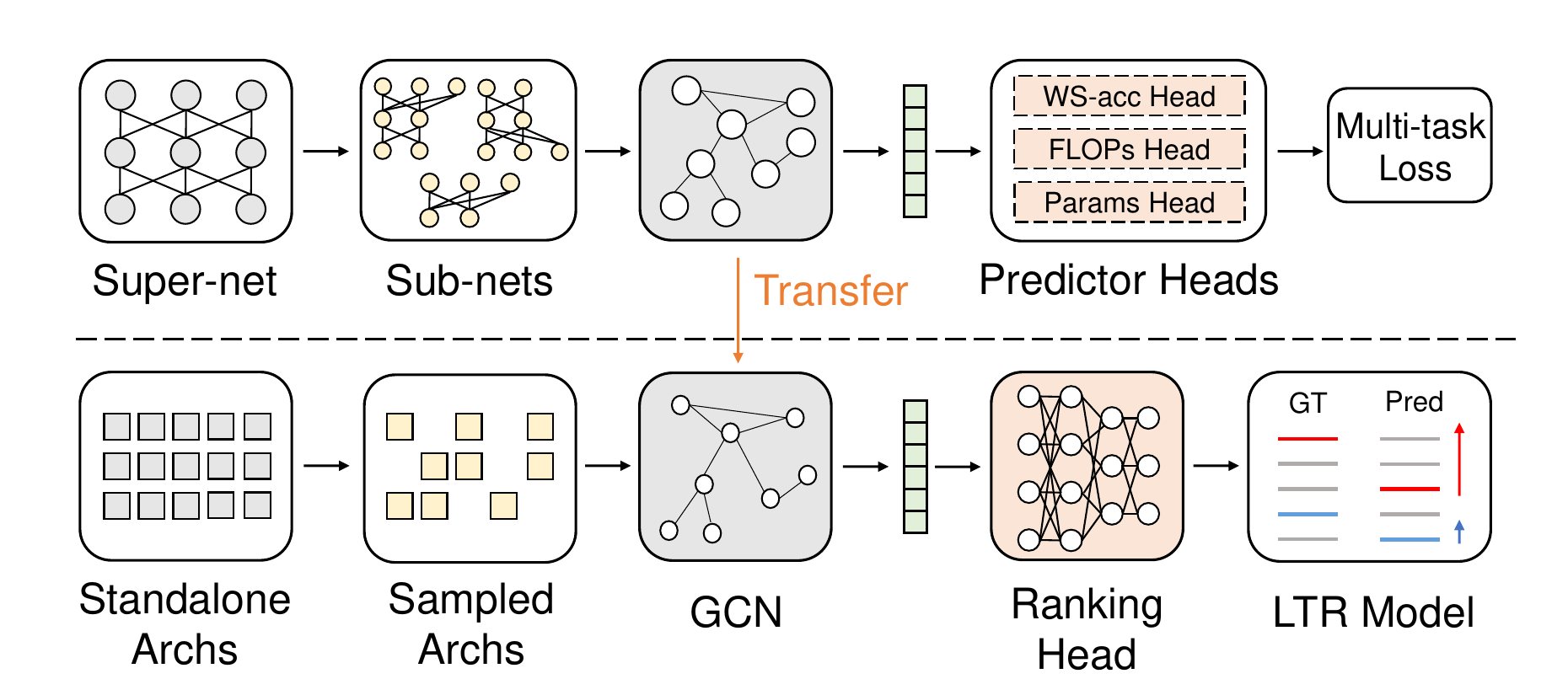}
    % \vspace{-2em}
    \caption{(\textbf{Top}) AceNAS first samples weak labels on super-net to train the ranking model with multi-task loss. (\textbf{Bottom}) The trained GCN is transferred to optimize the LTR model.}
    \label{fig:method}
    % \vspace{-1em}
\end{figure}

The overall illustration of \algname{} is shown in~\cref{fig:method}, which is divided into two stages. First, pre-training a GCN-based ranker with weakly supervised labels from a well-trained super-net. Second, transferring the pre-trained GCN into another ranker and fine-tuning it with LambdaRank with limited architecture-accuracy pairs.

% We design \algname{} which incorporates techniques from information retrieval and combines new NAS-specific designs and adaptations. In this section, we first formulate NAS as a Learning to Rank problem and justify NDCG as a new optimization objective for NAS. Then we design a new NAS ranking model based on LambdaRank, in which we use GCN to capture the representation of neural architectures. As it is hard and expensive to collect sufficient architecture-accuracy pairs, we pretrain the GCN with weak labels obtained from a well-trained super-net to greatly reduce architecture-accuracy pairs needed.

\subsection{LambdaRank}\label{sec:rankingmodel}

We follow \cite{wen2020neural,chau2020brp} to design a ranker with GCN~\cite{kipf2016semi}, that takes a neural network architecture as input and predicts its score. It first encodes neural architectures into an encoding with a few graph convolutional layers, then use a ranking head (\ie a Multi-layer perceptron) to output the prediction. Detailed designs can be found in \cref{sec:impl-details-gcn}. % It then fully trains many architectures, gets their accuracy, and builds a ``training dataset'' consisting of many architecture-accuracy pairs.
To train the ranker, we follow the state-of-the-art paradigm -- performance predictors~\cite{wen2020neural} -- to collect a number of data-points, \ie pairs of architecture and accuracy. Next, we describe how the optimization is done.
% The ranker is made up of a few graph convolutional layers, followed by a ranking head, which is a Multi-layer perceptron. Detailed designs can be found in \cref{sec:impl-details-gcn}. Here, we focus on how to optimize this ranker given a dataset.

Suppose we have a ranker $f_{\omega}$ parameterized with $\omega$ and a set of known architecture-accuracy pairs $\{(\alpha_1, \acc_1) , \ldots, (\alpha_m, \acc_m) \}$. Let $\{r_1, r_2, \ldots, r_m\}$ be their corresponding relevance (see \cref{sec:adapt-ndcg-to-nas}). The optimization goal can be formally written as,
\begin{equation}
\label{eq:lambdarank-optimization-goal}
\omega^{*} = \argmax_{\omega} \NDCG(f_\omega; \{\alpha_1, \ldots, \alpha_m\}, \{r_1, \ldots, r_m\})
\end{equation}
% We note that the optimization of \cref{eq:lambdarank-optimization-goal} is challenging because NDCG has a non-differentiable sorting operator. As a result, $\omega$ can not be optimized with vanilla back-propagation.

LambdaRank~\cite{burges2007learning}, a well reputed LTR algorithm, is capable of optimizing  \cref{eq:lambdarank-optimization-goal}.
% Fortunately, direct optimization of NDCG can be solved with LambdaRank~\cite{burges2007learning}, a well known LTR algorithm. 
At its core part, gradients can be directly computed with LambdaRank without actually computing the loss explicitly.
Specifically, for a pair $(\alpha_i,\alpha_j)$ whose ranking score is $(f_{\omega}(\alpha_i), f_{\omega}(\alpha_j))$ and $r_i > r_j$, the gradient of ranking model parameter $\omega$ is computed as,
% ... What is the relationship to RankNet?
% LambdaRank~\cite{burges2007learning} optimizes NDCG by directly computing the gradients (instead of relying on the loss functions). It is inspired by RankNet and the gradients computed for a pair $(\alpha_i,\alpha_j)$ has the following form:
\begin{align}
\label{eq:lambdarank-1}
    \frac{\delta \Loss}{\delta \omega} & = \lambda_{ij} \left( \frac{\delta f_{\omega}(\alpha_i)}{\delta \omega} - \frac{\delta f_{\omega}(\alpha_j)}{\delta \omega} \right) \\
% \end{equation}
% \noindent
% where $\lambda_{ij}$ is defined as:
% \begin{equation}
\label{eq:lambdarank-2}
    \lambda_{ij} & \equiv \frac{- \sigma}{1 + e ^{\sigma(f(\alpha_i) - f(\alpha_j)))}} \lvert \Delta_{\mathrm{NDCG}} \rvert
\end{align}
% \noindent
$\sigma$ is the hyper-parameter controlling the shape of sigmoid. $\lvert \Delta_{\mathrm{NDCG}} \rvert$ measures the change of NDCG if the ground-truth ranking position of $\alpha_i$ and $\alpha_j$ get swapped. Swapping higher ranked items gets more penalty, leading to a larger gradient. Note that if $\lvert \Delta_{\mathrm{NDCG}} \rvert$ is replaced with $1$, \cref{eq:lambdarank-1} reduces to RankNet~\cite{burges2005learning}, which minimizes the number of inversions in the full ranking. % Different from the ranking models that optimize the whole rank (\eg pair-wise ranking loss in RankNet~\cite{burges2005learning}), LambdaRank uses NDCG to put higher emphasis on good-performing architectures. It takes the position of an item (\eg a document in IR or an architecture in NAS) in the ranking distribution into consideration. In

\subsection{Weak supervision of weight sharing} \label{sec:ws}

% As it is computationally expensive to obtain sufficient number of architecture-accuracy pairs in ground-truth, the training of ranking model becomes particularly challenging and unstable.

% The approach to optimize NDCG as described above requires sufficient number of architecture-accuracy pairs, which can be still computationally expensive when the search space or dataset becomes large in scale. 
% To make the algorithm more practical, 
To reduce the required number of architecture-accuracy pairs and make the algorithms more practical, we propose to use weak supervision from a well-trained super-net (\ie accuracy evaluated using the weights from super-net) to pre-train the ranker. A super-net is trained with weight sharing approach (single-path random sampling~\cite{li2019random,guo2020single} in our experiments), % using uniform sampling, \ie each mini-batch trains a sampled architecture in the super-net~\cite{guo2020single},
and thus the computation cost is comparable to training a single architecture. The design of treating weight sharing accuracy as weakly supervised labels is inspired by the observation in previous research~\cite{zhang2020does} that weight sharing super-net is capable of differentiating good architectures from bad ones, with relatively high ranking correlation.% (\eg Kendall's tau could be higher than 0.6 on many search spaces).

% To empower our ranking model with knowledge from super-net, we replace ranking head in the model (\cref{fig:predictor}) with a \emph{WS-accuracy head}, which is another two-layer MLP to predict weight sharing accuracy. This model is trained with mean-squared-error (MSE) loss instead of using LambdaRank, because weight sharing labels are not qualified for identifying the best architectures from good-performing ones.

To empower our ranker with knowledge from super-net, we replace ranking head in the ranker with a \emph{WS-accuracy head} in the pre-training. The ranker is trained with mean-squared-error (MSE) loss instead of using LambdaRank, because weight sharing labels are not qualified for identifying the best architectures from good-performing ones. To further boost the effectiveness of pre-training, inspired by \cite{dai2020fbnetv3,chau2020brp}, we incorporate multi-task training by introducing additional two heads that predict FLOPs and number of parameters respectively. % . Specifically, Apart from WS-accuracy head, additional two heads are introduced to predict FLOPs and number of parameters, respectively.
The ranker is trained to minimize the following multi-task mean-squared-error (MSE) loss:
%The output of GCN is feeded into multiple MLPs (Multi-layer Perceptron),In the first stage, three of them are used, to predict weight-sharing accuracy, FLOPs and number of parameters respectively.
%Weight sharing is first introduced to reduce budgets since evaluate thousands of sub-nets on super-net is almost free compared to training. Although the ranking power of weight sharing based NAS is still controversial, ~\cite{zhang2020does} shows that architecture rank derived from super-net can falls into 
%a relatively stable state. Such that, it would be helpful to transfer the ranking knowledge learned in the super-net to our ranking model.
% \vspace{-.5em}
\begin{align}
% \label{eq:multitask}
% \begin{split}
\Loss_{\textrm{pretrain}} = & \Loss_{\textrm{mse}}(\acc_i, \acc_i^*) + \lambda_1 \cdot \Loss_{\textrm{mse}}(\flops_i, \flops_i^*) \nonumber \\
& ~~~~~ + \lambda_2 \cdot \Loss_{\textrm{mse}}(\params_i, \params_i^*) \label{eq:multitask}
% \end{split}
\end{align}
where variables marked with stars ($^*$) are predictions. % Empirically we find that the training is not sensitive to $\lambda_1$ and $\lambda_2$, after we normalize the ground truth labels by subtracting mean and dividing by standard deviation. Therefore we simply set $\lambda_1 = \lambda_2 = 1$.

\section{Evaluation}
\label{sec:experiments}

\subsection{Experiment setup}
\label{sec:impl-details}

%In this section, we describe the experiment settings. More hyper-parameters and implementation details are available in Supplement-A.

%We now describe experiment setups. More hyper-parameters and implementation details are available in \cref{sec:appendix-impl-details}.

\highlight{Search space.} We evaluate \algname{} on both the NAS benchmarks and the widely-used ProxylessNAS~\cite{cai2019proxylessnas} search space.
\begin{itemize}
    \item \textit{NAS Benchmarks}. We collect 12 benchmarks (10 different search spaces, and 3 different datasets) from NAS-Bench-101~\cite{ying2019bench}, NAS-Bench-201~\cite{dong2020nasbench201} and NDS~\cite{radosavovic2019network}. Each benchmark provides the ground-truth accuracy for every architecture,  so that a comprehensive evaluation of our method is computationally feasible.
    \item \textit{ProxylessNAS}. We also measure \algname{} on a large chain-wise search space that consists of 21 sequential MB-Conv searchable blocks. Unlike NAS Benchmarks, the search space is much larger (i.e., $2.58 \times 10^{17}$ candidates) and there is no ground-truth accuracy for candidate architectures. To be comparable with ProxylessNAS-mobile, 
   we follow \cite{bender2020can} and use a same latency lookup table to constrain the search space within 83ms -- 85ms. % for comparable latency to ProxylessNAS-mobile (83ms -- 85ms).
   % This search space contains $2.58 \cdot 10^{17}$ architectures. ~\cite{cai2019proxylessnas} is a large chain-wise NAS search space that consists of 21 sequential MB-Conv choice blocks and $2.58 \cdot 10^{17}$ candidates. We follow \cite{bender2020can} to use a latency lookup table to constrain the search space so that models have comparable latency to ProxylessNAS-mobile (83ms -- 85ms).
\end{itemize}
%\begin{description}
 %   \item[NAS Benchmarks] contain 12 benchmarks (10 different search spaces, and 3 different datasets) provided by NAS-Bench-101~\cite{ying2019bench}, NAS-Bench-201~\cite{dong2020nasbench201} and NDS~\cite{radosavovic2019network}. An architecture-accuracy lookup table is contained in each benchmark, so that a comprehensive evaluation of our method is computationally feasible.
%    \item[ProxylessNAS] is a chain-wise search space that consists of 21 sequential MB-Conv choice blocks and $2.58 \cdot 10^{17}$ candidates, proposed by \cite{cai2019proxylessnas}. We follow \cite{bender2020can} to use a latency lookup table to constrain the search space so that models have comparable latency to ProxylessNAS-mobile (83ms -- 85ms).
%\end{description}

\highlight{Algorithm details.} Our first step is to get the three weak labels (\ie validation accuracy from weight-sharing NAS, FLOPs, and parameter size). We encode the full search space into a super-net and adopt the widely-used uniform random sampling approach~\cite{guo2020single} to train the super-net. Based on the trained super-net, we randomly sample 4k architectures and obtain weak labels. The sampled architectures and their weak labels are used for the ranker pre-training. The second step is to fine-tune the ranker with train-from-scratch accuracy labels. \algname{} spawns trials and collects their fully-trained accuracy (this can be a table lookup if a benchmark is available) to train the ranker with LambdaRank. This step is repeated a few times with the \emph{iterative sampling} strategy~\cite{chau2020brp}. Finally, we use the fine-tuned ranker to predict best-performing architectures from the search space. %We fully train them and select the model with highest validation accuracy as the final architecture.
%Finally, another few architectures are selected by ranking model, trained and evaluated. 
A formal description of the algorithm workflow is included in \cref{sec:algorithm}.

%Based on the trained super-net, 4k weak labels are evaluated, which are validation accuracies (and FLOPs, parameter size). Secondly, a ranking model is trained on those weak labels. This is called ``pretraining''. Then, \algname{} spawn trials and collect their fully-trained accuracy (this can be a table lookup if a benchmark is available) to train the LambdaRank ranking model. This step is repeated a few times with an iterative sampling strategy, as introduced in the last section. Finally, another few architectures are selected by ranking model, trained and evaluated. A formal description of the algorithm workflow is included in Supplement-B.

\highlight{Evaluation metrics.}
We use \textbf{budget} to refer to the total number of architectures fully-trained in iterative sampling and final model selection. \textbf{Test accuracy} is used as an end-to-end metric, which is the test accuracy (after fully-trained) of the best model ever encountered during the whole search process. To purely assess the quality of a ranker, we use \textbf{top-k test accuracy}~\cite{yu2020train}, in which, we train and evaluate $k$ (\eg $k=10$ in NAS benchmarks) architectures top-ranked by the ranker, and get the test accuracy of the best one. \textbf{Test regret}~\cite{ying2019bench} is to measure the test accuracy gap between the best model found by NAS and the best one in the search space. Notably, we assume the test dataset is invisible during search. Therefore, we always select model on validation dataset and report its accuracy on test dataset.
%We use \textbf{budget} to refer to the total number of architectures trained in iterative sampling and final model selection (as described in the workflow). \textbf{Test accuracy} is used as an end-to-end metric, which is the test accuracy (after fully-trained) of best model ever encountered during the whole search process. To purely assess the quality of a ranking model, we use \textbf{top-k test accuracy}, in which, we train and evaluate $k$ architectures top-ranked by the ranking model, and get the test accuracy of the best one. \textbf{Test regret} is another metric, which measures the test accuracy gap between the best model found by NAS and the best model on search space. Notably, we assume the test dataset is invisible during search. Therefore, we always select model on validation dataset and report its accuracy on test dataset. % In other words, the ``best model'' here means the model with \emph{highest validation accuracy}. \yuge{make sure whether it's common practice}

\begin{table}[t]
\centering
\caption{Comparison with SOTA NAS methods. Budget here refers to the number of fully-trained architectures. $^\dag$ We reproduced BRP-NAS using FLOPs for pre-training.}
\label{tab:comparison-with-sota}
\begin{adjustbox}{width=\linewidth}
% \small
\begin{tabular}{ccccc}
\toprule
	\multirow{2}{*}{Method} & 
	\multicolumn{2}{c}{NAS-Bench-101} & \multicolumn{2}{c}{NAS-Bench-201}\\ 
	& Budget & Test Acc. & Budget & Test Acc. \\ 
	\midrule 
	Oracle & 423,624 & 94.34 & 15,625 & 73.48 \\
	Random & 1,000 & 93.42 & 100 & 69.94 \\
	\midrule 

%     Random \\
 GBDT-NAS~\cite{luo2020neural} & 2,000 & 94.14 & - & - \\
     SemiNAS~\cite{luo2020semisupervised} & 2,000 & 94.02 & - & - \\
    RE~\cite{real2019regularized} & 1,000 & 93.72 & 100 & 70.69 \\
    RL~\cite{zoph2016neural} & 1,000 & 93.58 & 100 & 70.68 \\
    BOHB~\cite{falkner2018bohb} & 1,000 & 93.72 & 100 & 69.71 \\
    % BANANAS~\cite{white2019bananas} & 500 & 94.08 & 
    
    BONAS~\cite{shi2020bridging} & 1,000 & 94.24 & - & - \\
    Unsup. encoding~\cite{yan2020does} & 400 & 94.10 &  - & 73.37 \\
    GA-NAS~\cite{rezaei2021generative} & 1561 & 94.23 & 444 & 73.28 \\
    ReNAS~\cite{xu2021renasrelativistic} & 423 & 93.95 & 90 & 72.12 \\
    Neural Predictor~\cite{wen2020neural} & 219 & 94.04 & - & - \\
    
    % AlphaX & 1072 & 94.23 \\
    BRP-NAS~\cite{chau2020brp} $^\dag$ & \textbf{110} & 94.05 & 110 & 72.79 \\
\midrule
    \algname{} & \textbf{110} & 94.10 & 110 & 73.38 \\
    \algname{} (small) & 30 & 93.92 & 30 & 72.04 \\
    \algname{} (large) & 1,000 & \textbf{94.32} & 500 & \textbf{73.47} \\ 
\bottomrule
\end{tabular}
\end{adjustbox}

% \vspace{-2em}
\end{table}

\subsection{\algname{} on NAS benchmarks}
\label{sec:performance-comp}

\begin{table}[t]
\centering
\caption{Comparisons of test accuracy by different GCN-based approaches on 12 benchmarks (10 search spaces and 3 datasets).}
\label{tab:comparison-on-benchmarks}
\begin{adjustbox}{width=\linewidth}
\begin{tabular}{cccc}
\toprule
Benchmark & Vanilla & BRP & \algname{} \\
\midrule
NAS-Bench-101 & 93.60 & 94.05 & \textbf{94.10} \\
NAS-Bench-201 (CIFAR-10) & 94.10 & 94.35 & \textbf{94.52} \\
NAS-Bench-201 (CIFAR-100)       & 71.96 & 72.93 & \textbf{73.38} \\
NAS-Bench-201 (ImageNet) & 45.70 & 46.31 & \textbf{46.34} \\
NDS-Amoeba  & 94.73 & \textbf{94.85} & 94.84  \\
NDS-DARTS  & 94.84 & 94.88 & \textbf{94.92} \\
NDS-DARTS-fix-w-d  & 94.01 & 94.12 & \textbf{94.16} \\
NDS-ENAS   & 94.78 & 94.91 & \textbf{94.97}  \\
NDS-ENAS-fix-w-d & 93.94 & 94.05 & \textbf{94.13}  \\
NDS-NASNet    & 94.76 & 94.92 & \textbf{95.02} \\
NDS-PNAS       & 94.95 & 94.99 & \textbf{95.06} \\
NDS-PNAS-fix-w-d    & 94.23 & 94.27 & \textbf{94.34} \\
\bottomrule
\end{tabular}
\end{adjustbox}

\end{table}

\begin{figure}[t]
    \centering
    \includegraphics[width=\linewidth]{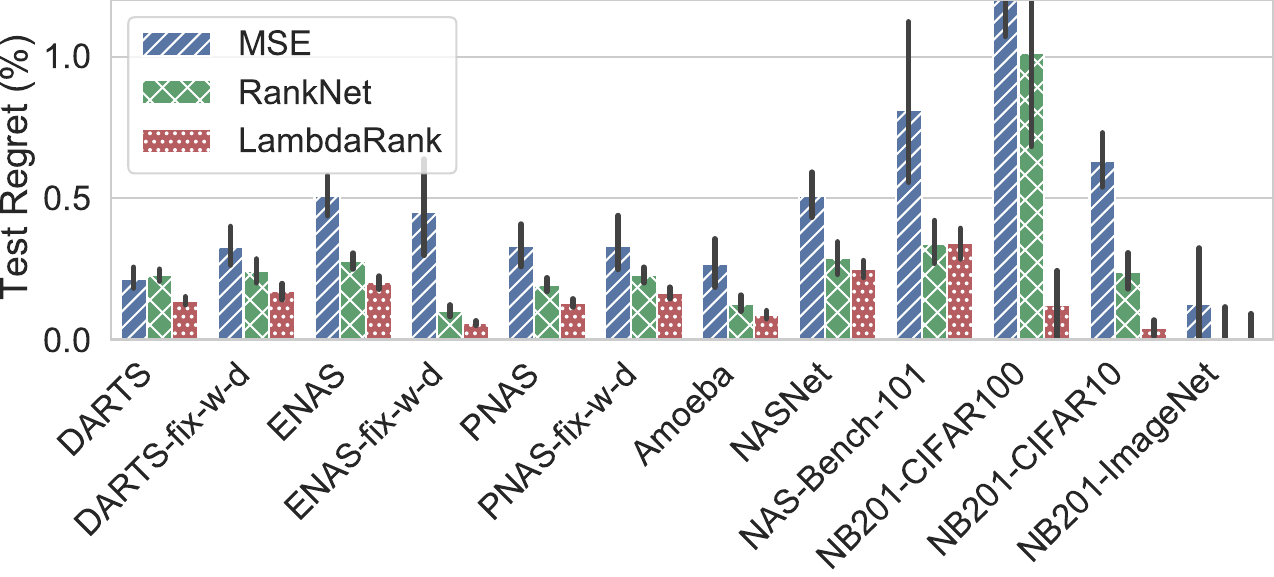}

    \includegraphics[width=\linewidth]{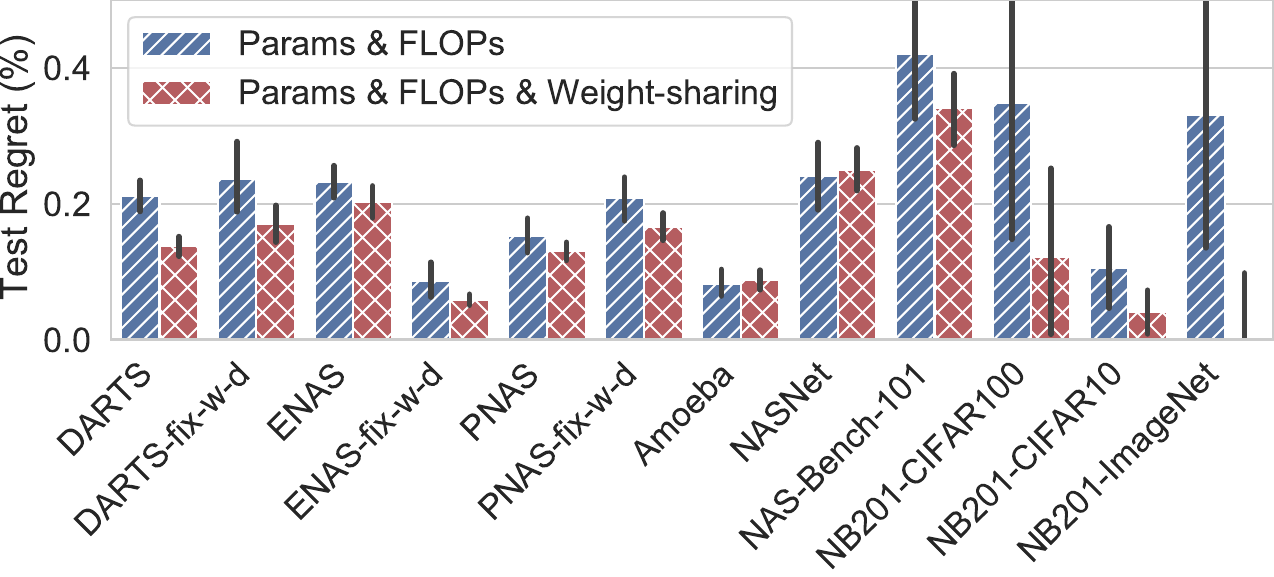}
    \vspace{-1.5em}
    \caption{Comparison among test regrets when using different LTR loss (\textbf{top}) and when pre-training with different labels (\textbf{bottom}). Budget is fixed to 110. }
    \label{fig:ablation-study-loss-resume}
    % \vspace{-1.5em}
\end{figure}

\highlight{Comparison with state-of-the-art methods.} We first evaluate {\algname} by comparing to prior works on NAS-Bench-101 and 201 (CIFAR-100). We list out the results in \cref{tab:comparison-with-sota}, where we run each method on each benchmark 50 times. To evaluate the effectiveness of our work, we set three different levels of budgets (\ie 30, 110, and 1000 fully-trained architectures). Note that \algname{} has an extra weight sharing stage, but its cost is comparable to fully train 2 -- 3 architectures, which is negligible. The budget is splited into 5 rounds of iterative sampling, and we sample another 10 architectures at last, for budget 30 and 110.  Compared to previous works, we achieve a comparable accuracy with only 110 ground-truth accuracy labels on NAS-Bench-101. This cost is 18$\times$ smaller than GBDT-NAS, and 8$\times$ smaller than RE, RL, BOHB, SemiNAS and BONAS. On NAS-Bench-201, we achieve 0.59\% -- 3.67\% higher accuracy when the same number of ground truths are sampled. We further decrease the budget and find that, even with a budget as small as 30, \algname{} is still competitive with several baselines. Notably, when we increase budget to 1,000 on NAS-Bench-101 and 500 on NAS-Bench-201, {\algname} significantly outperforms other baselines. It achieves similar accuracy with the best architecture (oracle), with only 0.02\% and 0.01\% gap on NAS-Bench-101 and NAS-Bench-201, respectively. This result indicates that \algname{} can achieve fast convergence without sacrifice of final performance. %  This result indicates that \algname{} does not easily get stuck in local optimum. \yuge{compare with pure-weight-sharing methods}

\highlight{Comparison with GCN-based approaches on different search spaces.} We then further demonstrate the effectiveness of {\algname} on other search spaces. We implement two most related GCN-based approaches on all benchmarks: \textit{(i)} Vanilla, a basic GCN predictor proposed in \cite{wen2020neural} and \textit{(ii)} BRP-NAS, a state-of-the-art binary-relation predictor that also leverages FLOPs and latency in pre-training. The results (when budget = 110) are shown in \cref{tab:comparison-on-benchmarks}. \algname{} outperforms baselines on 11 out of 12 benchmarks. We demonstrate a more detailed comparison under different budgets in \cref{sec:comparison-different-budgets} and \cref{sec:iterative-sampling}. % To show our effectiveness with fewer samples, we train all the predictors with 20, 40, 60, 80, and 100 architectures. We report the highest accuracy of top 10 models returned by predictors and repeat each experiment for 50 times. 

% We show the results in \cref{fig:top10-best}. On all 10 search spaces, {\algname} consistently find a higher accuracy model than Vanilla and BRP-NAS under the same number of samples. Remarkably, {\algname} also outperforms Vanilla and BRP-NAS under 20 samples, where we can see that {\algname} achieves higher accuracy than Vanilla (up to 0.62\%) and BRP-NAS (up to 0.11\%). 

% \vspace{-1em}

% \subsubsection{Improvement in NDCG}
% \label{sec:ndcg-experiments}

% \vspace{-.5em}

\subsubsection{Ablation study on NAS benchmarks}
\label{sec:ablation-study}

\highlight{LambdaRank vs. other ranking loss.} We compare LambdaRank with MSE loss and RankNet loss. MSE loss is employed in Neural Predictor~\cite{wen2020neural}.  RankNet~\cite{burges2005learning} loss is essentially the same as LambdaRank but does not take into account the changes in NDCG and treat all architecture-accuracy pairs equally (as discussed in \cref{sec:rankingmodel}). As shown in \cref{fig:ablation-study-loss-resume} (\textbf{top}), on all  search spaces, {\algname} with optimizing LambdaRank loss achieves much smaller test regrets than MSE and RankNet. We further conduct comparisons in \cref{sec:more-ablation-study} to demonstrate that LambdaRank loss is consistently outperforming the other two under different pre-training settings. %This echoes the findings in methodology section and demonstrates the effectiveness of LambdaRank. More ablations under different pre-training conditions are available in \cref{sec:more-ablation-study}.
%We first replace LambdaRank in {\algname} with either MSE loss or RankNet loss. RankNet~\cite{burges2005learning} loss is essentially the same as LambdaRank but does not take into account the changes in NDCG and treat all pairs equally. We report the test regret  of top-10 architectures. As shown in \cref{fig:ablation-study-loss-resume} (top), on all  search spaces, {\algname} with LambdaRank achieves much smaller test regrets than MSE and RankNet. This echoes the findings in methodology section and demonstrates the effectiveness of LambdaRank.

% \vspace{2px}

%We evaluate the effectiveness to train a ranking model. We implement a weight sharing guided search~\cite{pourchot2020share} as our baseline for comparison, which selects top-100 architectures with the highest accuracy on super-net. For \algname{}, we also sample 100 architectures, but the first 80 are selected by iterative sampling, and the rest 20 are selected by the final ranking model. As shown in \cref{fig:weight-sharing-comparison}, \algname{} outperforms the baseline with the same search cost (100 samples). Remarkably, we reduce test regret by 3\% on NAS-Bench-201.

% \vspace{2px}
\noindent\textbf{Weak labels.} We then evaluate the effectiveness of using weight sharing labels to pre-train GCN. Our comparison baseline is \textit{Parameters \& FLOPs}, that removes weight sharing from {\algname} (\ie \textit{Params. \& FLOPs \& Weight-sharing}).  \cref{fig:ablation-study-loss-resume} (\textbf{bottom}) shows the test regrets on 12 search spaces and datasets. With the weak supervision of weight sharing labels, we significantly reduce the test regret by up to 0.39\%. % \yuge{check weight sharing only}

\subsubsection{\algname{} under various weight sharing methods}
\label{sec:acenas-weight-sharing}
The previous experiment demonstrates that weak supervision by single-path weight sharing ~\cite{guo2020single} can improve the search accuracy of \algname{}. Then, the question naturally arises: does the quality of weak supervision impact the effectiveness of \algname{}? To verify this, we conduct experiments with various weight sharing methods in state-of-the-art NAS works.

Table~\ref{tab:ws-top-10-accuracy} summarizes the results on two search spaces. To measure the quality of various weak supervision methods, we train the super-net with various weight sharing methods (e.g., Gradient and RL). Then, we retrain the top 10 architectures indicated by the weight sharing method, and use the best accuracy as the method's quality.  %Then, instead of using the original architecture accuracy that is searched by the search algorithm, we evaluate the top-10 accuracy on the super-net as the supervision quality.
As shown in \cref{tab:ws-top-10-accuracy}, when equipped with a better weight sharing method, \algname{} can be even better. % DARTS (1st order) outperforms other weak supervision methods in both search spaces.
Remarkably, by applying the weak supervision of  DARTS (1st order), \algname{} achieves 94.57\% and 94.30\% test accuracy on NAS-Bench-201 (oracle: 94.57\%) and NDS-DARTS-fix-w-d (oracle: 94.32\%), respectively, which are close to the best in the whole search space. Besides,  although DARTS (2nd order) and ProxylessNAS have a much weaker supervision on the NAS-Bench-201 search space, {\algname} can still find good-performing architectures. This suggests that {\algname} is robust to low-quality weak supervision. More results can be found in \cref{sec:combination-weight-sharing}.

\begin{table}[t]
\caption{Top-10 accuracy of \algname{} under various weight-sharing methods. ``NDS-DARTS'' is abbr. for NDS-DARTS-fix-w-d. The maximum value in each column is bold, and the minimum is underlined.}
\label{tab:ws-top-10-accuracy}
\begin{adjustbox}{width=\linewidth}

% \begin{tabular}{cccc}
% \toprule
% Method &  Category                          & NB201 & NDS-DARTS \\
%  \midrule
% RandomNAS~\cite{li2019random}  & Single-path           &                      94.52 &                     94.16 \\
% DARTS (1st order)~\cite{liu2018darts}  & Gradient  &        \textbf{94.57} &      \textbf{94.30} \\
% DARTS (2nd order)~\cite{liu2018darts}  & Gradient  &            94.46 &                  94.12 \\
% FBNet~\cite{wu2019fbnet} & Gradient            &                94.51 &  \underline{94.10} \\
% ProxylessNAS~\cite{cai2019proxylessnas} & Gradient/RL &           \underline{94.44} &              94.16 \\
% ENAS~\cite{pham2018efficient} & RL       &                94.50 &                   94.12 \\ \bottomrule
% \end{tabular}

\begin{tabular}{@{\hskip0pt}c@{\hskip1pt}ccccc}
\toprule
\multirow{2}{*}{WS strategy} & 
\multirow{2}{*}{\begin{tabular}{@{}c@{}}Search \\ method\end{tabular}}           
& \multicolumn{2}{c}{NAS-Bench-201 space} & \multicolumn{2}{c}{NDS-DARTS space} \\
{} & {} & WS quality & \algname{} & WS quality & \algname{} \\ \midrule
RandomNAS & Single-path           &              \texttt{93.16} &              94.52 &              \texttt{93.52} &              94.16 \\
DARTS (1st order) & Gradient  &     \texttt{\textbf{94.45}} &     \textbf{94.57} &     \texttt{\textbf{94.14}} &     \textbf{94.30} \\
DARTS (2nd order) & Gradient  &  \texttt{\underline{84.60}} &              94.46 & \texttt{93.80} &              94.12 \\
FBNet & Gradient            &              \texttt{93.96} &              94.51 &  \texttt{\underline{93.50}} &  {\underline{94.10}} \\
ProxylessNAS & Gradient/RL &              \texttt{87.43} &  \underline{94.44} &              \texttt{93.65} &              94.16 \\
ENAS & RL       &              \texttt{93.67} &              94.50 &\texttt{94.01} &              94.12 \\ \bottomrule
\end{tabular}

% \begin{tabular}{ccccc}
% \toprule
% \multirow{2}{*}{Weight-sharing method}                                 & \multicolumn{2}{c}{NB201 space} & \multicolumn{2}{c}{DARTS space} \\
% {} & WS only & \algname{} & WS only & \algname{} \\ \midrule
% RandomNAS~\cite{li2019random}            &              93.16 &              94.52 &              93.52 &              94.16 \\
% DARTS (1st order)~\cite{liu2018darts}   &     \textbf{94.45} &     \textbf{94.57} &     \textbf{94.14} &     \textbf{94.30} \\
% DARTS (2nd order)~\cite{liu2018darts}   &  \underline{84.60} &              94.46 &              93.80 &              94.12 \\
% FBNet~\cite{wu2019fbnet}               &              93.96 &              94.51 &  \underline{93.50} &  \underline{94.10} \\
% ProxylessNAS~\cite{cai2019proxylessnas} &              87.43 &  \underline{94.44} &              93.65 &              94.16 \\
% ENAS~\cite{pham2018efficient}          &              93.67 &              94.50 &              94.01 &              94.12 \\ \bottomrule
% \end{tabular}

\end{adjustbox}

\end{table}

% \begin{figure}[t]
%     \centering
%     \includegraphics[width=\linewidth]{figures/weight-sharing-comparison-v2.pdf}
%     \caption{Comparison of test regret between weight-sharing-guided greedy search and \algname{} on 100 budget.}
%     \label{fig:weight-sharing-comparison}
%     % \vspace{-1.5em}
% \end{figure}

\begin{table}[t]
    \caption{Top-1 ImageNet accuracy on ProxylessNAS search space. $^\dagger$: accuracy numbers are from the original papers.}
    \label{tab:proxylessnas}
    \centering
    \small
    \begin{adjustbox}{width=\linewidth}
    \begin{tabular}{cccc}
        \toprule
       & Test Acc. (\%) & Latency (ms) \\
        \midrule
      
        TuNAS~\cite{bender2020can}$^\dagger$  &75.0 & 84.0 \\
        ProxylessNAS~\cite{cai2019proxylessnas}$^\dagger$  &74.6 & 84.4 \\
        MnasNet-B1~\cite{tan2019mnasnet}$^\dagger$&74.5 & 84.5 \\
         \midrule
         Neural Predictor~\cite{wen2020neural}  &74.75 & 84.95 \\
       
       % \algname{} (Stage 1) &100& 74.73 & 83.30 \\
        \textbf{\algname{}} & \textbf{75.13} & 84.59 \\
        \bottomrule
    \end{tabular}
    % \vspace{-.5em}
    \end{adjustbox}

    % \vspace{-1em}
\end{table}

\begin{figure}[t]
\includegraphics[width=\linewidth]{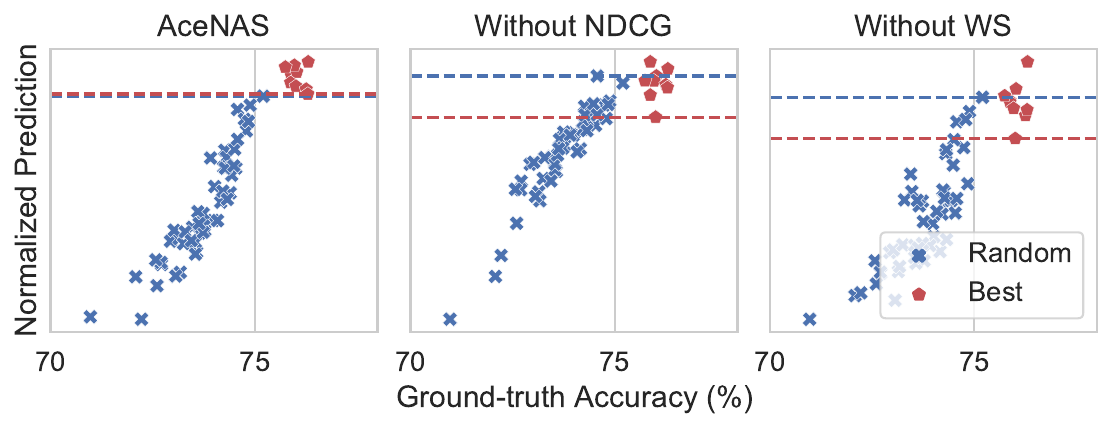}
\caption{Qualitative ablation study of \algname{} on ProxylessNAS. The x-axis is validation accuracy (ground-truth) and y-axis is the normalized prediction score (greater is better). In the middle figure, we replace NDCG with MSE loss. In the right figure, we train the ranker without weak supervision pre-training.}
\label{fig:proxyless-ablation}
% \vspace{-1em}
\end{figure}

\subsection{\algname{} on ProxylessNAS search space}
We now evaluate \algname{} on the large chain-wise search space in ProxylessNAS. We integrate \algname{} into a evolutionary search algorithm~\cite{real2019regularized} by replacing its fitness function with the output of our performance ranker.
The evolution is splitted into 5 stages, where the population size of first stage is 80, and each of the rest stages has 30 individuals. % an initial 80 architectures are randomly sampled for fine-tuning our pretrained ranking model. Then,  we sample 10 random and predict 20 best architectures in next every iteration.
A weight-shared super-net is trained before the first stage begins, which produces weak labels to pre-train the ranker, and the ranker is fine-tuned at the end of each stage.
Architecture training recipes follow the same settings in \cite{bender2020can}. % The 20 best is taken from as random set of 100,000 architectures, following \cite{wen2020neural}. %In the final stage, we follow \cite{wei2020npenas} to run evolution algorithm with the ranking model functioning as a surrogate function. 
\cref{tab:proxylessnas} lists out the results. Compared to state-of-the-art NAS methods on ProxylessNAS search space, 
\algname{} reaches the highest test accuracy of 75.13\% (an average over 3 runs that report 75.29\%, 75.09\%, 75.00\% respectively), outperforming the original ProxylessNAS by as much as 0.53\%. Compared to the state-of-the-art NAS methods, \algname{} achieves 0.38\% higher accuracy than Neural Predictor under a similar budget.  %Remarkably, at the first stage, with only 100 architectures trained, \algname{} has reached 74.73\%, which is comparable to Neural Predictor but at least twice faster.

%Results are shown in \cref{tab:proxylessnas}. With 234 architectures trained and evaluated (in the final stage), \algname{} reaches 75.93\% on validation set and 75.13\% on test set (an average over 3 runs that report 75.29\%, 75.09\%, 75.00\% respectively), outperforming ProxylessNAS by as much as 0.53\%. Remarkably, at the first stage, with only 100 architectures trained, \algname{} has reached 74.73\%, which is comparable to Neural Predictor but at least twice faster.

\highlight{Capability of finding top architectures.}
For a deeper understanding, we evaluate the effectiveness of NDCG and weak supervision pre-training to identify top architectures. We collect the state-of-the-art searched architectures on the ProxylessNAS search space as the best-performing architectures (first group), and randomly sample 50 architectures as the ordinary architectures (second group). 

Since the accuracy of the first group (best-performing architectures) is always higher than the accuracy of the second group (randomly sampled architectures), the scores given by an ideal ranker should be linear to the ground-truth accuracy and be able to separate the two groups accordingly. As shown in \cref{fig:proxyless-ablation} (\textbf{left}), \algname{} accurately discriminate the accuracy differences of two groups.

Besides \algname{}, we train two rankers as baselines: \emph{(i)} replace the LambdaRank with the MSE loss (\cref{fig:proxyless-ablation} (\textbf{middle})), and \emph{(ii)} train the ranker without pre-training (\cref{fig:proxyless-ablation} (\textbf{right})). 
% For each predictor, we visualize its prediction score of each model and the ground-truth model accuracy. 
The results show that neither of the two baselines can separate the two groups of architectures by the prediction scores.
\cref{fig:proxyless-ablation} (\textbf{middle}) indicates that, without LambdaRank, the ranker performs well as a whole (score is basically linear to ground-truth accuracy), but fails to distinguish best-performing architectures. \cref{fig:proxyless-ablation} (\textbf{right}) shows that pre-training also plays an important role. Without weak supervision pre-training, the prediction scores do not well converge to a line proportional to the accuracy. 

\section{Conclusion}

% Architecture ranking is one of the fundamental problems in NAS. In IR, learning to rank documents has been a well-studied problem. Over the past decades, the mainstream LTR methods have been shifting from pointwise and pairwise to listwise. %, gradually relaxing the objective.
% We observe that recent state-of-the-art NAS rankers~\cite{wen2020neural,chau2020brp} fall into pointwise and pairwise from the technical perspective. In this paper, we show that LambdaRank, a listwise method, can further benefit NAS. The NAS community might be embarking on a similar trajectory as in IR.

% Nonetheless, there are also essential differences between architectures and documents, and such differences should be addressed carefully. Our view of NAS from LTR perspective poses both opportunities and challenges to future works.

This paper proves theoretically and empirically that NDCG is a better metric for NAS performance rankers. We then introduce \algname{}, which directly optimizes NDCG and incorporates weak supervision of weight sharing. Extensive experiments under various settings show its effectiveness.

Our study demonstrates the possibility that techniques of LTR in IR community can be applied to the NAS problem, if the discrepancies are addressed carefully. We hope our work will broaden the horizon of future NAS research.

\balance

\nocite{langley00}

\bibliography{egbib}
\bibliographystyle{icml2022}

%%%%%%%%%%%%%%%%%%%%%%%%%%%%%%%%%%%%%%%%%%%%%%%%%%%%%%%%%%%%%%%%%%%%%%%%%%%%%%%
%%%%%%%%%%%%%%%%%%%%%%%%%%%%%%%%%%%%%%%%%%%%%%%%%%%%%%%%%%%%%%%%%%%%%%%%%%%%%%%
% APPENDIX
%%%%%%%%%%%%%%%%%%%%%%%%%%%%%%%%%%%%%%%%%%%%%%%%%%%%%%%%%%%%%%%%%%%%%%%%%%%%%%%
%%%%%%%%%%%%%%%%%%%%%%%%%%%%%%%%%%%%%%%%%%%%%%%%%%%%%%%%%%%%%%%%%%%%%%%%%%%%%%%
\newpage
\appendix
\onecolumn

\appendix

\numberwithin{figure}{section}
\numberwithin{table}{section}

% \onecolumn

\section{Definitions of notations}

We list out the definitions of notations used throughout the paper in \cref{tab:definition-notation}.

\begin{table}[h]
\centering
\caption{The definitions of notations used through the paper.}
\label{tab:definition-notation}
\begin{tabular}{|l|l|}
\hline
$\mathcal{A}$ & Search space  \\
$\alpha$ & An architecture/model in the search space \\
$n$ & Number of architectures to rank \\
$k$ & Number of architectures selected to be fully trained and evaluated \\
$\acc$ & Accuracy of architecture \\
$r$ & Relevance score of architecture (proportional to accuracy) \\
$f$ & Ranker (ranking function) \\
$f_{\omega}$ & Ranker parameterized with $\omega$ \\
$\pif$ & Ranked list (permutation) produced by $f$ \\
\hline
\end{tabular}
\end{table}

\section{Implementation details}
\label{sec:appendix-impl-details}

\subsection{Experiments on NAS benchmarks}

In this section, we first summarize the characteristics of 12 benchmarks we have experimented on. Then we introduce the settings used in weight-shared super-net training. Lastly, we elaborate the details to train GCN, including hyper-parameters and how to handle neural architectures with graph neural networks.

\subsubsection{Search space}

Apart from NAS-Bench-101~\cite{ying2019bench} and NAS-Bench-201~\cite{dong2020nasbench201}, which have been evaluated by many prior works, we leverage 8 more benchmarks from NDS~\cite{radosavovic2019network}. These search spaces are more practical compared to NAS-Bench-101 and NAS-Bench-201, as they originate from SOTA NAS works (\eg NASNet), search for more dimensions (\eg up to 13 op types, width and depth) and hence contain even more architectures compared to commonly-used spaces in NAS literature.

\begin{table}[htbp] 
    \centering
    \caption{Characteristics of search spaces in benchmarks: available datasets$^\star$, number of different cells in total, number of search dimension (SD), and number of architectures available in the benchmark. $^\star$ all search spaces have CIFAR-10 data; $^\dagger$ means it has CIFAR-100 data; $^\ddagger$ means it has ImageNet data (NAS-Bench-201 has ImageNet 16-120 data).}
    \label{tab:search-space-list}
    % \begin{adjustbox}{width=\linewidth}
    \begin{tabular}{cccc}
    \toprule
         Search space & \# Cells & \# SD & \# Benchmarked \\
    \midrule
         NAS-Bench-101 & 423,624 & 1 & 423,624 \\
         NAS-Bench-201 $^\dagger$ $^\ddagger$ & 15,625 & 1 & 15,625  \\ 
         DARTS $^\ddagger$ & $(16,777,216)^2$ & 3 & 5,000 \\
         DARTS-fixwd  & $(16,777,216)^2$ & 3 & 5,000 \\
         ENAS $^\ddagger$ & $(9,765,625)^2$ & 3 & 4,999 \\
         ENAS-fixwd  & $(16,777,216)^2$ & 3 & 5,000 \\
         PNAS $^\ddagger$ & $(1,073,741,824)^2$ & 3 & 4,999 \\
         PNAS-fixwd  & $(1,073,741,824)^2$ & 3 & 4,599 \\
         Amoeba $^\ddagger$ & $(1,073,741,824)^2$ & 3 & 4,983 \\
         NASNet $^\ddagger$ & $(137,858,491,849)^2$ & 3 & 4,846 \\
    \bottomrule
    \end{tabular}
    % \end{adjustbox}

    % \vspace{-2em}
\end{table}

Throughout all our experiments, we follow the guidelines of \cite{ying2019bench} at our best efforts. Firstly, during the search phase, we \textbf{never} use the test dataset. We always compare architectures based on the validation accuracy. Only in the final stage, the selected single architecture is tested on the test dataset (see \cref{alg:nasrank}). Secondly, to simulate the real-world scenario where an architecture is often trained only once, in each of our experiment, we randomly select one trial’s accuracy for each architecture, rather than use the average accuracy of 3 trials.
We repeat every setting with at least 50 experiment runs, thus one architecture can have different accuracies in different runs. 
Although following these settings result in performance drop for most cases, we believe such evaluation is more realistic and we encourage followed-up works to stick to these settings.
% Strictly following these two settings results in drop of performance.  Firstly, for each run, we randomly select one trial’s accuracy for each architecture, rather than use the av- erage accuracy of 3 trials. Secondly, we always select the ar- chitecture with the best validation accuracy and report its test accuracy. Such evaluation is more realistic (Ying et al. 2019). Similarly, the oracle on NAS-Bench-201 of ImageNet16-120 is 46.3% instead of 47%. Therefore, our solution is actually very close to oracle.

\subsubsection{Weight sharing}

In the weight sharing phase of \algname{}, we use RandomNAS (\ie single-path random sampling)~\cite{li2019random,guo2020single} to train a super-net before each search process starts. We follow \cite{pham2018efficient,guo2020single,stamoulis2019singlepath,cai2019once} for handling the dynamic channels and depths during super-net training in NAS-Bench-101 and NDS. On evaluation, 4k architectures are evaluated for each super-net. We calculate the batch normalization statistics on the fly. We run our training on a single Nvidia Tesla V100 with 16GB memory.

We list important hyper-parameters used in super-net training in \cref{tab:hyper-parameters-supernet-training}.

\begin{table}[htbp]
    \centering
    \caption{Important hyper-parameters used in super-net training. $^*$: In search spaces provided by NDS~\cite{radosavovic2019network}, we set batch size to 128 due to limited GPU memory.}
    \label{tab:hyper-parameters-supernet-training}
    \begin{tabular}{|l|l|}
        \hline
        Batch size & 192$^*$ \\
        Number of epochs & 600 \\
        Optimizer & SGD \\
        Initial learning rate & 0.05 \\
        Ending learning rate & 0 \\
        Learning rate schedule & Cosine decay \\
        Weight decay & 0.0001 \\
        Gradient clip & 5 \\
        Evaluate batch size & 512 \\
        \hline
    \end{tabular}

\end{table}

In introduction, \cref{sec:acenas-weight-sharing} and \cref{sec:combination-weight-sharing}, several other weight-sharing approaches have been implemented. Those implementations follow one-shot algorithms provided by Microsoft NNI~\cite{zhang2020retiarii}.

\subsubsection{NDCG}
\label{sec:appendix-ndcg-details}

As mentioned in \cref{sec:ndcg}, accuracy values are usually distributed in a large range and the distribution can be skewed. They can not be directly used as relevance scores to calculate NDCG. Therefore, we first normalize the accuracy values to a smaller range after ignoring outliers. Note that the outliers are only ignored when calculating lower and upper bound, but still taken into account in ranker training. In our experiments, we compute the 20\%-quantile\footnote{We use the implementation of \texttt{numpy.quantile}.} of the original distribution obtained from training data (\ie architecture-accuracy pairs) as lower bound, and directly use the maximum accuracy in the training data as upper bound. This is supported by the empirical observation made by \cite{anonymous2022nasbenchsuite} (Figure 2), that, the performances of worst quarter architectures scatter in a broader range. The values in this range are then linearly mapped to $[0, S]$ to get relevance scores ($S=20$ in experiments). Such recipe is always followed, even if the training data is lacking (only got a few architecture-accuracy pairs). Formally,
\begin{equation}
\label{eq:accuracy-relevance}
r_i = S \cdot \max \left( \frac{\textrm{acc}_i - \textrm{Quantile}(\{\textrm{acc}_i\}, 0.2)}{\max_{1\le i \le n} \textrm{acc}_i - \textrm{Quantile}(\{\textrm{acc}_i\}, 0.2)} , 0 \right)
\end{equation}
where $\textrm{acc}_i$ refers to the accuracy of the $i$-th architecture, and $r_i$ refers to the computed relevance score that is used in NDCG. The outer $\max(\cdot, 0)$ is because NDCG cannot take negative relevance scores, as pointed out by \cite{gienapp2020impact}.

\begin{figure}[htbp]
    \centering
    % \vspace{-0.5em}
    \includegraphics[width=0.5\linewidth]{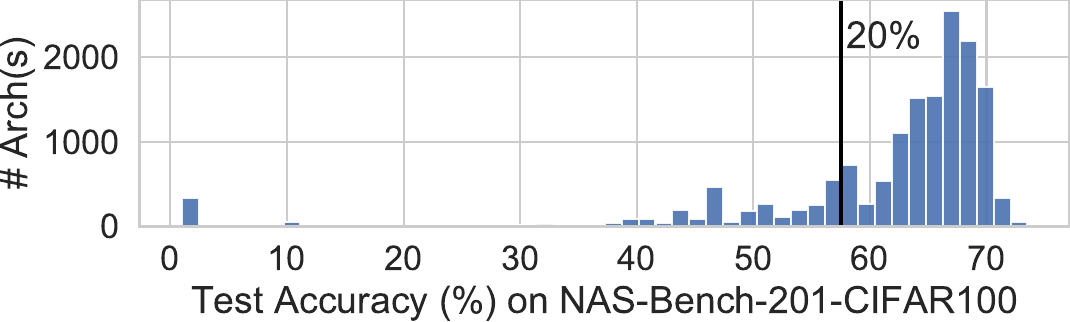}
    \caption{Test accuracy distribution in NAS-Bench-201.}
    % \vspace{-1em}
    \label{fig:distribution-nas-bench-201}
\end{figure}

\cref{fig:distribution-nas-bench-201} shows an illustration of this process on NAS-Bench-201. In this example, around 80\% of the accuracy values are located between 60\% and 70\%, while the other 20\% accuracy values span from 0 to 60\% (we call them outliers). We clip the accuracy to a narrower range to distinguish the top 80\% architectures.

In \cref{sec:ablation-ndcg-parameters}, we further compare different choices of $S$ and different approaches to compute the lower bound.

\subsubsection{GCN}
\label{sec:impl-details-gcn}

\begin{figure}[htbp]
    \centering
    % \vspace{-.5em}
    \includegraphics[width=0.7\linewidth]{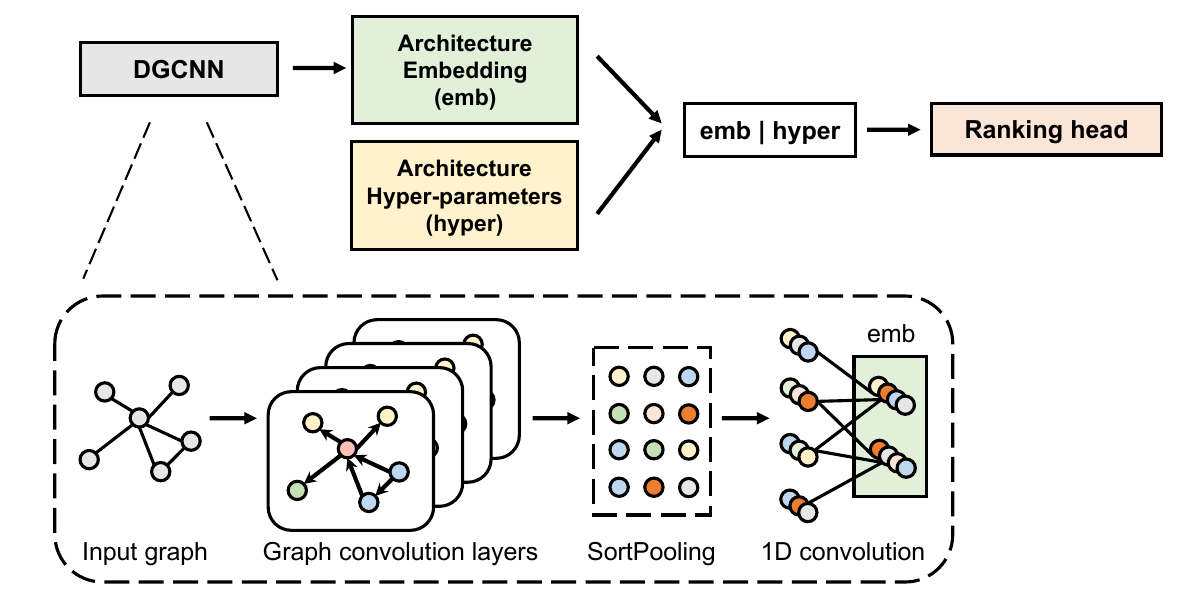}
    \caption{Architecture of our performance ranker.}
    \label{fig:predictor}
    % \vspace{-.5em}
\end{figure}

A crucial step to build a performance ranker for neural architectures, is to get an appropriate embedding of architectures, \ie converting dynamically constructed graphs with different depth and width into a fixed-length vector. Graph Convolutional Network (GCN)~\cite{kipf2016semi} is a natural fit for generating the embedding due to its advantage in dealing with graph-structured data, thus it has been adopted in recent works~\cite{chau2020brp,wen2020neural}. We also use GCN for the embedding. Specifically we choose Deep Graph Convolutional Neural Network (DGCNN)~\cite{zhang2018end} which performs well in our model. It has four directed graph convolution layers followed by sort-pooling~\cite{zhang2018end} and 1D convolution as shown in \cref{fig:predictor}.

Following \cite{wen2020neural}, we encode one type of cell into a directed graph. The type of operator is encoded into a one-hot tensor that is treated as node attributes, and the connections between operators are encoded as edges. Some other pseudo-nodes are necessary to make the graph connected, for example the nodes that are labeled as add/input/output/concatenate. In search spaces provided by NDS, the neural networks search for multiple different types of cells and a series of architecture hyper-parameters (\eg number of cells stacked, channel size multiplier). The graphs are then feeded into GCN and the embedded features are concatenated with hyper-parameter features.

Another component in the performance ranker is ranking head. Our ranking head is a Multi-layer Perceptron, which in our case are two fully-connected layers with ReLU and dropout in between. It predicts ranking score $s_i^*$ for an architecture $\alpha_i$ by taking $\alpha_i$'s embedding from DGCNN and corresponding hyper-parameters. With the ranking score, we use \cref{eq:lambdarank-1} and \cref{eq:lambdarank-2} to optimize the ranker. $\sigma$ in \cref{eq:lambdarank-2} is set to 1.

The ranker is trained using \emph{iterative sampling} which has been widely used in AutoML algorithms (\eg BOHB~\cite{falkner2018bohb} and BRP-NAS~\cite{chau2020brp}). We split the training process into multiple rounds. In each round, we train $n$ architectures  and get $n$ architecture-accuracy pairs, which are used to train the ranker. Then, the ranker is used to sample architectures for the next round. To balance exploration and exploitation, in each round, we sample $\varepsilon \cdot n$ best architectures with our ranker, while the other $(1 - \varepsilon) \cdot n$ are sampled randomly. $\varepsilon$ is set of 0.5 in our experiments.

To empower our ranker with knowledge from super-net, we replace ranking head in the ranker (\cref{fig:predictor}) with a WS-accuracy head, a FLOPs head, and a parameters head, which are three two-layer MLPs to predict weight sharing accuracy, FLOPs and number of parameters respectively. The whole ranker is then pretrained to minimize the loss defined in \cref{eq:multitask}. Empirically we find that the training is not sensitive to $\lambda_1$ and $\lambda_2$, after we normalize the ground truth labels by subtracting mean and dividing by standard deviation. Therefore we simply set $\lambda_1 = \lambda_2 = 1$.
% This ranker is trained with mean-squared-error (MSE) loss instead of using LambdaRank, because weight sharing labels are not qualified for identifying the best architectures from good-performing ones.

We use PyTorch\footnote{We refer to \cite{haowei01pytorch} when implementing LTR algorithms.}  for GCN implementation. We list important hyper-parameters of ranker training in \algname{} in \cref{tab:hyper-parameters-ranking-model}. For BRP-NAS~\cite{chau2020brp} and Neural predictor~\cite{wen2020neural}, we follow the hyper-parameters used in their paper.

\begin{table}[htbp]
    \centering
    \caption{Important hyper-parameters used in ranker training. In pre-training stage, we use the same hyper-parameters, except initial learning rate = 0.001, weight decay = $10^{-5}$ and early stop is disabled.}
    \label{tab:hyper-parameters-ranking-model}
    \begin{tabular}{|l|l|}
        \hline
        Hidden units & 128 \\
        Batch size & 20 \\
        Number of epochs & 300 \\
        Optimizer & Adam \\
        Initial learning rate & 0.005 \\
        Ending learning rate & 0 \\
        Learning rate schedule & Cosine decay \\
        Weight decay & 0.0005 \\
        Early stop patience & 50 \\
        \hline
    \end{tabular}

\end{table}

% The training budget is split into 5 rounds. For example, if 100 architectures are sampled in total, 20 architectures are sampled in each round. We take $\alpha$ = 0.5, which means half are sampled with ranking model and half are sampled randomly. Then $k$ architectures top-ranked by the ranking model are retrained from scratch. We select the model with highest validation accuracy, and report its test accuracy as the final result (\ie top-$k$ accuracy).

\subsection{Experiments on ProxylessNAS}

To run experiment on ProxylessNAS, we first train a super-net with single-path random sampling~\cite{li2019random,guo2020single}. The hyper-parameters used are slightly different from those listed in \cref{tab:hyper-parameters-ranking-model}. We list them in \cref{tab:proxyless-hyper-parameters-supernet-training}. We followed the implementation in \cite{bender2020can} to sample skip connection at 0.5 probability, although we did not apply other tricks, \eg merging convolution kernels. After super-net training is done, we sampled 10000 architectures, where half of them satisfy the latency constraint (83 -- 85ms) and the other half are randomly sampled from the distribution used in super-net training phase.

\begin{table}[htbp]
    \centering
    \caption{Important hyper-parameters used in super-net training of ProxylessNAS. $^*$: We split the 2048 batch size into 16 GPUs and in each mini-batch every GPU samples architectures independently.}
    \label{tab:proxyless-hyper-parameters-supernet-training}
    \begin{tabular}{|l|l|}
        \hline
        Batch size & 2048 $^*$ \\
        Number of epochs & 360 \\
        Warm-up epochs & 5 \\
        Optimizer & SGD \\
        Initial learning rate & 0.48 \\
        Ending learning rate & 0 \\
        Learning rate schedule & Cosine decay \\
        Weight decay & 0.00005 \\
        Accelerator & 16 GPUs \\
        \hline
    \end{tabular}

\end{table}

To train GCN, we used hyper-parameters identical to \cref{tab:hyper-parameters-ranking-model}, except that initial learning rate is decreased to 0.001.

To train the searched architecture (both in the validation setting and test setting), we followed the settings proposed by \cite{bender2020can}. Concretely, during the search process, each architecture is trained for 90 epochs. The final selected architectures are retrained with a longer schedule (360 epochs) with dropout rate 0.15 before the final fully-connected layer. We re-implement it with PyTorch as the original implementation supports TPU only. To align the batch size (4096) with the original setting, we use 16 V100 GPU so that each GPU takes a mini-batch of 256 samples. Ideally, Sync-BN~\cite{peng2018megdet,zhang2018context} should be applied to synchronize batch normalization on all GPUs, however, we find that it harms the training speed by about 50\%. To balance training speed and performance, we used Distribute-BN that synchronizes running statistics of batch normalization at the end of each epoch.

Training a single architecture with 90 epochs take around 5 hours on 16 V100 GPUs, which is around 80 GPU hours. It sums up to 780 GPU days to train all 234 architectures in the search process. To speedup this process, we leverage the metrics of randomly selected architectures from \cite{bender2020can} when \algname{} requests for a new random architecture to be run. This saves the training cost of 140 architectures and thus reduces the total cost to 313 GPU days.

\section{Pseudo-code of \algname{}}
\label{sec:algorithm}

In Algorithm \ref{alg:nasrank}, we present the pseudo-code of \algname{}.

\begin{algorithm*}
	\renewcommand{\algorithmicrequire}{\textbf{Input:}}
	\renewcommand{\algorithmicensure}{\textbf{Output:}}
	\caption{\algname{}}
	\label{alg:nasrank}
	\begin{algorithmic}
		\REQUIRE Search space $\Arch$, budget for each round $n$, budget after training $k$, number of rounds $R$, exploration-exploitation factor $\varepsilon$. Dataset $\mathcal{D}_{\textrm{train}}$, $\mathcal{D}_{\textrm{val}}$ and $\mathcal{D}_{\textrm{test}}$.
		\ENSURE The ranker $f_{\omega}$, the best architecture $\alpha^{\dag}$, the best test accuracy $\textrm{acc}^{\dag}$.
		\STATE {$\triangleright$ \textit{Pre-training}}
		\STATE {Build a weight sharing super-net $S$ based on $\Arch$.}
		\WHILE {not converged}
		    \STATE {Random sample one sub-net $\alpha$ from super-net.}
	        \STATE {Optimize weights corresponding to $\alpha$ and update in $S$.}
		\ENDWHILE
		
		\STATE {Sample sufficient architectures from $\Arch$ and evaluate accuracy, FLOPs and number of parameters on $S$}
		\STATE {Optimize $f_{\omega}$ to minimize $\Loss_{mse}(\acc_i, \acc_i^*) + \lambda_1 \cdot \Loss_{mse}(\flops_i, \flops_i^*) + \lambda_2 \cdot \Loss_{mse}(\params_i, \params_i^*)$.}

        \STATE
		\STATE {$\triangleright$ \textit{Fine-tuning}}
		\STATE {Initialize sampled architectures $A = \varnothing$}
		\FOR{$i=1,...,R$}
		  %  \IF{$i=1$}
		  %      \STATE {$A' \leftarrow$ randomly sampled $n$ architectures.}
		  %  \ELSE
		    \STATE {$\triangleright$ \textit{When the search space is very large, the following step can be solved with evolution.}}
	        \STATE {$A' \leftarrow$ best $\varepsilon \cdot n$ architectures predicted by $f_{\omega}$ and $(1 - \varepsilon) \cdot n$ random architectures.}
		  %  \ENDIF
		    \STATE{$A \leftarrow A \cup \{(\alpha, \textsc{TrainAndEval}(\alpha, \mathcal{D}_{\textrm{train}}, \mathcal{D}_{\textrm{val}}) ~\vert~ \alpha \in A'\}$. \quad $\triangleright$ \textit{This is the most costly step.}}
			\STATE {Fine-tune ranker $f_{\omega}$ on $A$ with LambdaRank.}
		\ENDFOR

        \STATE {$A' \leftarrow$ top-$k$ architectures predicted by $f_{\omega}$.}
        \STATE {$A \leftarrow A \cup \{(\alpha, \textsc{TrainAndEval}(\alpha, \mathcal{D}_{\textrm{train}}, \mathcal{D}_{\textrm{val}}) ~\vert~ \alpha \in A'\}$.}

        \STATE{$\alpha^{\dag} \leftarrow$ architecture with best validation accuracy on $A$.}
        \STATE{$\textrm{acc}^{\dag} \leftarrow \textsc{TrainAndEval}(\alpha, \mathcal{D}_{\textrm{train}}, \mathcal{D}_{\textrm{test}})$}
	\end{algorithmic} 
\end{algorithm*}

\section{Proof of theorems in \cref{sec:ndcg}}
\label{sec:proof}

\subsection{Proof of \cref{theorem:1}}

\begin{proof}
According to the assumption, we have,
\begin{equation}
\sum_{i=1}^n \invlog{i} (2^{r_{\pif(i)}}-1) \ge \nu \left( \sum_{i=1}^n \invlog{i} (2^{r_{i}}-1) \right)
\end{equation}

Let $\rstar = \max_{1 \le i \le k} r_{\pif(i)}$. Since $r_1 \ge r_2 \ge \cdots \ge r_n$ and $\rstar \ge r_{\pif(i)}$ for $1 \le i \le k$, we get the following inequality:
\begin{equation}
\sum_{i=1}^k \invlog{i} (2^{\rstar}-1) + \sum_{i=k+1}^n \invlog{i} (2^{r_{i-k}} - 1) \ge \sum_{i=1}^n \invlog{i} (2^{r_{\pif(i)}}-1) \ge \nu \left( \sum_{i=1}^n \invlog{i} (2^{r_{i}}-1) \right)
\end{equation}
Expand the inequality, we get,
\begin{equation}
\label{eq:ndcg-inequality-1}
2^{\rstar} \left( \sum_{i=1}^k \invlog{i} \right) \ge \sum_{i=1}^{n-k} 2^{r_i} \left(\frac{\nu}{\log_2(i+1)} - \invlog{i+k} \right) + \sum_{i=n-k+1}^{n} 2^{r_i} \left( \frac{\nu}{\log_2(i+1)} \right) + (1 - \nu) \sum_{i=1}^n \invlog{i}
\end{equation}

Recall that $c_i$ is defined to be a sequence of length $n$:
\begin{equation}
\cnu{i} = \begin{cases}
\invlog{i} - \frac{1}{\nu \log_2(i+k+1)} &\text{ if $1 \le i \le n -k$} \\
\invlog{i} &\text{ if $n-k+1 \le i \le n$} 
\end{cases}
\end{equation}
By assumption, it is easy to verify that $\cnu{i} > 0$ for $1 \le i \le n$.

Then \cref{eq:ndcg-inequality-1} can be rewritten as:
\begin{equation}
2^{\rstar} \left( \sum_{i=1}^k \invlog{i} \right) \ge \nu \sum_{i=1}^{n} 2^{r_i} \cnu{i} + (1 - \nu) \sum_{i=1}^n \invlog{i}
\end{equation}

Take log on both sides:
\begin{align}
\rstar & \ge \log_2 \left( \nu \sum_{i=1}^{n} 2^{r_i} \cnu{i} + (1 - \nu) \sum_{i=1}^n \invlog{i} \right) - \log_2 \left( \sum_{i=1}^k \invlog{i} \right) \nonumber \\
& \ge \nu \log_2 \left( \sum_{i=1}^{n} 2^{r_i} \cnu{i} \right) + (1 - \nu) \log_2 \left( \sum_{i=1}^n \invlog{i} \right) - \log_2 \left( \sum_{i=1}^k \invlog{i}  \right) \label{eq:ndcg-inequality-3} \\
& \ge \nu \left( \frac{\sum_{i=1}^{n} r_i \cnu{i}}{\sum_{i=1}^n \cnu{i}} +  \log_2 \left( \sum_{i=1}^n \cnu{i} \right) \right) + (1 - \nu) \log_2 \left( \sum_{i=1}^n \invlog{i} \right) - \log_2 \left( \sum_{i=1}^k \invlog{i}  \right) \label{eq:ndcg-inequality-4} \\
& =\nu \left( \frac{\sum_{i=1}^{n} r_i \cnu{i}}{\sum_{i=1}^n \cnu{i}} \right) + \nu \log_2 \frac{L(k) + \left(1 - \frac{1}{\nu} \right) (L(n) - L(k)) }{L(n)} + \log_2 \frac{L(n)}{L(k)} \nonumber \\
& = \nu \left( \frac{\sum_{i=1}^{n} r_i \cnu{i}}{\sum_{i=1}^n \cnu{i}} \right) + \log_2 \left( 1 - \frac{L(n) - L(k)}{\nu \cdot L(n)} \right)^{\nu} + \log_2 \frac{L(n)}{L(k)} \label{eq:ndcg-inequality-5}
\end{align}
Here \cref{eq:ndcg-inequality-3} and \cref{eq:ndcg-inequality-4} is obtained by using Jensen's inequality twice, and we use $L(n)$ to denote the prefix sum of $\frac{1}{\log_2(i+1)}$, \ie $L(n) = \sum_{i=1}^n \frac{1}{\log_2 (i+1)}$. On the other hand, since $k > > (n+1) - (n+1)^{\nu}$, $\nu > \frac{\log_2(n-k+1)}{\log_2(n+1)}$,
\begin{align}
\frac{L(n) - L(k)}{\nu \cdot L(n)} < & \frac{\log_2 (n+1) (L(n) - L(k))}{\log_2 (n-k+1) L(n)} \\
\le & \frac{\ln (n+1) \left( \li (n+1) - \li(k+1) \right) }{ \ln (n-k+1) \li(n+2) } \label{eq:ndcg-inequality-6} \\
\le & 1 \label{eq:ndcg-inequality-7}
\end{align}
where ${\li (x)=\int _{2}^{x}{\frac {dt}{\ln t}}=\operatorname {li} (x)-\operatorname {li} (2)}$, is the ``offset logarithmic integral''. \cref{eq:ndcg-inequality-6} is by using left/right Riemann sums to bound the sum of $\frac{1}{\log_2 (i+1)}$. See \cref{lemma:rhonk} for the proof of the last step.

Combining \cref{eq:ndcg-inequality-5} and \cref{eq:ndcg-inequality-7}, along with Bernoulli's inequality, we have,
\begin{equation}
\begin{split}
\rstar & \ge \nu \left( \frac{\sum_{i=1}^{n} r_i \cnu{i}}{\sum_{i=1}^n \cnu{i}} \right) + \log_2 \left( 1 - \nu \frac{L(n) - L(k)}{\nu \cdot L(n)} \right) + \log_2 \frac{L(n)}{L(k)} \\
& = \nu \left( \frac{\sum_{i=1}^{n} r_i \cnu{i}}{\sum_{i=1}^n \cnu{i}} \right)
\end{split}
\end{equation}

\end{proof}

A followed corollary of \cref{theorem:1} is,

\begin{corollary}
If $\NDCGExpr \ge \nu$ and $k > (n+1) - (n+1)^{\nu}$,  $\TopKExpr \ge h(\nu)$, where $h(\nu)$ is a increasing function of $\nu$.
\end{corollary}

\begin{proof}
We let,
\begin{equation}
h(\nu) = \nu \left( \frac{\sum_{i=1}^{n} r_i c_i }{\sum_{i=1}^n c_i } \right)
\end{equation}
We note that $\left( \sum_{i=1}^{n} r_i c_i  \right) \Big/ \left( \sum_{i=1}^n c_i \right)$ is a $c_i$-weighted average of $r_i$.  For $i \le n - k$, $c_i$ is increasing when $\nu$ increases. Furthermore, the increment is larger when $i$ is smaller, where $\frac{1}{\log_2 (i+k+1)}$ is larger. Hence, the weighted average leans to larger $r_i$'s as $\nu$ increases.

We conclude $h(\nu)$ is a monotonically increasing function of $\nu$.
\end{proof}

\begin{lemma}
\label{lemma:rhonk}
Let $\rho (n,k) \coloneqq \frac{\ln (n+1) \left( \li (n+1) - \li(k+1) \right) }{ \ln (n-k+1) \li(n+2) }$. $\rho(n,k) \le 1$ for all integers $n \ge k \ge 1$.
\end{lemma}

\begin{proof}
Firstly, we show that $\rho(n,1) \le 1$.

We define $h(n)$ to be,
\begin{equation}
h(n) = \ln \left( \frac{n+1}{n} \right) \li (n+1) - \frac{\ln (n)}{\ln(n+2)} 
\end{equation}

Since,
\begin{equation}
\frac{\delta h(n)}{\delta n} = \log_{(1+n)}\left( 1+\frac{1}{n} \right) - \frac{\ln \left( \frac{n^{1/(n+2)}}{(n+2)^{1/n}} \right)}{ \ln^2 (n+2) } - \frac{\li (n+1)}{n(n+1)} < 0
\end{equation}
$h(n)$ is a monotonically decreasing function of $n$.

On the other hand, we have,
\begin{equation}
\begin{split}
\ln(n+1) \li (n+1) - \ln(n) \li(n+2) & < \ln(n+1) \li(n+1) - \ln(n) \left(\li(n+1) + \frac{1}{\ln(n+2)} \right) \\
& = h(n) \le h(1) = 0
\end{split}
\end{equation}
Hence,
\begin{equation}
\rho(n,1) = \frac{\ln(n+1) \li (n+1) }{\ln(n) \li(n+2)} \le 1
\end{equation}

Then, we show $\rho(n,k)$ is a monotonically decreasing function of $k$.
\begin{equation}
\frac{\delta \rho (n,k)}{\delta k} = \frac{\ln (n+1) \left( \ln (k+1) ( \li(n+1) - \li(k+1)) - (n-k+1) \ln (n-k+1) \right)}{(n+1-k) \ln (k+1) \ln^2 (n+1-k) ( \li (n+2))}
\label{eq:rhonk-derivative}
\end{equation}

Obviously, the sign of \cref{eq:rhonk-derivative} depends on the sign of $\left( \ln (k+1) \cdots \ln (n-k+1) \right)$. Let $\alpha = k+1$, $\beta = n-k+1$, so that $\alpha + \beta = n+2$, and $2 \le \alpha \le n + 1$, $1 \le \beta \le n$. Then,
\begin{equation}
\textrm{sign} \left( \frac{\delta \rho (n,k)}{\delta k}  \right) = \textrm{sign} \left( \ln \alpha (\li(\alpha+\beta-1) - \li (\alpha)) - \beta \ln \beta \right)
\label{eq:rhonk-derivative}
\end{equation}

Since we have,
\begin{equation}
\li(\alpha+\beta-1)-\li(\alpha) = \int_{\alpha}^{\alpha+\beta-1} {\frac {dt}{\ln t}} \le \frac{\beta-1}{\ln \alpha}
\label{eq:rhonk-derivative-cond}
\end{equation}
Combining \cref{eq:rhonk-derivative} and \cref{eq:rhonk-derivative-cond}, we get,
\begin{equation}
\textrm{sign} \left( \frac{\delta \rho (n,k)}{\delta k}  \right) \le \textrm{sign} \left( \beta (1 - \ln \beta) - 1 \right) < 0
\end{equation}

Hence, we conclude $\rho(n,k) \le \rho(n,1) \le 1$.

\end{proof}

\subsection{Proof of \cref{theorem:2}}

\begin{proof}
For a ranker $f$, let $\TopKExpr \ge y_t$, where $1 \le t \le n - k + 1$. Since $\E[\NDCGExpr]$ = $\frac{\E[\textrm{DCG}(f; \mathbf{a}, \mathbf{r})]}{\textrm{IDCG}(f; \mathbf{a}, \mathbf{r})}$, we only need to prove that, $\E[\textrm{DCG}(f; \mathbf{a}, \mathbf{r})]$ is a monotonically decreasing function of $t$.

Since we want to have at least one of $\pif(1), \pif(2), \ldots, \pif(k)$ to be at most $t$, the total number of different $\pif$ that satisfies such condition is,
\begin{equation}
C_{\textrm{tot}}(n,k,t) = n! - \frac{(n-t)!(n-k)!}{(n-t-k)!}
\end{equation}
Then, we compute the number of permutations where $\pif(i)=j$ ($1 \le j \le n$), which we denote as $C(i,j;n,k,t)$. Apparently, $i=1,2,\ldots,k$ are symmetric, \ie $C(i_1,j) = C(i_2,j)$ for $1 \le i_1, i_2 \le k$. The same is for $i=k+1,k+2,\ldots,n$, $j=1,2,\ldots,t$, and $j=t+1,t+2,\ldots,n$. Thus, $C(i,j)$ can be divided into four cases:
\begin{equation}
\label{eq:comb-four-cases}
C(i,j;n,k,t)=\begin{cases}
(n-1)! &\text{ if $1 \le i \le k$ and $1 \le j \le t$} \\
\left(C_{\textrm{tot}}(n,k,t) - t \cdot (n-1)! \right) \big/ \left( n-t \right) &\text{ if $1 \le i \le k$ and $t + 1 \le j \le n$} \\
(n-1)! - \frac{(n-t)! (n-k-1)!}{(n-t-k)!} &\text{ if $k+1 \le i \le n$ and $1 \le j \le t$} \\
\left(C_{\textrm{tot}}(n,k,t) - t \cdot \left( (n-1)! - \frac{(n-t)! (n-k-1)!}{(n-t-k)!} \right) \right) \big/ \left( n-t \right) &\text{ if $k+1 \le i \le n$ and $t + 1 \le j \le n$}
\end{cases}
\end{equation}
Note that the expressions in \cref{eq:comb-four-cases} can be further simplified, but we keep them in their raw forms so that readers can easily see where they come from.

On the other hand, $\E[\textrm{DCG}(f; \mathbf{a}, \mathbf{r})]$ can be evaluated as:
\begin{equation}
\label{eq:comb-dcg}
\begin{split}
\E[\textrm{DCG}(f; \mathbf{a}, \mathbf{r})]
&= \E \left[ \sum_{i=1}^n \frac{1}{\log_2 (i+1)} \left( 2^{r_{\pif(i)}}-1 \right) \right] \\
&= \sum_{i=1}^n \frac{1}{\log_2 (i+1)} \left[ \E \left(2^{r_{\pif(i)}} \right) -1 \right] \\
&= \sum_{i=1}^n \frac{1}{\log_2 (i+1)} \left[ \sum_{j=1}^n 2^{r_j} \cdot \Pr(\pif(i)=j)  -1 \right] \\
&= \logsum{1}{n} \left[ \sum_{j=1}^n 2^{r_j} \cdot \frac{C(i,j;n,k,t)}{C_{\textrm{tot}}(n,k,t)}  -1 \right]
\end{split}
\end{equation}

By substituting \cref{eq:comb-four-cases} into \cref{eq:comb-dcg}, we get:
\begin{align}
\E[\textrm{DCG}(f; \mathbf{a}, \mathbf{r})]
&= \bigg[
\left( \logsum{1}{k} \right) \left( \relexpsum[j]{1}{t}  \right) \left(n-1 \right)! \nonumber \\
& \quad \quad + \left( \logsum{1}{k} \right) \left( \relexpsum[j]{t+1}{n}  \right) \left( \frac{n!}{n-t} - \frac{(n-t-1)!(n-k)!}{(n-t-k)!} - \frac{t \cdot (n-1)!}{n-t} \right) \nonumber \\
& \quad \quad  + \left( \logsum{k+1}{n} \right) \left( \relexpsum[j]{1}{t}  \right) \left( (n-1)! - \frac{(n-t)! (n-k-1)!}{(n-t-k)!} \right) \nonumber \\
& \quad \quad + \left( \logsum{k+1}{n} \right) \left( \relexpsum[j]{t+1}{n}  \right) \bigg( \frac{n!}{n-t} - \frac{(n-t-1)!(n-k)!}{(n-t-k)!} \nonumber \\
& \quad \quad \quad \quad \quad - t \cdot \left( \frac{(n-1)!}{n-t} - \frac{(n-t-1)! (n-k-1)!}{(n-t-k)!} \right) \bigg)
& \bigg] \nonumber \\
& \quad \quad \Big/ \left( n! - \frac{(n-t)!(n-k)!}{(n-t-k)!} \right)
\label{eq:comb-complex}
\end{align}

For notation simplicity, for the rest of this proof, we use $L(n) = \logsum{1}{n}$, and $R(n) = \relexpsum{1}{n}$. After this, \cref{eq:comb-complex} can be further rewritten and simplified as:
\begin{equation}
\label{eq:comb-complex-simp}
\begin{split}
\E[\textrm{DCG}(f; \mathbf{a}, \mathbf{r})]
= & \frac{ (nR(t)L(k) + kR(n)L(n) - (nR(n)+kR(t))L(n)+ tR(n)(L(n)-L(k)) (n-k-1)!(n-t-1)! }{n!(n-t-k)! - (n-t)!(n-k)!} \\
& \quad \quad + \frac{R(n)L(n) (n-1)!(n-k-t)!}{n!(n-t-k)! - (n-t)!(n-k)!}
\end{split}
\end{equation}

Define $\gamma(n,t,k)$ as ${n \choose k} \big/ {n-t \choose k}$. We rewrite \cref{eq:comb-complex-simp} with $\gamma(n,t,k)$:
\begin{equation}
\label{eq:comb-complex-simp-final}
\begin{split}
\E[\textrm{DCG}(f; \mathbf{a}, \mathbf{r})]
= & \frac{ (nR(t)L(k) + kR(n)L(n) - (nR(n)+kR(t))L(n)+ tR(n)(L(n)-L(k)) \gamma(n,t,k) }{(n-t)(n-k)(1 - \gamma(n,t,k))} \\
& \quad \quad + \frac{R(n)L(n) }{n (1 - \gamma(n,t,k))}
\end{split}
\end{equation}

By Stirling's formula, when $n \gg k$, $\gamma(n,t,k)$ can be approximated as,
\begin{equation}
\label{eq:comb-gamma}
\gamma(n,t,k) = {n \choose k} \bigg/ {n-t \choose k} = \frac{n!(n-t-k)!}{(n-t)!(n-k)!} \approx \left(\frac{n-t}{n} \right)^k
\end{equation}
Thus, by replacing $\gamma(n,t,k)$ with \cref{eq:comb-gamma}, \cref{eq:comb-complex-simp-final} becomes,
\begin{equation}
\E[\textrm{DCG}(f; \mathbf{a}, \mathbf{r})]
= \left( \frac{R(n)L(n)}{n} + \frac{ \left( \frac{n-t}{n} \right)^k \left(  L(k) (n R(t) - t R(n)) - L(n) (( n-k-t) R(n) + k R(t) ) \right) }{(n-t)(n-k)} \right) \bigg/ \left(1 - \left(\frac{n-t}{n} \right)^k \right)
\end{equation}

We then take the partial derivative of $\E[\textrm{DCG}(f; \mathbf{a}, \mathbf{r})]$ with respect to $t$:
\begin{equation}
\label{eq:comb-complex-simp-final-derivative}
\frac{\delta \E[\textrm{DCG}(f; \mathbf{a}, \mathbf{r})]}{\delta k} = \frac{\zeta (n L(k)-k L(n)) \left((n \left(\zeta-1\right) + kt)R(n) +n (n-t) \left(1-\zeta \right) \frac{\delta R(t)}{\delta t} -n  \left(\zeta+k-1\right) R(t) \right)}{n (n-k) (n-t)^2 \left(\zeta-1\right)^2}
\end{equation}
Note that in this equation (and for the rest of this proof), we use $\zeta = \left(1-\frac{t}{n}\right)^k$, which is implicitly an function of $n$, $t$, and $k$, to simplify the notation.

In \cref{eq:comb-complex-simp-final-derivative}, obviously the denominator is greater than 0. $0 < \zeta < 1$. $nL(k) - kL(n) > 0$ because $L(n) \approx \li(n) \cdot \ln 2 \sim \frac{n}{\ln n}$. For the coefficient of $\frac{\delta R(t)}{\delta t}$ and $R(t)$, $n (n-t) \left(1-\zeta \right) > 0$, while $-n  \left(\zeta+k-1\right) < 0$. On the other hand, we have,
\begin{equation}
\frac{\delta R(t)}{\delta t} \approx 2^{r_t}, \quad \text{ and}
\end{equation}
\begin{equation}
R(t) = \relexpsum{1}{t} \ge t 2^{r_t}
\end{equation}
% \begin{equation}
% n 2^{r_t} \ge \relexpsum{1}{n} = R(n)
% \end{equation}
Therefore,
\begin{equation}
\label{eq:comb-complex-simp-final-derivative-ineq}
\begin{split}
\textrm{sign} \left( {\frac{\delta \E[\textrm{DCG}(f; \mathbf{a}, \mathbf{r})]}{\delta k}} \right)
&= \textrm{sign}  \left((n \left(\zeta-1\right) + kt)R(n) +n (n-t) \left(1-\zeta \right) \frac{\delta R(t)}{\delta t} -n  \left(\zeta+k-1\right) R(t) \right) \\
% & \ge \textrm{sign}  \left((n \left(\zeta-1\right) + kt)R(n) +(n-t) \left(1-\zeta \right) R(n) -n  \left(\zeta+k-1\right) R(t) \right) \\
% & = \textrm{sign} \left( (k + \zeta - 1) ( t R(n) - n  R(t) ) \right)
& \le \textrm{sign} \left( (n \left(\zeta-1\right) + kt)R(n) + n (n-t) \left(1-\zeta \right) 2^{r_t} -n  \left(\zeta+k-1\right) t 2^{r_t} \right) \\ 
&.= \textrm{sign} \left( (R(n) - n 2^{r_t})( (\zeta - 1) n + kt) \right)
\end{split}
\end{equation}

By Bernoulli's inequality, $\zeta = (1-\frac{t}{n})^k \ge 1 - \frac{kt}{n}$. Hence, $(\zeta - 1)n + kt \ge 0$. Since $R(n) \le n 2^{r_t}$ by assumption, $\frac{\delta \E[\textrm{DCG}(f; \mathbf{a}, \mathbf{r})]}{\delta k} > 0$. We conclude that $\E[\textrm{DCG}(f; \mathbf{a}, \mathbf{r})]$ is a function that is monotonically decreasing to $k$, thus complete the proof.
\end{proof}

\section{More experiment results}

\subsection{Empirical study of NDCG}

\subsubsection{Correlation between NDCG and accuracy}
\label{sec:correlation-ndcg-accuracy}

\begin{figure*}[htbp]
    \centering
    \includegraphics[width=\textwidth]{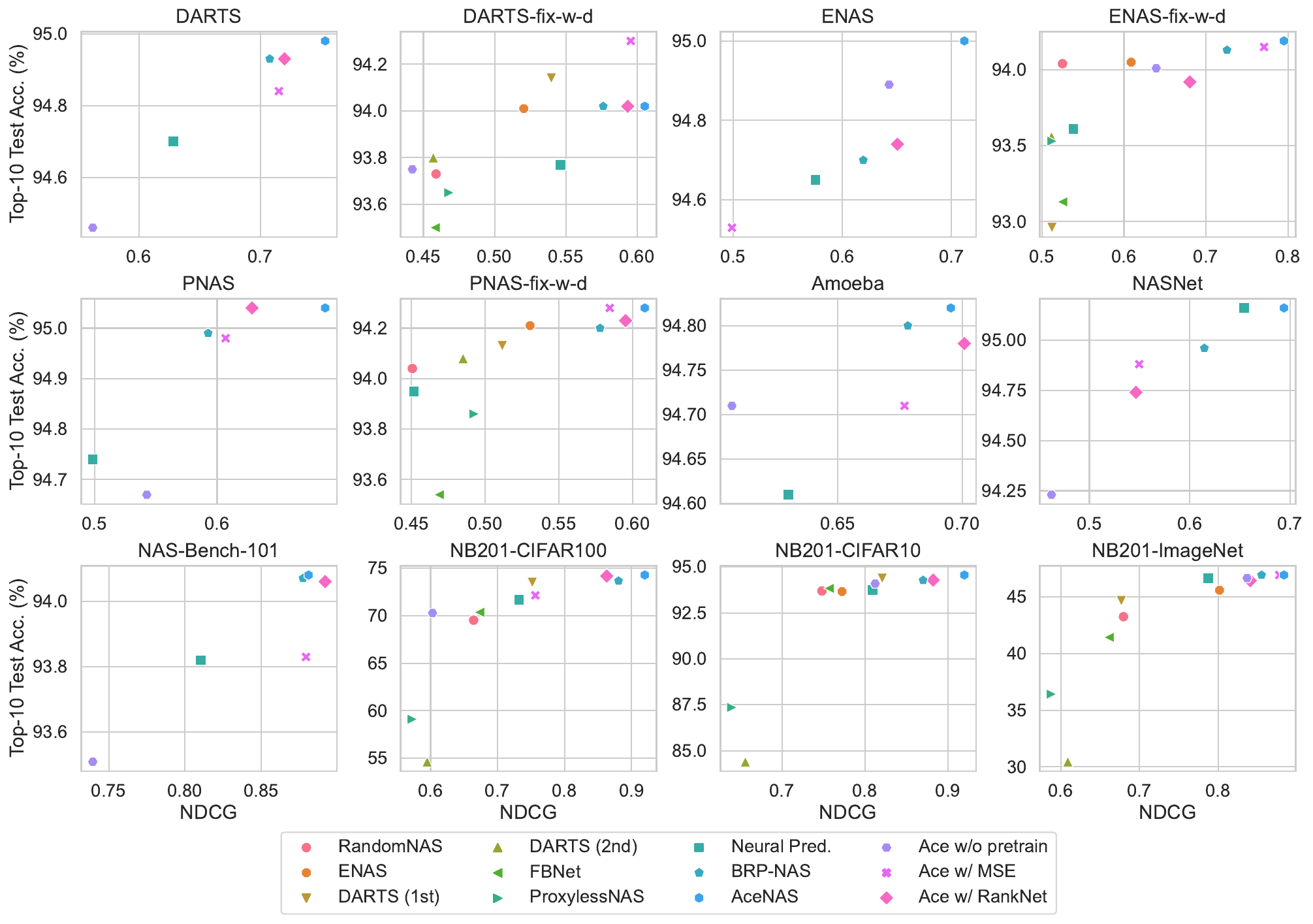}
    \caption{Scatter-plot of NDCG vs. top-10 test accuracy of a various kind of rankers, experimented on various search spaces.}
    \label{fig:ndcg-top10-corr}
\end{figure*}

To answer the question whether NDCG is a good indicator, we first collect a number of rankers on each search space, which spans rankers based on weight sharing, several performance predictors, and \algname{}. We run each of them once, and measure their NDCG and top-10 test accuracy in \cref{fig:ndcg-top10-corr}.  The upward trend in this figure indicates a positive correlation between NDCG and test accuracy. Notably on the spaces of NAS-Bench-201, for rankers achieving test accuracy between 70\% and 75\%, NDCG spans a wide range from 0.6 and 0.9, indicating that NDCG is numerically sensitive to top-performing architectures.

\begin{figure*}[htbp]
    \centering
    \includegraphics[width=\textwidth]{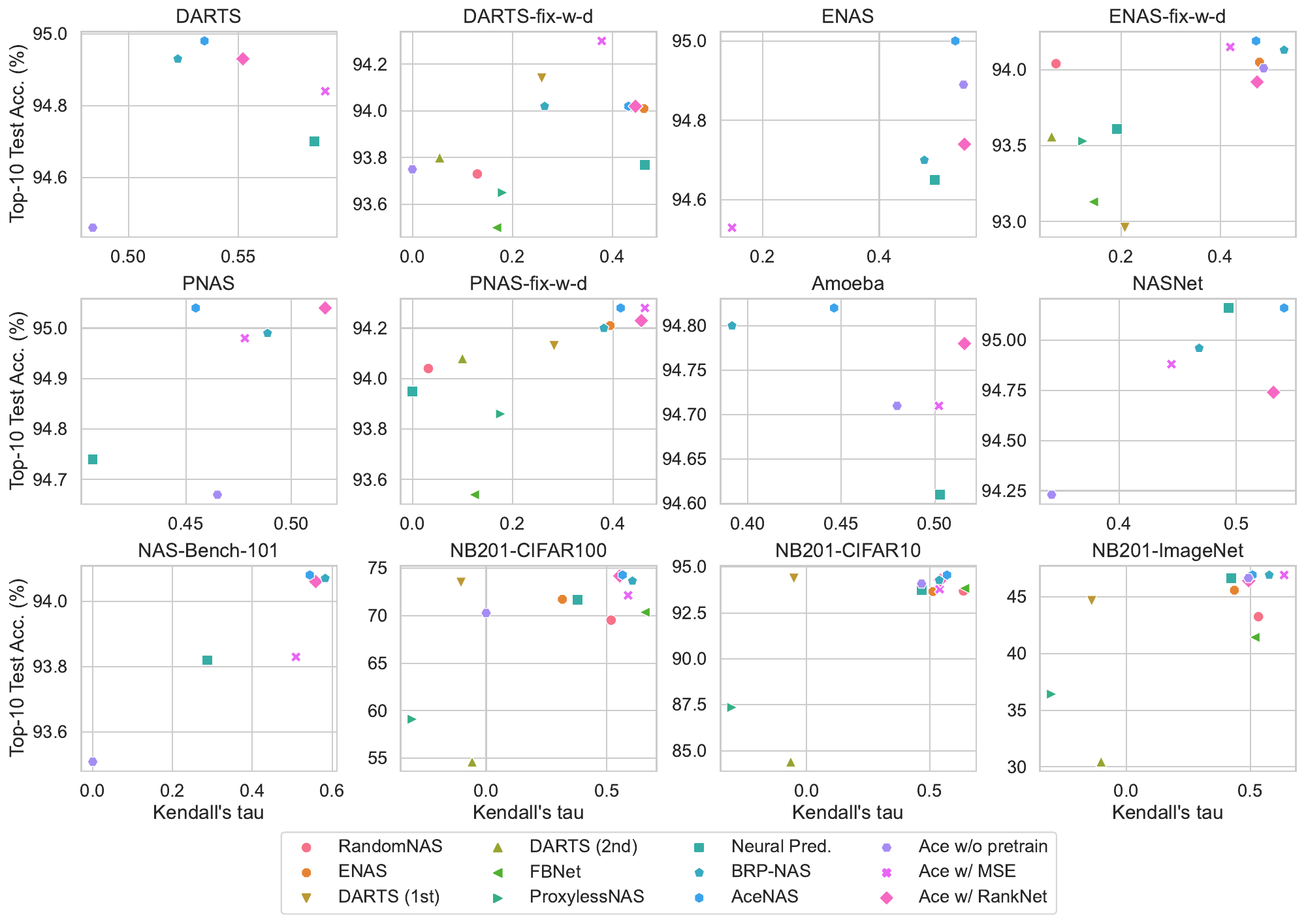}
    \caption{Scatter-plot of KdT vs. top-10 test accuracy of a various kind of rankers, experimented on various search spaces.}
    \label{fig:kdt-top10-corr}
\end{figure*}

A similar scatter-plot is shown in \cref{fig:kdt-top10-corr}, which illustrates the relationship between KdT and test accuracy. In this figure, we observe many cases where a ranker with a high KdT does not have a good test accuracy. Another interesting finding is that \algname{}'s variant (AceNAS w/ RankNet) enjoys KdT close to \algname{} on most search spaces, and sometimes it is even better, but final accuracy is relatively low. Comparing \cref{fig:ndcg-top10-corr} with \cref{fig:kdt-top10-corr}, we can conclude once more, that KdT is a worse metric for rankers than NDCG.

% We collect 12 NAS performance rankers, spanning from sampling-based performance predictors to weight-sharing based methods (implementation detailed in \cref{sec:impl-details}). For each method, we train a ranker and measure its NDCG and top-10 test accuracy on each search space and dataset, respectively. Results on more search spaces are available in \cref{sec:correlation-ndcg-accuracy}. The scatter plot is shown in \cref{fig:ndcg-accuracy-12-ranker}. The upward trend in this figure indicates a positive correlation between NDCG and test accuracy. Notably, for rankers achieving test accuracy between 70\% and 75\%, NDCG spans a wide range from 0.6 and 0.9, indicating that NDCG is numerically sensitive to top-performing architectures.

\begin{figure*}[htbp]
    \centering
    \includegraphics[width=\textwidth]{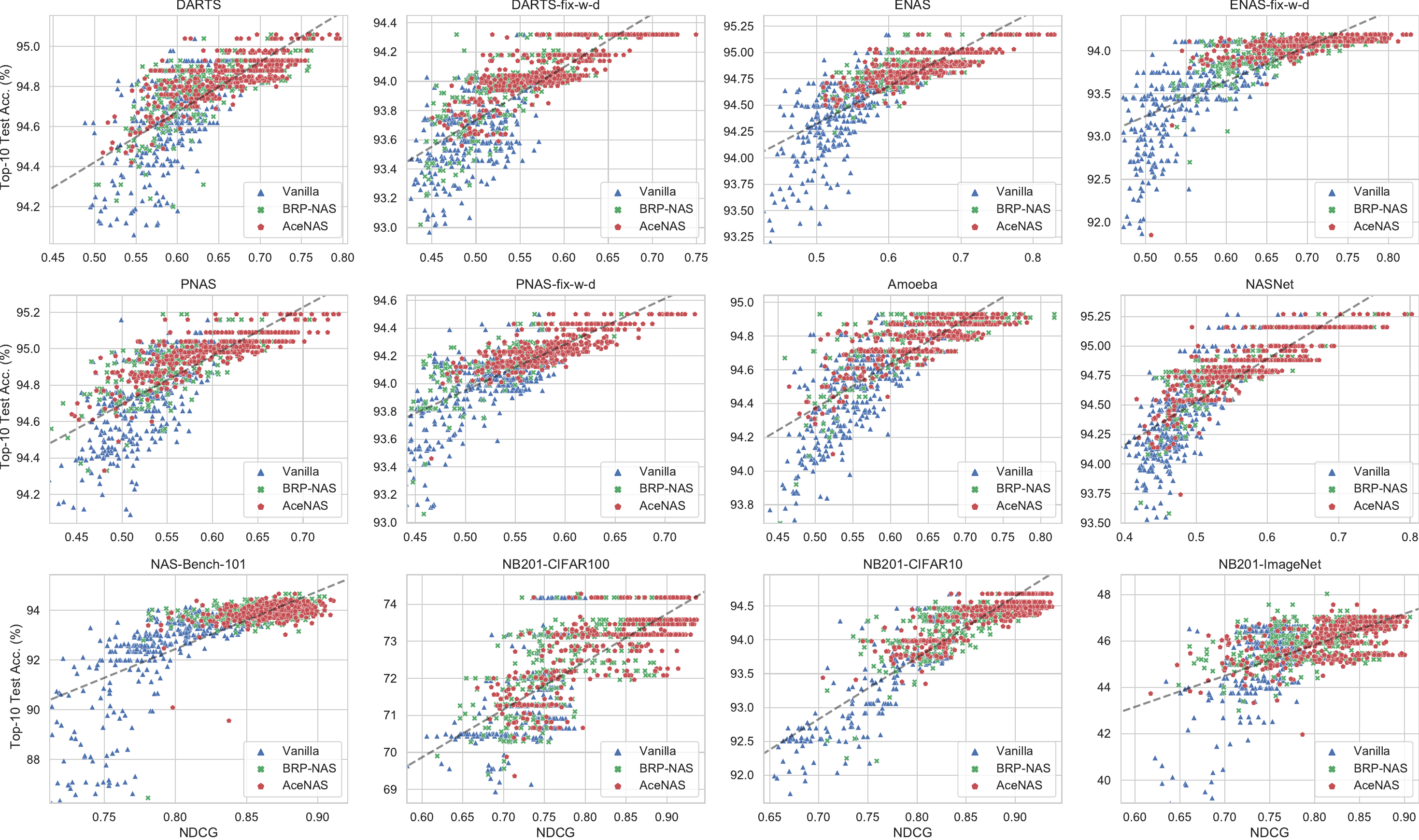}\textbf{}
    \caption{\algname{} achieves better NDCG and accuracy on different search spaces. The upward trends show the correlation between NDCG and accuracy.}
    \label{fig:ndcg-top10-corr-12fig}
\end{figure*}

Moreover, we collect NDCG metric of all experiments and runs covered in \cref{sec:experiments}, and show them in \cref{fig:ndcg-top10-corr-12fig}. In this figure, each point corresponds to an experiment (\ie a performance ranker trained with a specific seed and budget). In this figure, we see a upward trend which is similar to \cref{fig:ndcg-top10-corr}. When looking at points with different markers and colors, we further point out the distribution of different methods, which illustrates how \algname{} improves NDCG and accuracy simultaneously. Compared to points from Vanilla and BRP-NAS, most points from \algname{} are at the top-right corner, meaning that they enjoy both a better NDCG and a better test accuracy at the same time.

\begin{figure}[htbp]
\centering
\includegraphics[width=0.6\linewidth]{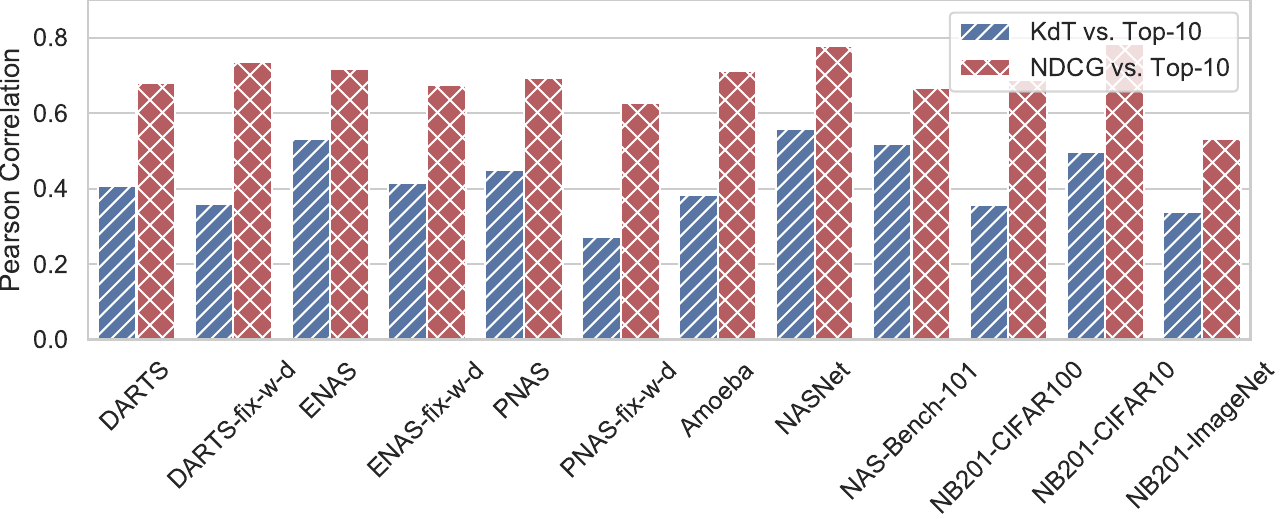}
\caption{Pearson correlation coefficients of different ranking metrics with respect to top-10 accuracy.}
\label{fig:kendall-tau-ndcg-comparison}
\end{figure}

We also calculate the pearson correlation between NDCG and test accuracy (\ie x-y correlation in \cref{fig:ndcg-top10-corr}) and we find that NDCG is well correlated with the end-to-end goal of NAS (up to $\sim$0.8 correlation). This is higher than the correlation between KdT and test accuracy. The results are shown in  \cref{fig:kendall-tau-ndcg-comparison}.

\subsubsection{NDCG relevance score}
\label{sec:ablation-ndcg-parameters}

In this section, we study two important factors in \cref{eq:accuracy-relevance}, and exploit different kinds of design choices. The experiment settings are the same as \cref{sec:ablation-study}. We show the top-10 test regret (the gap between top-10 test accuracy and oracle test accuracy), and the total budget is fixed to 110.

\begin{figure}[htbp]
    \centering
    \includegraphics[width=\textwidth]{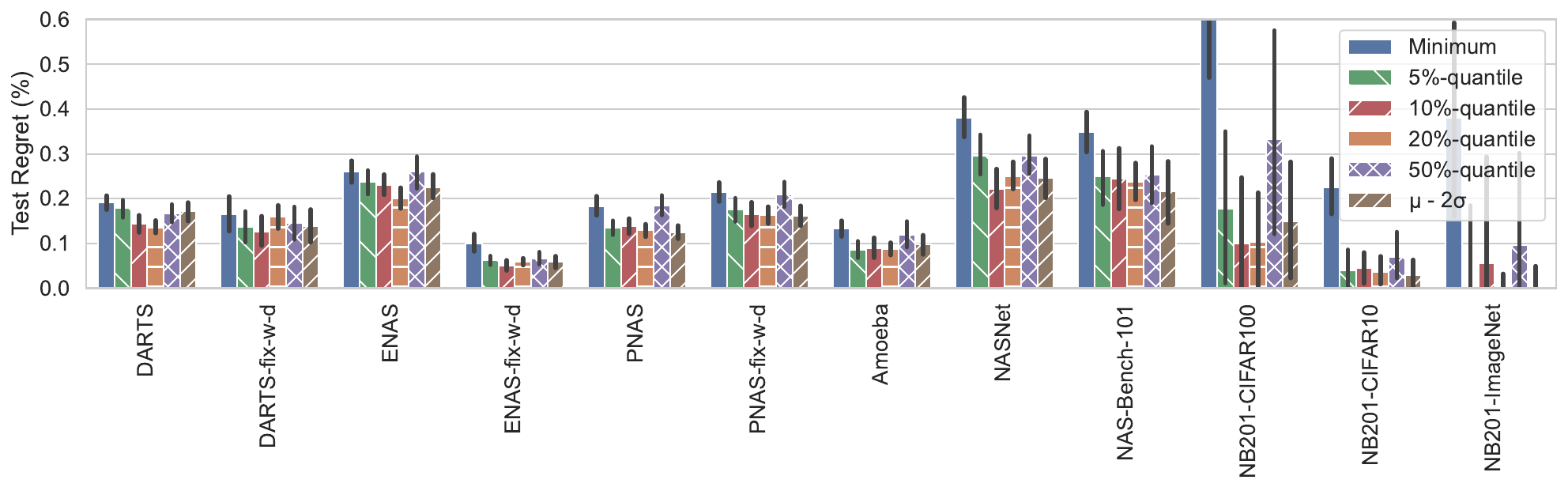}
    \caption{Comparison of different methods to compute lower bound to normalize accuracies in NDCG computation.}
    \label{fig:ablation-ndcg-outlier-removal}
\end{figure}

We first compare different approaches to calculate upper bounds and lower bounds for normalization. Recall that, by default, the upper bound is the estimated with the maximum accuracy in the training samples, while the lower bound is the 20\% quantile. Such strategy is compared with: \emph{(i)} directly using minimum as lower bound; \emph{(ii)} using 5\%, 10\% or 50\% quantile as lower bound; \emph{(iii)} using mean minus two times standard deviation as lower bound by assuming data to conform to a gaussian distribution. The results are shown in \cref{fig:ablation-ndcg-outlier-removal}, where we see that all the strategies perform similarly, except the minimum strategy which does not take outliers into consideration. The performance drop is particularly large on search spaces provided by NAS-Bench-201, where the accuracy distribution is highly skewed (see \cref{fig:distribution-nas-bench-201}). Other than minimum, 50\%-quantile also performs a little worse than others. We hypothesize that the ranking information of the bottom items is lost due to an over-aggressive clipping. As a result, the ranker fails to fully leverage all the data in training.

\begin{figure}[htbp]
    \centering
    \includegraphics[width=\textwidth]{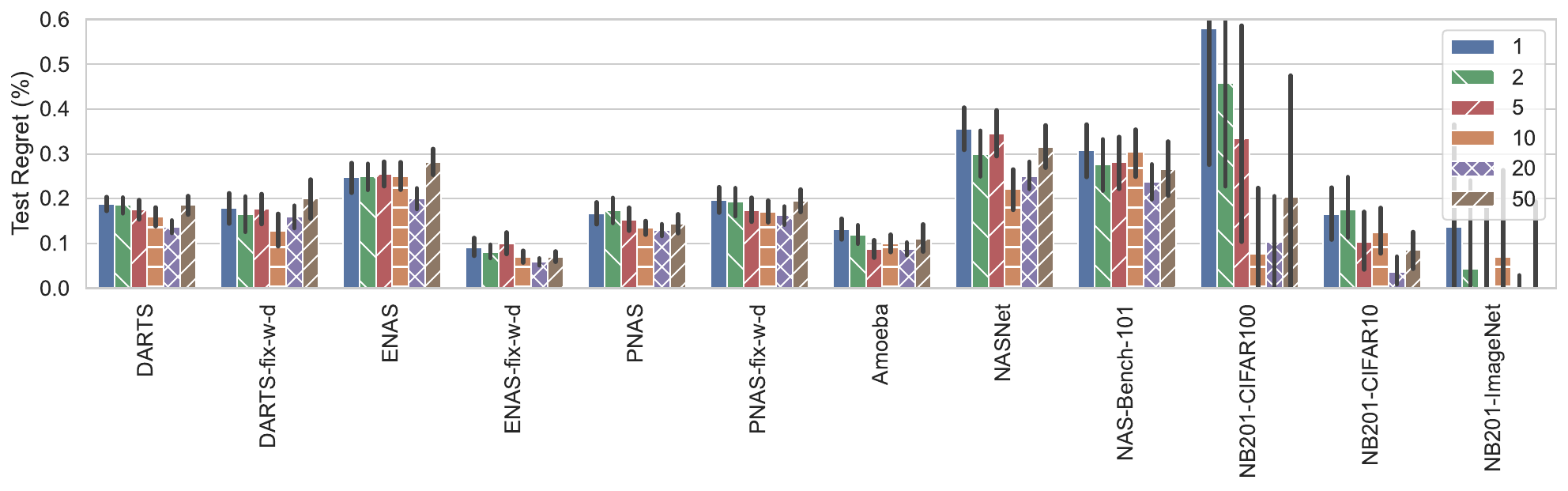}
    \caption{Comparison of different relevance scales ($S$) in NDCG computation.}
    \label{fig:ablation-ndcg-relevance-scale}
\end{figure}

We then compare different relevance scale, \ie $S$ in \cref{eq:accuracy-relevance}. By default $S=20$. Other choices of $S$ include 1, 2, 5, 10 and 50. We show the comparison in \cref{fig:ablation-ndcg-relevance-scale}. Intuitively, higher $S$ puts a higher emphasis on top-ranked architectures, which according to the claim of this paper, is more likely to produce higher performance. The experiment results show that this claim is generally true, except there is a slight performance drop when $S=50$. We believe that the drop when $S=50$ is because too large numbers used in the power of 2 result in numerical instability.

\begin{figure}[htbp]
    \centering
    
    \begin{subfigure}{\linewidth}
        \centering
        \includegraphics[width=\textwidth]{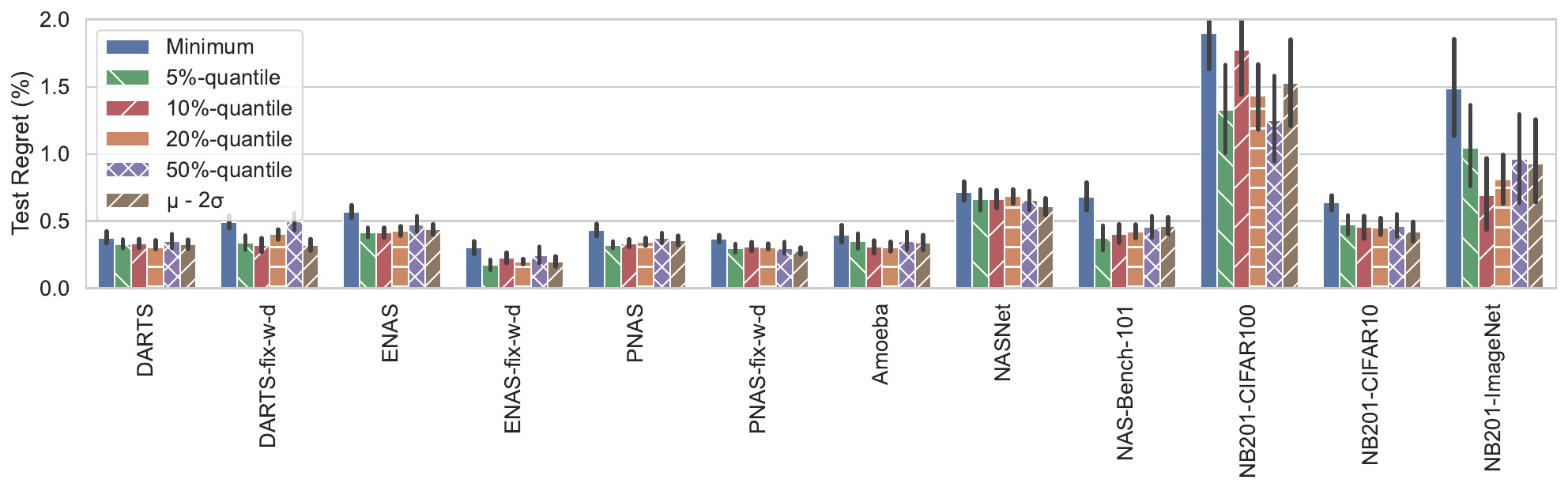}
        \subcaption{Comparison of different methods to compute lower bounds.}
    \end{subfigure}
    \begin{subfigure}{\linewidth}
        \centering
        \includegraphics[width=\textwidth]{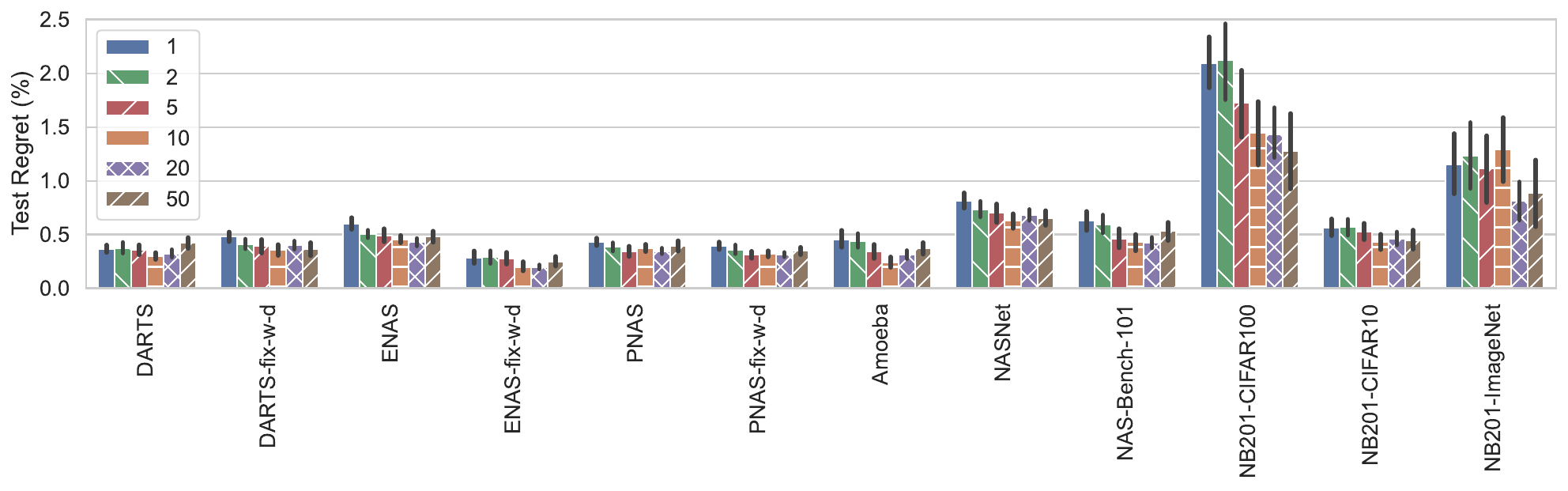}
        \subcaption{Comparison of different relevance scales.}
    \end{subfigure}
    \caption{Comparison of different design choices in computing relevance scores, when budget is small (30).}
    \label{fig:ablation-ndcg-b30}
\end{figure}

To verify whether \cref{eq:accuracy-relevance} still works well when the sampled architectures become fewer, we also repeat the experiments in \cref{fig:ablation-ndcg-outlier-removal} and \cref{fig:ablation-ndcg-relevance-scale} with a smaller budget (30). Intuitively, when the budget is smaller, the estimation of 20\%-quantile is less accurate because it is calculated on fewer samples, but the overall conclusions are similar to what we have made with budget 110. The results are shown in  \cref{fig:ablation-ndcg-b30}.

\subsection{Performance under different budgets}
\label{sec:comparison-different-budgets}

\begin{figure*}[ht]
    \centering
    \includegraphics[width=\textwidth]{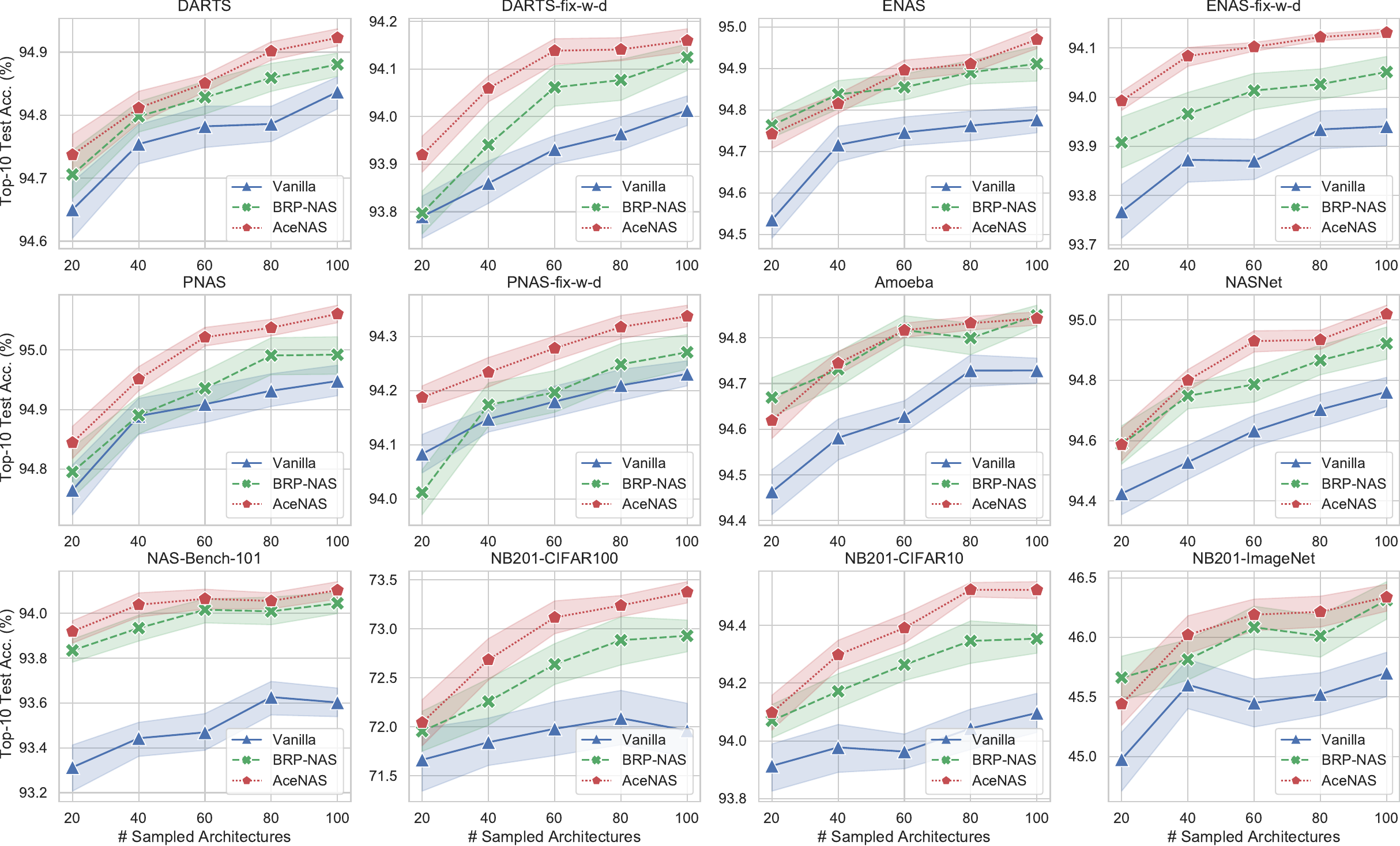}
    \caption{{\algname} consistently surpasses Vanilla and BRP-NAS. Shaded regions here refer to 95\% confidence interval. Please note that the number of samples on the x-axis does not take the final 10 samples (\ie, top-10) into account.}
    \label{fig:top10-best}
\end{figure*}

To show our effective under different budgets, we compare the highest accuracy of top-10 architectures returned by the ranker (\ie top-10 accuracy), and we repeat each experiment for 50 runs. We show the results in \cref{fig:top10-best}. On all 12 benchmarks, \algname{} consistently finds a better architecture than the two most-related GCN baselines (\ie Vanilla and BRP). Remarkably, \algname{} achieves higher accuracy than Vanilla (up to 0.62\%) and BRP-NAS (up to 0.11\%) under the smallest budget (\ie 20).

\subsection{Iterative sampling visualization}
\label{sec:iterative-sampling}

\begin{figure*}[htbp]
\centering
\includegraphics[width=\textwidth]{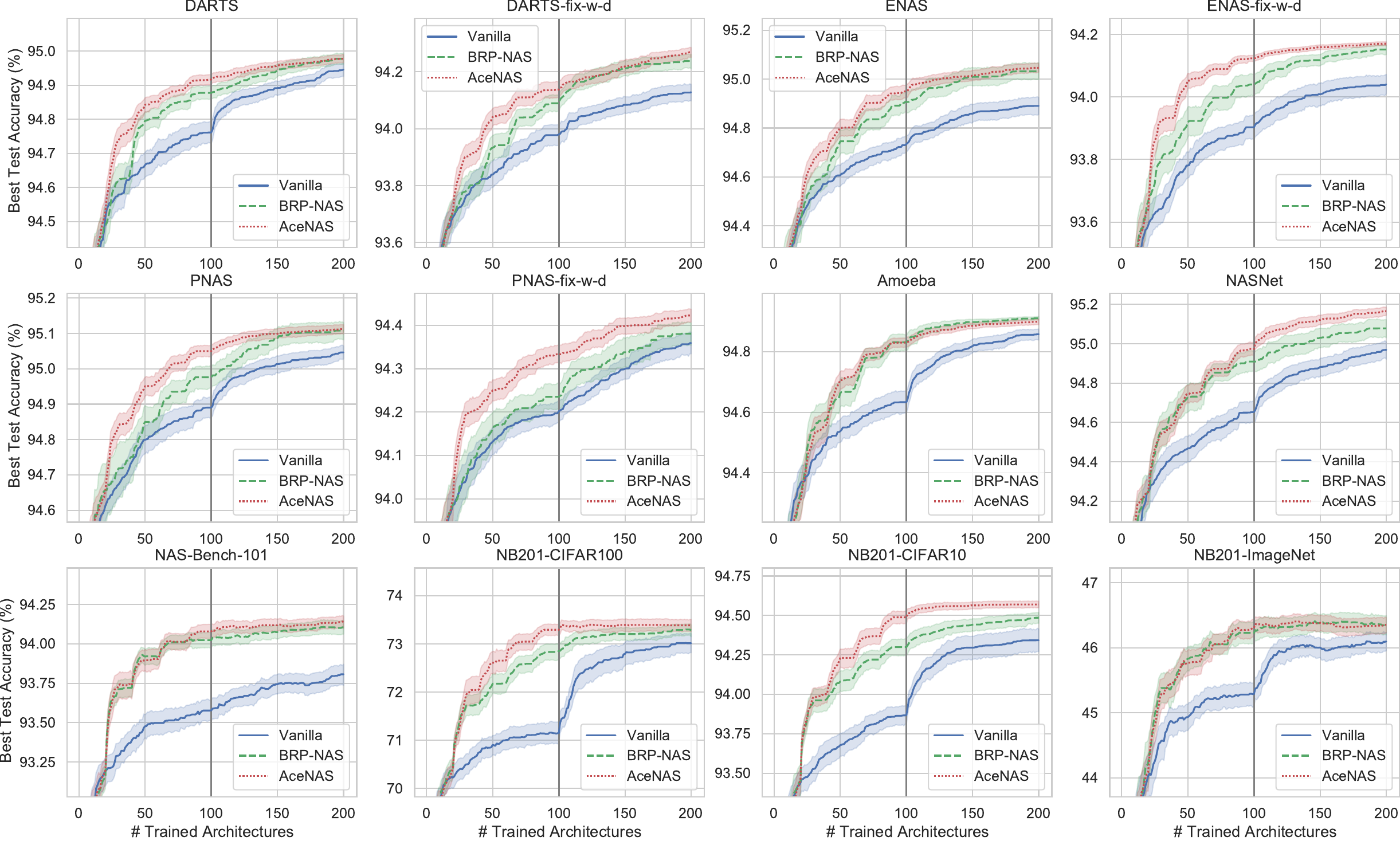}
\caption{The test accuracy of the architecture with best validation accuracy, with respect to number of architectures trained. After 100 architectures (the vertical black line), the ranker does not update any more. The budget of 100 architectures is splitted into 5 rounds. Each line is an average of 50 runs. The shaded regions are 95\% confidence interval.}
\label{fig:nplusk}
\end{figure*}

In \cref{fig:nplusk}, we visualize the whole process of \algname{} on NAS benchmarks. As the 100 architectures are iteratively sampled in 5 folds, we can see clear jumps on the curve at the point of 20, 40, 60, 80 and 100.

\begin{figure}[htbp]
\centering
    \begin{subfigure}{.4\linewidth}
        \centering
        \includegraphics[width=\textwidth]{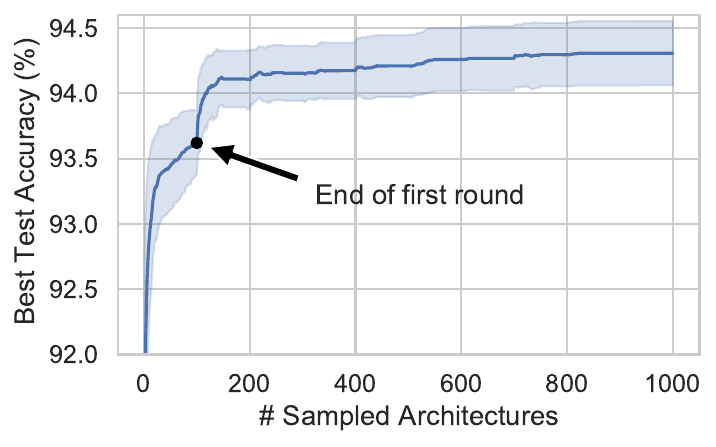}
        \subcaption{NAS-Bench-101 (budget = 1000)}
        \label{fig:nasbench101-process}
    \end{subfigure}
    \begin{subfigure}{.4\linewidth}
        \centering
        \includegraphics[width=\textwidth]{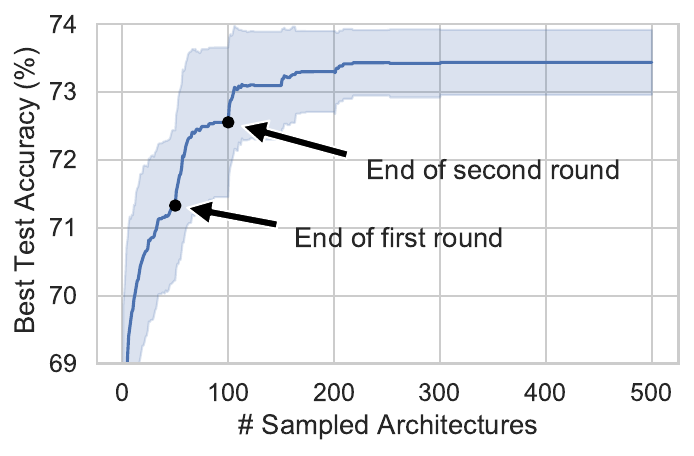}
        \subcaption{NAS-Bench-201, CIFAR-100 (budget = 500)}
        \label{fig:nasbench201-process}
    \end{subfigure}
    \begin{subfigure}{.6\linewidth}
        \centering
        \includegraphics[width=\textwidth]{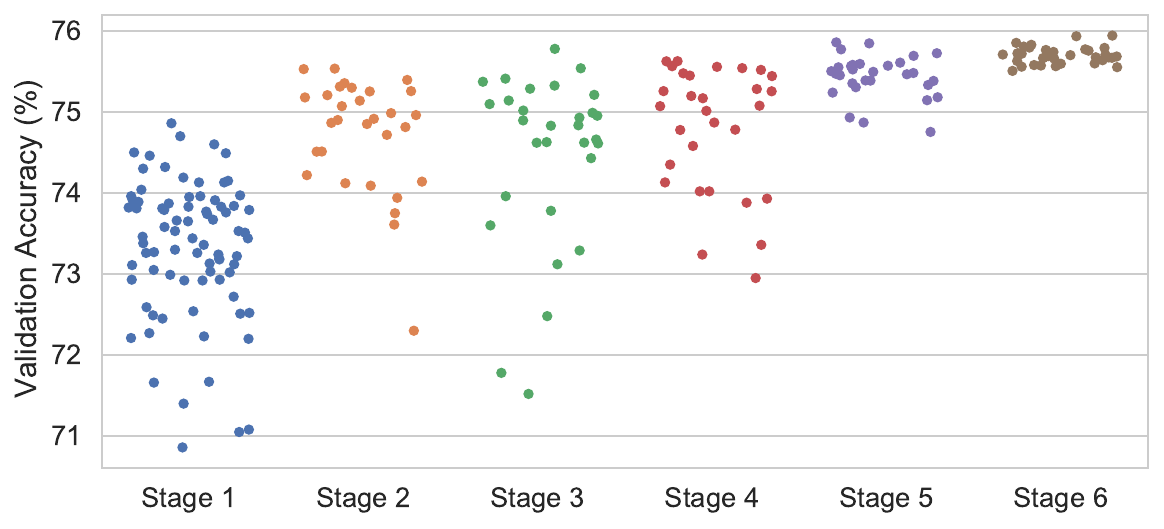}
        \subcaption{\algname{} optimization process on ProxylessNAS.}
        \label{fig:proxyless-process}
    \end{subfigure}
\caption{On NAS-Bench-101 (left), with total budget of 1000, we update the ranker every 100 samples. The black dot denotes the first time ranker is updated. On NAS-Bench-201 (CIFAR-100) (right), with total budget of 500, we update the ranker every 50 samples. The black dots denote the first and second time ranker is updated.}
\label{fig:nasbench101-201-process}
\end{figure}

Given sufficient budgets, \algname{} can improve even more. In \cref{fig:nasbench101-process}, we search on NAS-Bench-101. With a total budget of 1000, we fine-tune the ranker when every 100 samples are collected. In \cref{fig:nasbench201-process}, we search on NAS-Bench-201 (CIFAR-100). We divide the total budget of 500 into 10 rounds (50 each). The accuracy bumps after the first round indicates the effectiveness of ranker on selecting good-performing architectures. Notably, on NAS-Bench-101, accuracy continues to improve after over 800 architectures are sampled. This result demonstrates that \algname{} is not likely to fall into local optimum despite fast convergence.

% \yuge{Modify this figure.}
Similarly, we show the sampling process on ProxylessNAS search space. In \cref{fig:proxyless-process}, we conduct 6 stages of evolution. In the first phase, 80 architectures are sampled, and in the rest stages, 30 stages on sampled. We guarantee all the architectures sampled satisfy our latency constraint. % In the final stage, we exploit the ranker by using evolution to find the architectures with the best predicted scores.

\subsection{More ablation study on ranking loss}
\label{sec:more-ablation-study}

\begin{figure}[htbp]
\centering
    \begin{subfigure}{.48\linewidth}
        \centering
        \includegraphics[width=\textwidth]{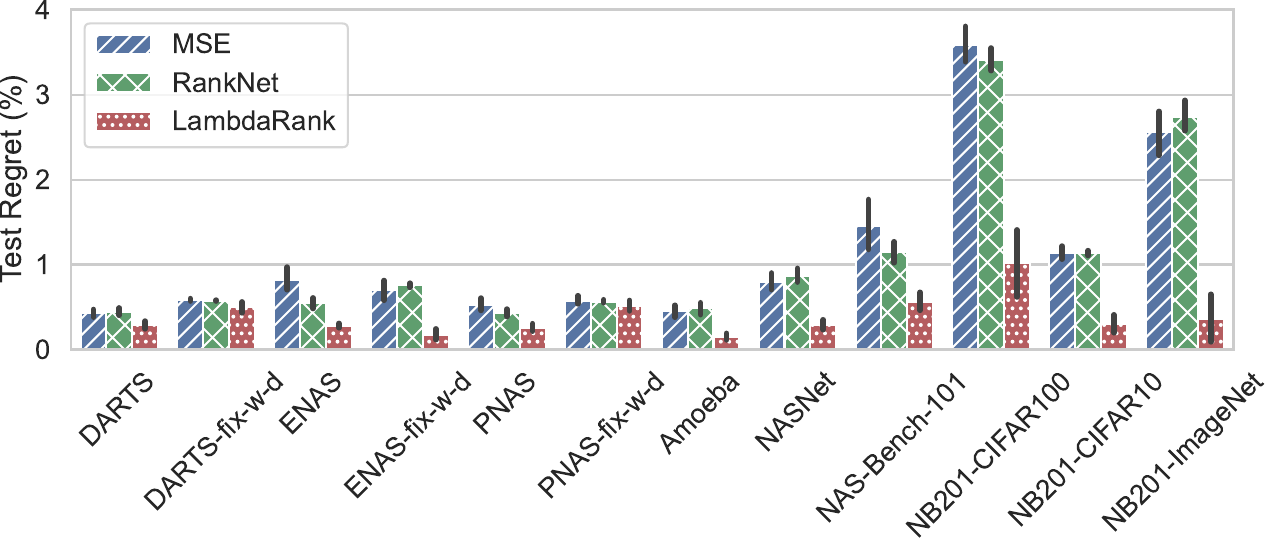}
        \subcaption{No pre-trainining.}
        % \label{fig:nasbench101-process}
    \end{subfigure}
    \begin{subfigure}{.48\linewidth}
        \centering
        \includegraphics[width=\textwidth]{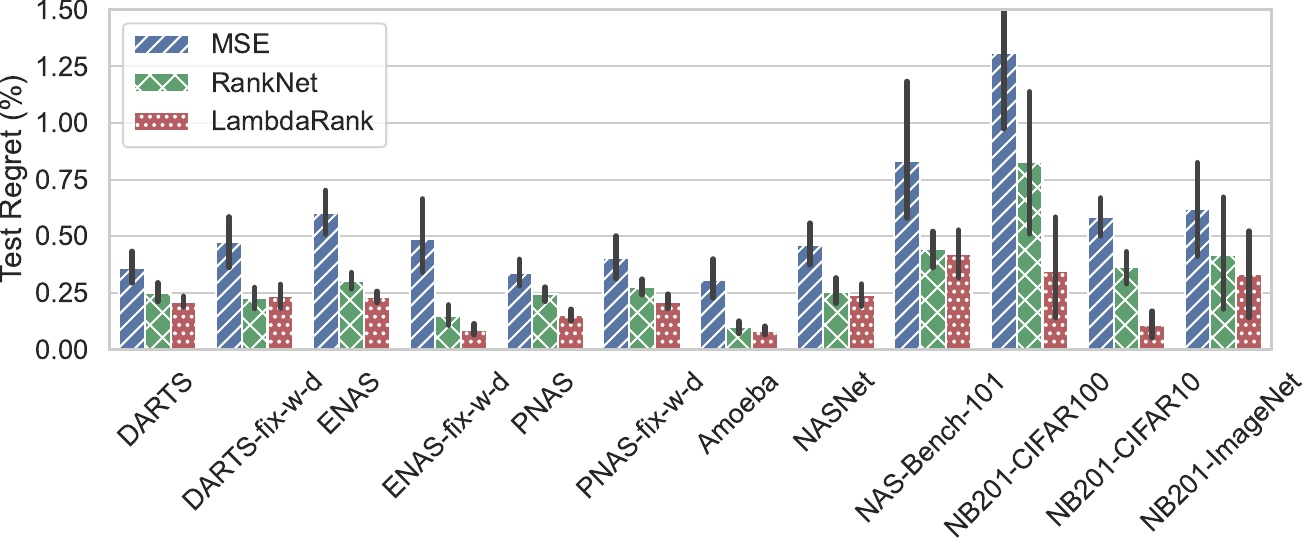}
        \subcaption{Pre-trainining with Parameters \& FLOPs.}
        % \label{fig:nasbench201-process}
    \end{subfigure}
\caption{Comparison among test regrets when using different LTR losses. \textbf{(left)}: when no pre-trainin is use. \textbf{(right)}: when pretraining with ``Parameters and FLOPs'' as weak labels. Each bar is an average of 50 runs.}
\label{fig:ablation-study-wrt-different-pretrain}
\end{figure}

We include the ablation study of ranking loss conditioned on multiple settings of pre-training, \ie no pre-trainng at all and pre-training with parameters and FLOPs. The results are shown in \cref{fig:ablation-study-wrt-different-pretrain}. We observe a consistent superiority that LambdaRank is better than RankNet, and RankNet is better than vanilla MSE loss.

\subsection{\algname{} with different weight-sharing approaches}
\label{sec:combination-weight-sharing}

In \cref{sec:acenas-weight-sharing}, we combine \algname{} with 6 different weight sharing methods, and show the performance when each method is used to obtain the weak labels for \algname{}. In \cref{tab:ws-top-10-accuracy-full}, we show results on more search spaces. Note that we only run experiments on 6 instead of all 12 search spaces because supporting several methods on some of the search spaces are not straightforward (\eg NAS-Bench-101). The conclusions are similar to what we have made in \cref{sec:acenas-weight-sharing}. One extra observation is that, the performance of weight-sharing is highly related to the choice of search space. There is no weight-sharing method that performs consistently well on all search spaces. This echoes the findings in \cite{zhang2020does}.

\begin{figure}[ht]
    \centering
    \includegraphics[width=0.8\textwidth]{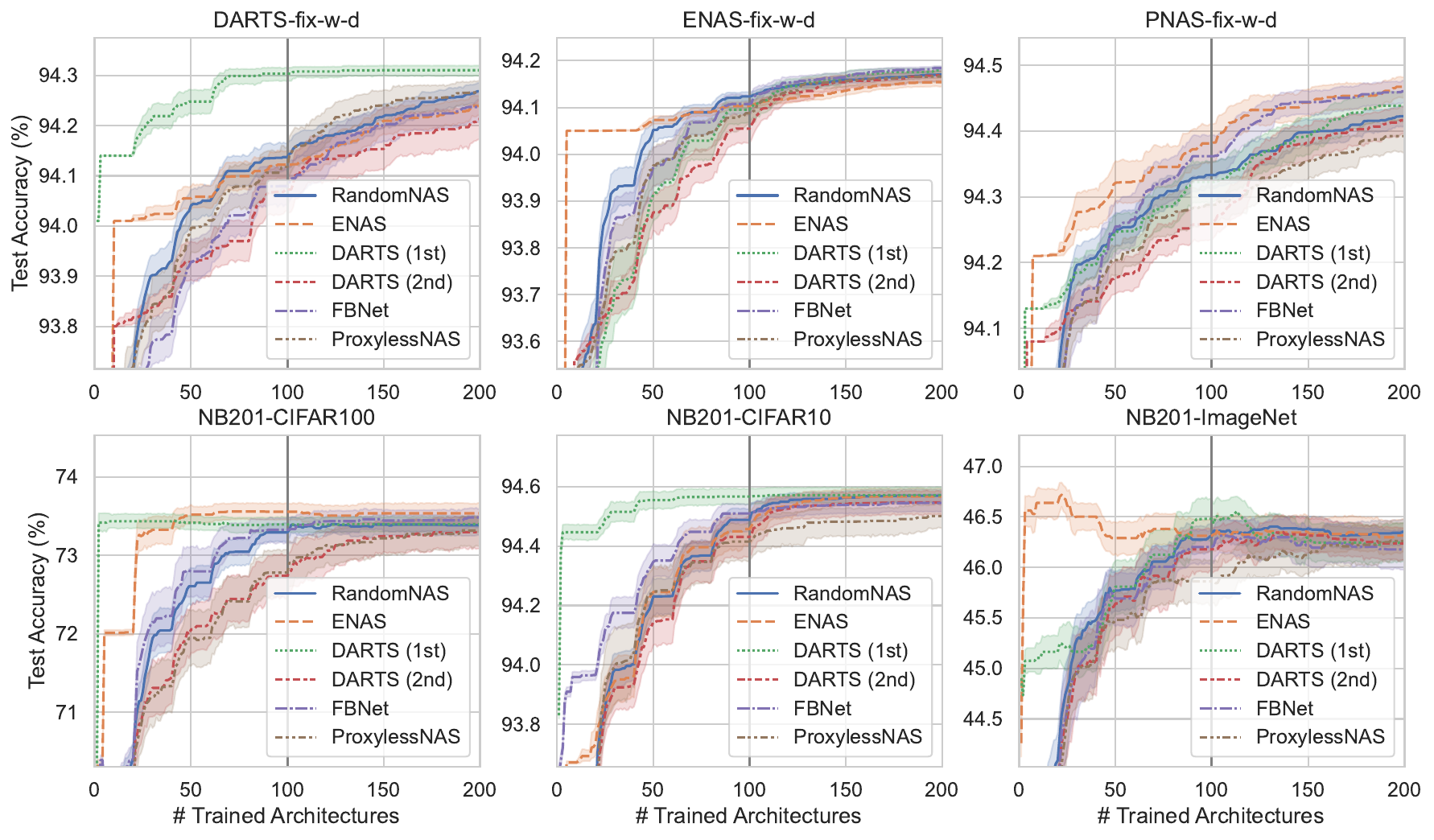}
    \caption{Iterative sampling visualization of \algname{} + different weight sharing approaches. Settings resemble \cref{fig:nplusk}: first 100 samples are used to fine-tune the ranker, and the ranker itself is fixed after the vertical line.}
    \label{fig:ws-gcn-train-plot200}
\end{figure}

Similar to \cref{fig:nplusk}, we visualize the iterative sampling process in \cref{fig:ws-gcn-train-plot200} and compare the combination of \algname{} with different weight sharing NAS. It can be seen that the methods with a better starting point tend to have a better final performance. Another notable phenomenon is that are NAS-Bench-201-ImageNet, \algname{} + ENAS first finds an architecture with accuracy exceeding 46.5\%, and then decreases to around 46.3\%. This is a result of the discrepancy between validation accuracy and test accuracy, and the algorithm coincidentally finds an architecture that is good performing on test dataset.

% \cref{tab:ws-top-10-accuracy-full} and \cref{fig:ws-gcn-train-plot200}.

\begin{table}[ht]
\caption{Top-10 test accuracy when \algname{} is combined with weak labels obtained by different weight-sharing methods. Each number is an average of 50 runs. The best method in each column is highlighted with bold text, while the worst method is underlined.}
\label{tab:ws-top-10-accuracy-full}
\begin{adjustbox}{width=\linewidth}
\begin{tabular}{ccccccccccccc}
\toprule
\multirow{2}{*}{Method} & \multicolumn{2}{c}{NB201-CIFAR100} & \multicolumn{2}{c}{NB201-CIFAR10} & \multicolumn{2}{c}{NB201-ImageNet} & \multicolumn{2}{c}{DARTS-fix-w-d} & \multicolumn{2}{c}{ENAS-fix-w-d} & \multicolumn{2}{c}{PNAS-fix-w-d} \\
{} &            WS quality &              \algname{} &            WS quality &              \algname{} &             WS quality &              \algname{} &            WS quality &              \algname{} &            WS quality &              \algname{} &            WS quality &              \algname{} \\
\midrule
RandomNAS~\cite{li2019random}        &              69.70 &              73.38 &              93.16 &              94.52 &               43.26 &              46.34 &              93.52 &              94.16 &              93.45 &     \textbf{94.13} &              93.87 &              94.34 \\
DARTS (1st order)~\cite{liu2018darts}     &     \textbf{73.44} &              73.40 &     \textbf{94.45} &     \textbf{94.57} &               45.10 &     \textbf{46.51} &     \textbf{94.14} &     \textbf{94.30} &  \underline{92.96} &              94.11 &              94.13 &              94.33 \\
DARTS (2nd order)~\cite{liu2018darts} &  \underline{55.70} &  \underline{72.93} &  \underline{84.60} &              94.46 &   \underline{28.15} &              46.19 &              93.80 &              94.12 &              93.56 &  \underline{94.09} &              94.08 &  \underline{94.27} \\
FBNet~\cite{wu2019fbnet}       &              70.01 &              73.40 &              93.96 &              94.51 &               41.49 &              46.30 &  \underline{93.50} &  \underline{94.10} &              93.13 &              94.12 &  \underline{93.54} &              94.37 \\

ProxylessNAS~\cite{cai2019proxylessnas}    &              59.69 &              72.93 &              87.43 &  \underline{94.44} &               37.20 &  \underline{45.97} &              93.65 &              94.16 &              93.53 &              94.11 &              93.86 &              94.29 \\
ENAS~\cite{pham2018efficient}    &              72.02 &     \textbf{73.56} &              93.67 &              94.50 &      \textbf{46.64} &              46.34 &              94.01 &              94.12 &     \textbf{94.05} &              94.11 &     \textbf{94.21} &     \textbf{94.39} \\
\bottomrule
\end{tabular}
\end{adjustbox}

\end{table}

\subsection{\algname{} vs. pure weight-sharing}
\label{sec:acenas-vs-pure-ws}

\begin{figure}[htbp]
    \centering
    \includegraphics[width=.5\linewidth]{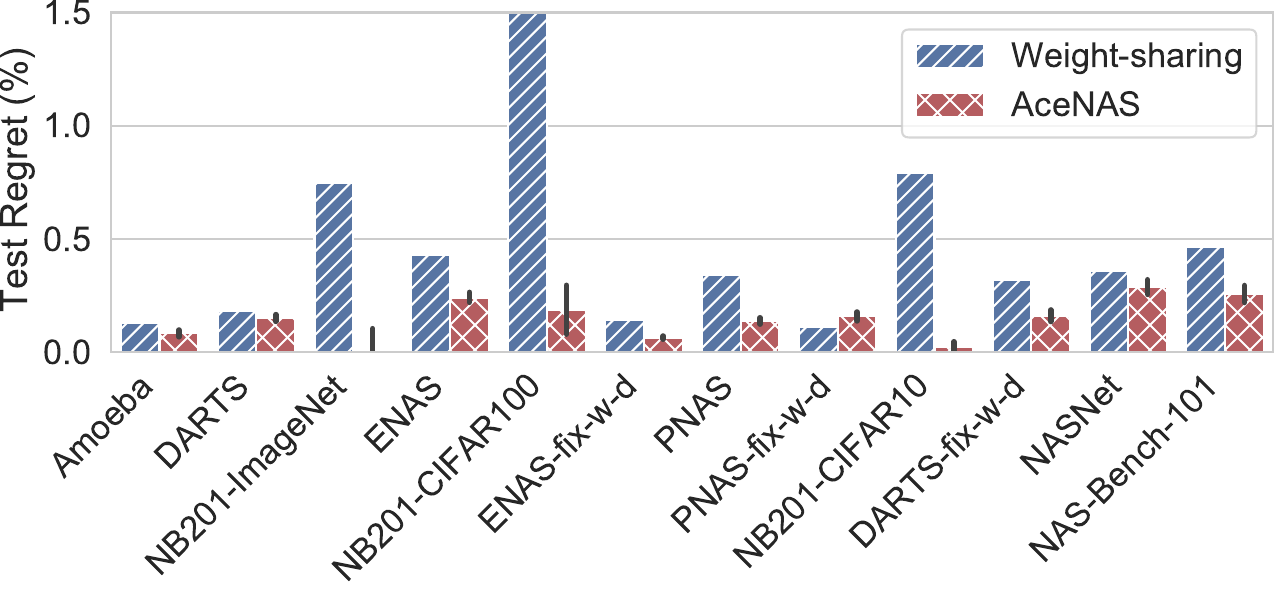}
    \caption{Comparison of test regret between weight-sharing-guided greedy search and \algname{} on 100 budget.}
    \label{fig:weight-sharing-comparison}
    % \vspace{-1.5em}
\end{figure}

Another natural question is, would pure weight-sharing without \algname{} be competitive with \algname{} if they are given more budgets to fully train architectures? Here, we compare \algname{} against a ``weight sharing guided search''~\cite{pourchot2020share} that selects top-100 architectures with the highest accuracy on weight-shared super-net. For fair comparison, \algname{} samples the same number of architectures. Specifically, 80 fully-trained architectures are iterative sampled, while the final ranking models predict the best 20 architectures.  As shown in \cref{fig:weight-sharing-comparison}, \algname{} outperforms the weight-sharing based prediction. Remarkably, we reduce the test regret by 3\% on NAS-Bench-201 (CIFAR-100). This experiment indicates that \algname{} is a necessary component, and weight-sharing alone does not work well.

\subsection{Pretraining quality}

\begin{figure}[htbp]
\centering
\includegraphics[width=0.5\linewidth]{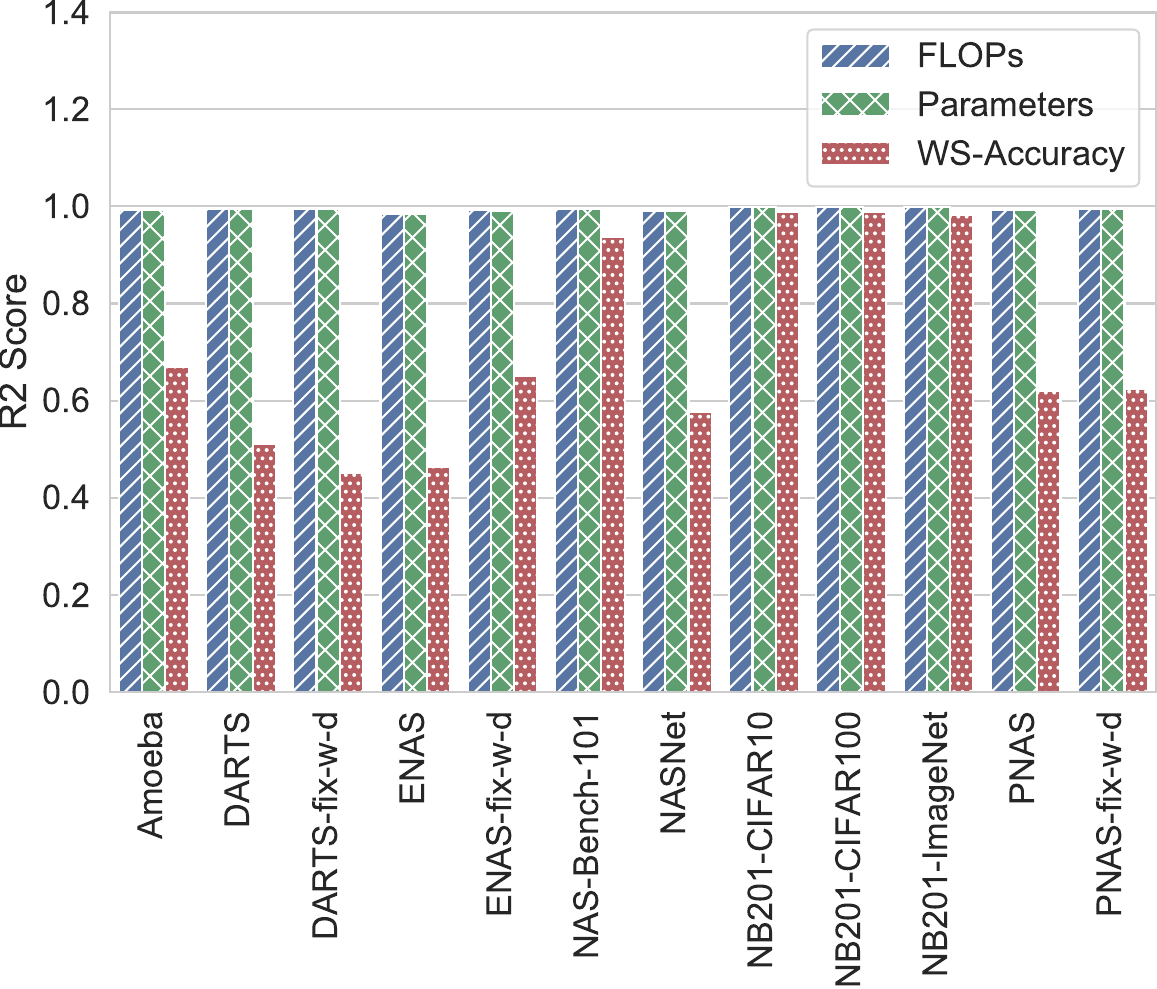}
\caption{The validation $R^2$ score in pretraining stage, indicating how good our ranker is good at predicting weight sharing accuracy, number of parameters and FLOPs.}
\label{fig:pretraining-quality}
\end{figure}

In \cref{fig:pretraining-quality}, we show how our ranker is good at capturing the information (\ie WS-accuracy, FLOPs and number of parameters) in the pretraining stage. Apart from 4k architectures that were used in training, we sampled extra 1k architectures per search space to evaluate the model. We compute $R^2$ scores (coefficient of determination), which can be as good as 1. We found that for parameters and FLOPs, our model hits $>0.98$ for all search spaces, which means that the model is very good at predicting the model size. On WS-accuracy, $R^2$ scores vary between 0.4 and 1, implying that it always learn information from super-net, at least to some extent. Clearly, on some benchmarks (\eg NAS-Bench-101 and NAS-Bench-201), it looks better than others. We conjecture that search spaces with lower diversity (\eg spaces in NAS-Bench series) are easier to learn.

\subsection{Quality of searched architectures}

For NAS benchmarks, we show in \cref{tab:architecture-rank} the test accuracy, test regret and rank of our searched architecture. \emph{\algname{} is approximately able to find the best architecture out of one thousand architectures.} Remarkably, on NAS-Bench-201-ImageNet, it almost finds the best validation architecture on the search space (the negative test regret estimation is caused by variance). Despite that the best validation architecture has been located, it still ranks 2.14 out of 1000, which means that some architectures have a even better test accuracy, which is due to the gap between validation dataset and test dataset, and the best on validation is not necessarily the best on test.

\begin{table}[htbp]
    \centering
    \caption{Test accuracy and rank of the best architecture, averaged over 50 runs. For test accuracy, we report the mean and standard deviation. For test regret and rank, we only report the mean. $^*$: the gap between test accuracy of searched architecture and the best validation architecture. $^\dag$: the average rank of test accuracy within all test accuracies.}
    \label{tab:architecture-rank}
    \small
    \begin{tabular}{cccc}
    \toprule
    Search space & Test acc. & Test regret$^*$ & Rank (\textperthousand) $^\dag$ \\
    \midrule
    NAS-Bench-101 & \meanstd{94.10}{0.20} & 0.24 & 0.33 \\
    NB201-CIFAR100 & \meanstd{73.38}{0.51} & 0.10 & 0.16 \\
    NB201-CIFAR10 & \meanstd{94.52}{0.15} & 0.04 & 0.05 \\
    NB201-ImageNet & \meanstd{46.34}{0.56} & -0.08 & 2.14 \\
    Amoeba & \meanstd{94.84}{0.07} & 0.09 & 0.80 \\
    DARTS & \meanstd{94.92}{0.07} & 0.14 & 1.20 \\
    DARTS-fix-w-d & \meanstd{94.16}{0.12} & 0.16 & 1.00 \\
    ENAS & \meanstd{94.97}{0.11} & 0.20 & 0.60 \\
    ENAS-fix-w-d & \meanstd{94.13}{0.04} & 0.06 & 0.40 \\
    NASNet & \meanstd{95.02}{0.15} & 0.25 & 0.41 \\
    PNAS & \meanstd{95.06}{0.06} & 0.13 & 0.60 \\
    PNAS-fix-w-d & \meanstd{94.34}{0.09} & 0.16 & 1.54 \\
    \bottomrule
    \end{tabular}

\end{table}

\begin{figure*}[htbp]
\begin{subfigure}{\textwidth}
\includegraphics[trim={0 2cm 0 1.5cm},clip,width=\textwidth]{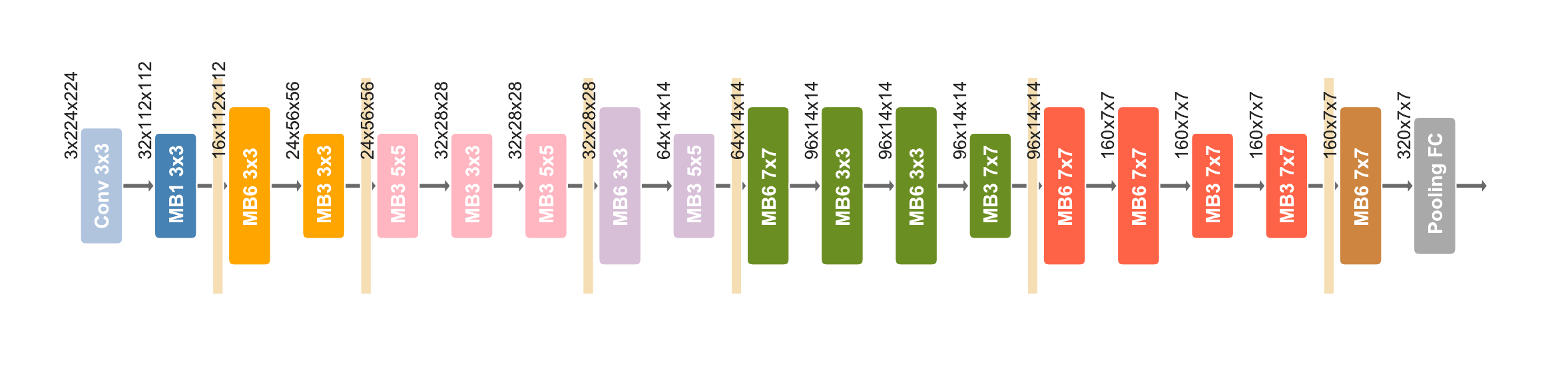}
\caption{AceNAS-M1}
\end{subfigure}
\hfill
\begin{subfigure}{\textwidth}
\includegraphics[trim={0 2cm 0 1.5cm},clip,width=\textwidth]{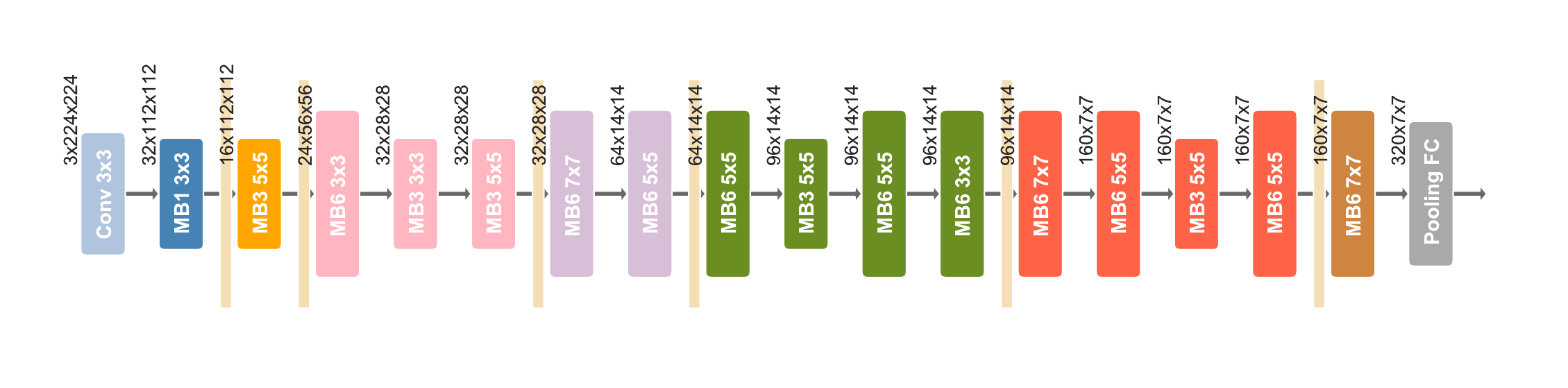}
\caption{AceNAS-M2}
\end{subfigure}
\hfill
\begin{subfigure}{\textwidth}
\includegraphics[trim={0 2cm 0 1.5cm},clip,width=\textwidth]{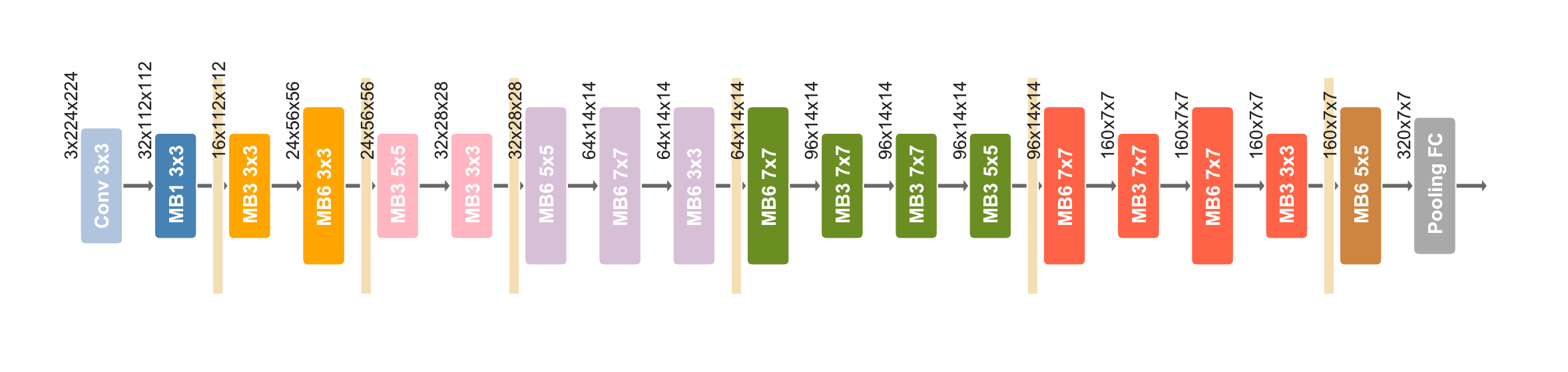}
\caption{AceNAS-M3}
\end{subfigure}
\caption{Searched architecture on ProxylessNAS search space.}
\label{fig:acenas-m}
\end{figure*}

For ProxylessNAS search space, we show the architectures found by 3 different runs, and name them AceNAS-M1, AceNAS-M2, and AceNAS-M3, respectively (M for Mobile). The network structures are shown in \cref{fig:acenas-m} and the accuracy and latency on ImageNet test set (commonly called validation set for historical reasons) are shown in \cref{tab:acenas-m-acc}. Pre-trained checkpoints of these models will be released.

\begin{table}[htbp]
    \centering
    \caption{Test accuracy and latency of architectures searched on ProxylessNAS (shown in \cref{fig:acenas-m}).}
    \label{tab:acenas-m-acc}
    \begin{tabular}{ccc}
    \toprule
    Architecture & Test acc. (\%) & Latency (ms) \\
    \midrule
    AceNAS-M1 & 75.25 & 84.60 \\
    AceNAS-M2 & 75.07 & 84.59 \\
    AceNAS-M3 & 75.11 & 84.92 \\
    \bottomrule
    \end{tabular}
\end{table}

%%%%%%%%%%%%%%%%%%%%%%%%%%%%%%%%%%%%%%%%%%%%%%%%%%%%%%%%%%%%%%%%%%%%%%%%%%%%%%%
%%%%%%%%%%%%%%%%%%%%%%%%%%%%%%%%%%%%%%%%%%%%%%%%%%%%%%%%%%%%%%%%%%%%%%%%%%%%%%%

\end{document}